\documentclass[10pt,journal,compsoc]{IEEEtran}
\ifCLASSOPTIONcompsoc
\usepackage[utf8]{inputenc} % allow utf-8 input
\usepackage[T1]{fontenc}    % use 8-bit T1 fonts
\usepackage{hyperref}       % hyperlinks
\usepackage{url}            % simple URL typesetting
\usepackage{booktabs}       % professional-quality tables
\usepackage{amsfonts}       % blackboard math symbols
\usepackage{nicefrac}       % compact symbols for 1/2, etc.
\usepackage{microtype}      % microtypography
\usepackage{bm}
\usepackage{amsmath}
\usepackage{algorithmic}
\usepackage{algorithm}
\usepackage{multirow}
\usepackage{graphicx}
\usepackage{subfigure}
\usepackage{wrapfig}
\usepackage{amsthm}
\usepackage{multicol}
\usepackage{amssymb}
\usepackage{pdfpages}
\usepackage{makecell}

\usepackage{marvosym}

\usepackage{xcolor}
\definecolor{_green}{RGB}{82,208,83}

\usepackage{pifont}% http://ctan.org/pkg/pifont
\newcommand{\cmark}{\ding{51}}%
\newcommand{\xmark}{\ding{55}}%

  \usepackage[nocompress]{cite}
\else
  \usepackage{cite}
\fi
\ifCLASSINFOpdf
\else
\fi

\usepackage{stfloats}
\vspace{7.5in}
\IEEEpubid{
\fbox{
\parbox{0.975\textwidth}{
\copyright 2024 IEEE. Personal use of this material is permitted. Permission from IEEE must be obtained for all other uses, in any current or future media, including\hfill
reprinting/republishing this material for advertising or promotional purposes, creating new collective works, for resale or redistribution to servers or lists, or reuse of\hfill
any copyrighted component of this work in other works. \vspace{-0.5mm}
}%
}}

\begin{document}

\title{Uni-AdaFocus: Spatial-temporal Dynamic Computation for Video Recognition}


% \author{\IEEEmembership{TPAMI}

\author{
  Yulin~Wang$^{\bm{\dagger}}$,
  Haoji~Zhang$^{\bm{\dagger}}$,
  Yang~Yue,
  % Yansong~Tang,~\IEEEmembership{Member,~IEEE},
  Shiji~Song,~\IEEEmembership{Senior~Member,~IEEE}, \\
  Chao~Deng,
  Junlan~Feng,~\IEEEmembership{Fellow,~IEEE},
  and~Gao~Huang~\!\textsuperscript{\Letter},~\IEEEmembership{Member,~IEEE}
  \IEEEcompsocitemizethanks{
  \IEEEcompsocthanksitem 
  Y. Wang, Y. Yue, S. Song, and G. Huang are with the Department of Automation, BNRist, Tsinghua University, Beijing, China. 
  % Email: \{wang-yl19, yueyang22\}@mails.tsinghua.edu.cn, \{shijis, gaohuang\}@tsinghua.edu.cn. 
  % \IEEEcompsocthanksitem 
  G. Huang is also with Beijing Academy of Artificial Intelligence, Beijing, China. 
  H. Zhang is with the Shenzhen International Graduate School, Tsinghua University, Guangdong, China. 
  % H. Zhang and Y. Tang are with the Shenzhen International Graduate School, Tsinghua University, Guangdong, China. 
  % Email: \{hodge013140, tangyansong15\}@gmail.com.
  % \IEEEcompsocthanksitem 
  C. Deng and J. Feng are with the China Mobile Research Institute, Beijing, China.
  \IEEEcompsocthanksitem
  Email: wang-yl19@mails.tsinghua.edu.cn, gaohuang@tsinghua.edu.cn. 
  % Email: \{dengchao, fengjunlan\}@chinamobile.com.
  \IEEEcompsocthanksitem 
  $^{\bm{\dagger}}$Equal contribution.\ \ \ \ \  \textsuperscript{\Letter}Corresponding author.
}}

% and the end of the author line. \emph{i.e.}, if you had this:

\markboth{Uni-AdaFocus: Spatial-temporal Dynamic Computation for Video Recognition}%
{Shell \MakeLowercase{\textit{et al.}}: Bare Advanced Demo of IEEEtran.cls for IEEE Computer Society Journals}



\IEEEtitleabstractindextext{%
\begin{abstract}

  This paper presents a comprehensive exploration of the phenomenon of data redundancy in video understanding, with the aim to improve computational efficiency. Our investigation commences with an examination of \emph{spatial redundancy}, which refers to the observation that the most informative region in each video frame usually corresponds to a small image patch, whose shape, size and location shift smoothly across frames. Motivated by this phenomenon, we formulate the patch localization problem as a dynamic decision task, and introduce a spatially adaptive video recognition approach, termed AdaFocus. In specific, a lightweight encoder is first employed to quickly process the full video sequence, whose features are then utilized by a policy network to identify the most task-relevant regions. Subsequently, the selected patches are inferred by a high-capacity deep network for the final prediction. The complete model can be trained conveniently in an end-to-end manner. During inference, once the informative patch sequence has been generated, the bulk of computation can be executed in parallel, rendering it efficient on modern GPU devices. Furthermore, we demonstrate that AdaFocus can be easily extended by further considering the \emph{temporal} and \emph{sample-wise} redundancies, \emph{i.e.}, allocating the majority of computation to the most task-relevant video frames, and minimizing the computation spent on relatively ``easier'' videos. Our resulting algorithm, Uni-AdaFocus, establishes a comprehensive framework that seamlessly integrates spatial, temporal, and sample-wise dynamic computation, while it preserves the merits of AdaFocus in terms of efficient end-to-end training and hardware friendliness. In addition, Uni-AdaFocus is general and flexible as it is compatible with off-the-shelf backbone models (\emph{e.g.}, TSM and X3D), which can be readily deployed as our feature extractor, yielding a significantly improved computational efficiency. Empirically, extensive experiments based on seven widely-used benchmark datasets (\emph{i.e.}, ActivityNet, FCVID, Mini-Kinetics, Something-Something V1\&V2, Jester, and Kinetics-400) and three real-world application scenarios (\emph{i.e.}, fine-grained diving action classification, Alzheimer's and Parkinson's diseases diagnosis with brain magnetic resonance images (MRI), and violence recognition for online videos) substantiate that Uni-AdaFocus is considerably more efficient than the competitive baselines. Code \& pre-trained models are available at \url{https://github.com/blackfeather-wang/AdaFocus}, \url{https://github.com/LeapLabTHU/AdaFocusV2}, and \url{https://github.com/LeapLabTHU/Uni-AdaFocus}.

\end{abstract}

}

\maketitle

% \begingroup\renewcommand\thefootnote{\IEEEauthorrefmark{1}}
% \footnotetext{\emph{Equal contribution.}}

% not be used in maketitle, but will appear (\emph{i.e.}, to be "transported")
% is not selected <OR> if conference mode is selected - because compsoc
% conference papers position the abstract like regular (non-compsoc)
\IEEEdisplaynontitleabstractindextext

% \fi
\IEEEpeerreviewmaketitle

% \vspace{-1ex}
\section{Introduction}
% \vspace{-1ex}
\label{sec:introduction}

% The explosive growth of online videos (\emph{e.g.}, on YouTube or TikTok) has fueled the demands for automatically recognizing human actions, events, or other contents within them, which benefits applications like recommendation \cite{davidson2010youtube, deldjoo2016content, gao2017unified}, surveillance \cite{collins2000system, chen2019distributed} and content-based searching \cite{ikizler2007searching}. 

The proliferation of online videos, exemplified by the platforms such as YouTube and TikTok, has necessitated the development of automated methods for identifying human actions, events, and other elements within them. This is crucial for facilitating the applications such as recommendation \cite{davidson2010youtube, deldjoo2016content, gao2017unified}, surveillance \cite{collins2000system, chen2019distributed}, and content-based searching \cite{ikizler2007searching}. In recent years, remarkable success in accurate video recognition has been achieved by leveraging deep networks \cite{feichtenhofer2019slowfast, zhu2017deep, feichtenhofer2016convolutional, carreira2017quo, tran2015learning, hara2018can}. However, the noteworthy performance of these models usually comes at the price of high computational costs. In real-world scenarios, computation directly translates into power consumption, carbon emission and practical latency, which should be minimized under economic, environmental or safety considerations.

% The proliferation of online video content, exemplified by platforms such as YouTube and TikTok, has necessitated the development of automated methods for identifying human actions, events, and other elements within these media. This is crucial for facilitating applications such as recommendation \cite{davidson2010youtube, deldjoo2016content, gao2017unified}, surveillance \cite{collins2000system, chen2019distributed}, and content-based searching \cite{ikizler2007searching}. 

% In recent years, considerable advancements in accurate video recognition has been achieved through the utilization of deep learning architectures \cite{feichtenhofer2019slowfast, zhu2017deep, feichtenhofer2016convolutional, carreira2017quo, tran2015learning, hara2018can}. Despite their noteworthy performance, these models are often associated with substantial computational overhead. It is crucial to address these concerns, as computation directly impacts power consumption, carbon emissions, and real-world latency, which must be minimized to ensure both economic viability and safety.

\begin{figure}[t]
    % \vskip -0.1in
    \begin{center}
    \centerline{\includegraphics[width=0.95\columnwidth]{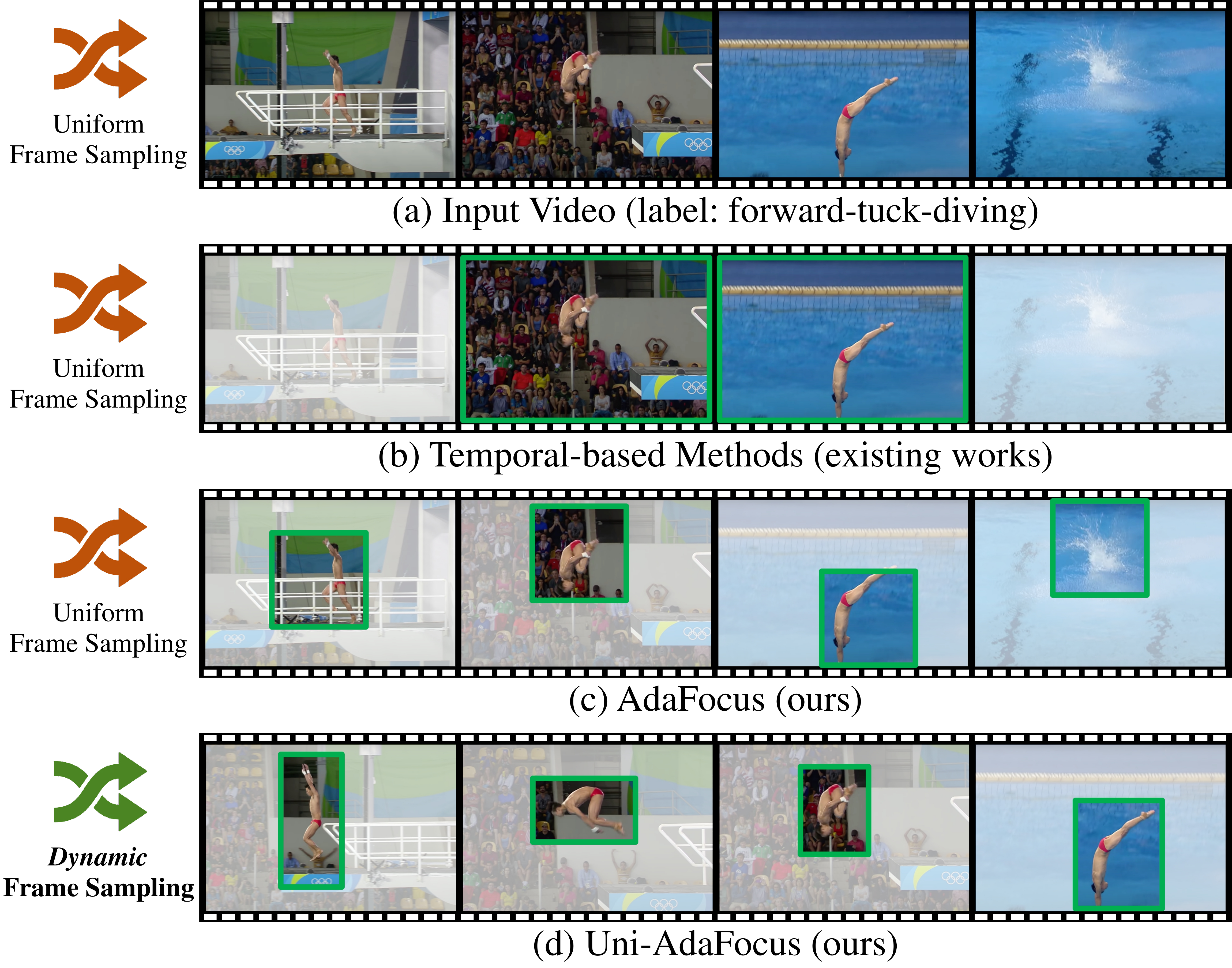}}
    \vskip -0.15in
    \caption{\textbf{Comparisons between existing temporal-based methods and our proposed approaches.} Most existing works aim to reduce computational costs by selecting a few informative frames to process. Orthogonal to them, AdaFocus reveals that a superior computational efficiency can be achieved by reducing the \emph{spatial redundancy}. Built upon this finding, we further demonstrate that it is feasible to formulate flexible and highly efficient spatial-temporal dynamic computation simultaneously in a unified framework (Uni-AdaFocus), \emph{i.e.}, attending to the most valuable spatial regions of the most task-relevant video frames. \label{fig:fig1}
    % The majority of current research focuses on decreasing computational expenses by selecting a limited number of informative frames for processing. In contrast, AdaFocus demonstrates that superior computational efficiency can be attained by capitalizing on the \emph{spatial redundancy}. Building on this insight, we further illustrate the feasibility of formulating a flexible and highly efficient spatial-temporal dynamic computation within a unified framework (Uni-AdaFocus). This is achieved by directing attention to the most valuable spatial regions of the most task-relevant video frames.
    % aims to perform efficient inference by attending to the task-relevant patch of each frame. 
    % Importantly, 
    }
    \end{center}
    % \vspace{-7ex}
    \vskip -0.125in
\end{figure}

% Efforts to mitigate the challenge of computational overhead in video recognition have led to numerous recent studies focusing on reducing the intrinsic \emph{temporal redundancy} \cite{wu2019liteeval, wu2019multi, wu2019adaframe, korbar2019scsampler, gao2020listen, meng2020ar, lin2022ocsampler}. As illustrated in Figure~\ref{fig:fig1} (b), concentrating on the most task-relevant video frames and allocating the majority of computational resources to these frames is an efficient approach. Nevertheless, another significant source of redundant computation in image-based data, specifically \emph{spatial redundancy}, has not been extensively investigated in the realm of efficient video recognition. Research on 2D-image classification has demonstrated that deep networks, such as ConvNets or vision Transformers, can generate accurate predictions by examining only a few discriminative regions within the entire image \cite{mnih2014recurrent, fu2017look, xie2020spatially, han2021dynamic, 9851927, 10091227}. Conducting inference on these relatively small regions can significantly reduce the computational demands of visual backbones, as processing a 96x96 patch necessitates approximately 18\% of the computation required for a 224x224 image.

To address this issue, a number of recent works propose to reduce the inherent \emph{temporal redundancy} in video recognition \cite{wu2019liteeval, wu2019multi, wu2019adaframe, korbar2019scsampler, gao2020listen, meng2020ar, lin2022ocsampler}. As illustrated in Figure~\ref{fig:fig1} (b), it is efficient to concentrate on the most task-relevant video frames, and allocate the majority of computation to them rather than all frames. Nevertheless, another important source of redundant computation in image-based data, specifically \emph{spatial redundancy}, has rarely been explored in the realm of efficient video recognition. In fact, it has been shown in 2D-image classification that deep networks (\emph{e.g.}, ConvNets or vision Transformers) are able to produce correct predictions by examining only a few discriminative regions instead of the entire images \cite{mnih2014recurrent, fu2017look, xie2020spatially, han2021dynamic, 9851927, 10091227}. By performing inference on these relatively small regions, one can dramatically reduce the computational cost of visual backbones (\emph{e.g.}, processing a 96x96 patch requires $\sim$18\% computation of inferring a 224x224 image).

% This paper investigates the potential for harnessing \emph{spatial redundancy} to enable efficient video recognition. We introduce a novel adaptive focus (AdaFocus) technique to dynamically identify and attend to task-relevant regions within each frame. Specifically, our approach initially employs a lightweight deep model to briefly examine each frame, obtaining cost-effective and coarse global information. Subsequently, we train a policy network based on this information to select the region with the highest value for recognition. The localization of task-relevant regions involves a non-differentiable process, necessitating the use of a reinforcement learning algorithm. Ultimately, we apply a high-capacity, accurate local encoder exclusively to the chosen regions. As the proposed regions typically consist of small patches with reduced dimensions, this approach results in significant computational savings. An illustration of AdaFocus can be found in Figure~\ref{fig:fig1} (c). By unevenly distributing computation across the spatial dimension of video frames according to their contributions to the recognition task, our method achieves a considerable increase in efficiency while maintaining accuracy.

In this paper, we are interested in whether this \emph{spatial redundancy} can be effectively leveraged to facilitate efficient video recognition.  We start by introducing a novel adaptive focus (AdaFocus) approach to dynamically localize and attend to the task-relevant regions of each frame. In specific, our method first takes a quick glance at each frame with a lightweight deep model to acquire cheap and coarse global information. Then we train a policy network on its basis to select the most valuable region for recognition. This procedure leverages the reinforcement learning algorithm due to the non-differentiability of localizing task-relevant regions. Finally, we activate a high-capacity and accurate local encoder to process only the selected regions. Since the proposed regions are usually small patches with a reduced size, considerable computational costs can be saved. An illustration of AdaFocus can be found in Figure~\ref{fig:fig1} (c). Our method allocates computation unevenly across the spatial dimension of video frames according to the contributions to the recognition task, leading to significant improvements in efficiency with a preserved accuracy.

% Building upon the foundational AdaFocus framework, we thoroughly investigate the optimal design of efficient spatial dynamic computation algorithms and enhance AdaFocus in several key aspects. Firstly, we simplify the training process of AdaFocus by reformulating it as an end-to-end algorithm, which eliminates the need for a complex three-stage training procedure with reinforcement learning. This results in reduced training costs, improved test accuracy, and greater accessibility for practitioners. Secondly, we explore the integration of appropriate supervision signals for learning to select task-relevant regions and propose a deep-feature-based approach for training more effective selection policies. Lastly, we introduce a deformable patch mechanism designed to flexibly adapt to task-relevant regions with varying shapes, sizes, scales, and locations.

On top of the vanilla AdaFocus framework, we delve deep into the optimal design of efficient spatial dynamic computation algorithms, and further improve AdaFocus in several important aspects. Firstly, we simplify the training of AdaFocus by reformulating it as an end-to-end algorithm, eliminating the need for the complicated three-stage training procedure with reinforcement learning. This yields reduced training cost, improved test accuracy, and greater accessibility for practitioners. Secondly, we present a discussion on how to introduce appropriate supervision signals for learning to select task-relevant regions, and propose a deep-feature-based approach for training more effective region selection policies. Lastly, we propose a deformable patch mechanism that enables AdaFocus to adapt flexibly to the task-relevant regions in various scales, shapes, and locations.

% It is important to highlight that the fundamental formulation of AdaFocus does not account for temporal-wise and sample-wise redundancies, meaning that computation is uniformly distributed across the temporal dimension and throughout different videos. However, our approach is amenable to temporal-adaptive and sample-adaptive dynamic inference techniques. For instance, AdaFocus can be augmented by concentrating computational resources on the most informative video frames or by decreasing computation allocated to relatively "easier" samples. In this paper, we illustrate the feasibility of achieving these objectives by incorporating a dynamic frame sampling algorithm and a conditional-exit mechanism.

It is worth noting that the basic formulation of AdaFocus does not account for the temporal-wise and sample-wise redundancies, meaning that the computation is uniformly allocated along the temporal dimension and across different videos. Therefore, our method is compatible with the ideas of temporal-adaptive and sample-adaptive dynamic inference. For instance, it can be extended by concentrating computational resources on the most informative video frames and by decreasing the computation spent on relatively ``easier'' samples. In this paper, we demonstrate that these goals can be attained by introducing a dynamic frame sampling algorithm as well as a conditional-exit mechanism.

% Incorporating the aforementioned methodological advancements, we present unified AdaFocus (Uni-AdaFocus, as depicted in Figure~\ref{fig:fig1} (d)), a comprehensive framework that cohesively integrates spatial, temporal, and sample-wise dynamic computation. Importantly, Uni-AdaFocus is compatible with a wide range of off-the-shelf backbone models, such as TSM \cite{lin2019tsm} and X3D \cite{feichtenhofer2020x3d}, which can be readily implemented as the feature extractor in Uni-AdaFocus to enhance their computational efficiency. Furthermore, Uni-AdaFocus's inference cost can be adjusted online without necessitating additional training (by modifying the criteria for sample-conditional computation). This adaptability allows Uni-AdaFocus to fully utilize varying computational resources or achieve a desired performance level with minimal power consumption, meeting the practical demands of numerous real-world applications, such as search engines and mobile applications.

Incorporating the aforementioned methodology innovations, we present unified AdaFocus (Uni-AdaFocus, see Figure~\ref{fig:fig1} (d)), a holistic framework that seamlessly integrates spatial, temporal, and sample-wise dynamic computation. Importantly, it is compatible with a wide range of off-the-shelf backbone models (\emph{e.g.}, TSM \cite{lin2019tsm} and X3D \cite{feichtenhofer2020x3d}), which can be conveniently deployed as the feature extractor in Uni-AdaFocus for improving their computational efficiency. Furthermore, the inference cost of Uni-AdaFocus can be adjusted online without additional training (by modifying the criterion for sample-conditional computation). This adaptability enables it to fully utilize fluctuating computational resources or achieve the desired level of performance flexibly with minimal power consumption, both of which are the practical demands of numerous real-world applications, such as search engines and mobile applications.

% reducing the computation spent on uninformative video clips or relatively `easier' samples.
% The vanilla AdaFocus framework does not model temporal redundancy, i.e., all frames are processed with identical computation, while the only difference lies in the locations of the selected regions. 
% We show that our method is compatible with existing temporal-based techniques, and can be extended via reducing the computation spent on uninformative frames, as presented in Figure~\ref{fig:fig1} (d). This is achieved by introducing an additional policy network that determines whether to skip some less valuable frames. This algorithm is referred to as AdaFocus+.

% Comprehensive experimental results demonstrate that Uni-AdaFocus consistently surpasses competing baselines by significant margins, attaining a new state-of-the-art performance in terms of both theoretical computational efficiency and practical inference speed.

Empirically, the effectiveness of Uni-AdaFocus is validated based on seven widely-used benchmark datasets and three real-world application scenarios. Extensive experiments demonstrate that Uni-AdaFocus consistently outperforms the competitive baselines by significant margins, attaining a new state-of-the-art performance in terms of both theoretical computational efficiency and practical inference speed.

% Empirically, the effectiveness of Uni-AdaFocus is validated based on seven widely-used benchmark datasets (\emph{i.e.}, ActivityNet, FCVID, Mini-Kinetics, Something-Something V1\&V2, Jester, and Kinetics-400) and three real-world application scenarios (\emph{i.e.}, fine-grained diving action classification, Alzheimer's and Parkinson's diseases diagnosis utilizing brain magnetic resonance images (MRI), and violence recognition for online videos). Extensive experimental results demonstrate that Uni-AdaFocus consistently outperforms the competitive baselines by significant margins, attaining a new state-of-the-art performance in terms of both theoretical computational efficiency and practical inference speed

% We evaluate the effectiveness of AdaFocus on five video recognition benchmarks (i.e., ActivityNet, FCVID, Mini-Kinetics, Something-Something V1\&V2). Experimental results show that AdaFocus by itself consistently outperforms all the baselines by large margins, while AdaFocus+ further improves the efficiency. For instance, AdaFocus+ has 2-3x less FLOPs\footnote{In this paper, FLOPs refers to the number of multiply-add operations.} than the recently proposed AR-Net \cite{meng2020ar} when achieving the same accuracy. We also demonstrate that our method can be deployed on top of the state-of-the-art networks (e.g., TSM \cite{lin2019tsm}) and effectively improve their computational efficiency.

This paper extends previous conference papers that introduced the basic AdaFocus framework \cite{Wang_2021_ICCV} and preliminarily discussed its end-to-end training \cite{wang2021adafocus}. Moreover, our deep-feature-based approach for training the patch selection policy (Section \ref{sec:improved_spatial_sub1}) is conceptually relevant to, and improved upon \cite{wang2022adafocusv3} (see Table \ref{tab:_discuss_vs_adaV3} for a detailed comparison). We have improved these earlier works substantially in several important aspects, which are summarized in Appendix \ref{app:summary_of_change}.

\section{Related Works}
\label{sec:related}
% \vspace{-0.4ex}

% Convolutional networks (ConvNets) have made a noteworthy impact on the field of large-scale video recognition, demonstrating exceptional accuracy in recent benchmark studies [1-5]. The methods employed within this field can be broadly categorized into several distinct approaches. One such method entails the concurrent capture of spatial and temporal data through the use of 3D Convolutional Networks (ConvNets), as demonstrated by works such as C3D [6], I3D [7], ResNet3D [8], and X3D [9]. An alternative technique involves the initial extraction of frame-wise features, followed by temporal data aggregation using specialized architectures. This approach can be seen in studies utilizing temporal averaging [10], the deployment of recurrent networks [11-13], and temporal channel shift [14-17]. A third category of work employs two-stream architectures to model short-term and long-term temporal relationships, as seen in [18-21]. The recent success of Vision Transformers (ViTs) [22] has inspired a significant number of studies to explore the potential of self-attention-based models in advancing effective video understanding [23-30].

\textbf{Video recognition.}
Convolutional networks (ConvNets) have made a noteworthy impact on the field of automatic video recognition, demonstrating exceptional accuracy on large-scale benchmarks \cite{caba2015activitynet,TPAMI-fcvid,kay2017kinetics,goyal2017something,materzynska2019jester}. The methods employed within this field can be broadly categorized into several distinct approaches. One such method entails the concurrent capture of spatial and temporal information through the use of 3D convolution, as demonstrated by the works such as C3D \cite{tran2015learning}, I3D \cite{carreira2017quo}, ResNet3D \cite{hara2018can}, X3D \cite{feichtenhofer2020x3d}, etc. An alternative technique involves the initial extraction of frame-wise features, followed by temporal-wise aggregation using specialized architectures. This approach can be seen in studies utilizing temporal averaging \cite{wang2016temporal}, recurrent networks \cite{donahue2015long, li2018recurrent, yue2015beyond}, and temporal channel shift \cite{lin2019tsm, sudhakaran2020gate, meng2021adafuse}. A third category of work employs two-stream architectures to model short-term and long-term temporal relationships, as seen in \cite{feichtenhofer2016convolutional, feichtenhofer2017spatiotemporal, feichtenhofer2019slowfast, gong2021searching}. More recently, driven by the success of vision Transformers (ViTs) \cite{dosovitskiy2021image}, a considerable number of works focus on facilitating effective video understanding with self-attention-based models \cite{arnab2021vivit, liu2022video, bertasius2021space, neimark2021video}. Notwithstanding the accomplishments of the aforementioned studies works, the expensive computational cost of deep networks, particularly 3D-ConvNets, usually constrain their practical application. Recent research efforts have been made towards improving the efficiency of video recognition \cite{tran2018closer, zolfaghari2018eco, tran2019video, liu2020teinet, liu2021tam, feichtenhofer2020x3d}.

\textbf{Temporal redundancy.}
A prominent strategy for facilitating efficient video recognition entails the minimization of the temporal redundancy within videos \cite{yeung2016end, wu2019adaframe, gao2020listen, korbar2019scsampler, wu2019multi, meng2020ar, ghodrati2021frameexit, kim2021efficient, sun2021dynamic, lin2022ocsampler, xia2022nsnet, xia2022temporal}. Since not all frames are equally important for a given task, the model should ideally allocate fewer computational resources towards less informative frames \cite{han2021dynamic}. Several effective algorithms have been proposed along this direction. For example, LiteEval \cite{wu2019liteeval} adaptively selects an LSTM model with appropriate size at each time step in a recurrent recognition procedure. Adaptive resolution network (AR-Net) \cite{meng2020ar} processes different frames with adaptive resolutions to save unnecessary computation on less important frames. VideoIQ \cite{sun2021dynamic} processes video frames using different precision according to their relative importance. FrameExit \cite{ghodrati2021frameexit} learns to conclude the inference process after seeing a few sufficiently informative frames. Compared to these approaches, the contributions of this paper lie in that 1) we develop the methodologies of reducing \emph{spatial redundancy} (AdaFocus), that is, to concentrate major computation on the task-relevant regions of video frames; 2) we demonstrate that AdaFocus is compatible with the spirit of reducing temporal redundancy by proposing a dynamic frame sampling algorithm tailored for our method; 3) we integrate the spatial, temporal, and sample-wise dynamic computation into a Uni-AdaFocus framework, yielding state-of-the-art computational efficiency.

In particular, OCSampler \cite{lin2022ocsampler} proposes a novel and effective framework that learns to select task-relevant frames with reinforcement learning. Our work is related to \cite{lin2022ocsampler} in the basic paradigm of formulating frame selection as a sequential weighted sampling problem without replacement, where the distribution is dynamically parameterized conditioned on each video utilizing a policy network. However, we develop novel theoretical analyses, which directly consider the expected loss of this problem as an optimization objective, and reveal that it can be decomposed into a differentiable form solved by the Monte Carlo method, yielding an efficient end-to-end trainable algorithm (Section \ref{sec:frame_sample}). Compared to \cite{lin2022ocsampler}, our method does not rely on reinforcement learning or multi-stage training, considerably reduces both the theoretical complexity and the practical training wall-time, yet significantly improves the performance (Table \ref{tab:_discuss_vs_oc}).

\textbf{Spatial-wise dynamic networks}
perform computation adaptively on top of different spatial locations of the inputs \cite{han2021dynamic, jaderberg2015spatial}. The AdaFocus network studied in this paper can be classified into this category as well. Many of the spatially adaptive networks are designed from the lens of inference efficiency \cite{han2021dynamic, ren2018sbnet, yang2020resolution, wang2019adaptively, chen2021dynamic}. For example, recent investigations have revealed that 2D images can be efficiently processed via attending to the task-relevant or more information-rich image regions \cite{figurnov2017spatially, 9851927, xie2020spatially, verelst2020dynamic}. In the realm of video understanding, the exploitation of spatial redundancy as a means to reduce computational cost remains a relatively unexplored area. It has been shown by the attention-based methods \cite{meng2019interpretable, li2021generalized} that the contributions of different frame regions to the recognition task are not equivalent. Some preliminary studies \cite{Wang_2021_ICCV, wang2021adafocus} have begun to underscore the potential benefits of this approach.

% In particular, the Spatial Transformer Networks [1] employ an interpolation-based mechanism, akin to the differentiable patch selection technique utilized in AdaFocus. However, their primary focus lies in proactively transforming the feature maps to learn spatially invariant representations, whereas our objective is to localize and concentrate on the task-relevant regions of video frames to enhance inference efficiency. Furthermore, we demonstrate that a direct implementation of this mechanism does not produce competitive outcomes in our problem context. To overcome the optimization challenges, our algorithm necessitates the introduction of improved designs.

The spatial transformer networks \cite{jaderberg2015spatial} are trained based on an interpolation-based mechanism, which is similar to the differentiable patch selection technique in AdaFocus. However, they focus on actively transforming the feature maps for learning spatially invariant representations, whereas our objective is to localize and attend to the task-relevant regions of the video frames for improving the computational efficiency. Moreover, we demonstrate that a straightforward implementation of this mechanism fails to yield competitive results in our problem. To address the optimization challenges, our algorithm necessitates the introduction of improved designs, as discussed in this paper.

% SePiCo: Semantic- Guided Pixel Contrast for Domain Adaptive Semantic Segmentation
% Generalized Domain Conditioned Adaptation Network
% Deep Residual Correction Network for Partial Domain Adaptation
% Adapting Across Domains via Target-Oriented Transferable Semantic Augmentation Under Prototype Constraint

\begin{figure}[!h]
    \vskip -0.1in
    \begin{center}
    \centerline{\includegraphics[width=0.9\columnwidth]{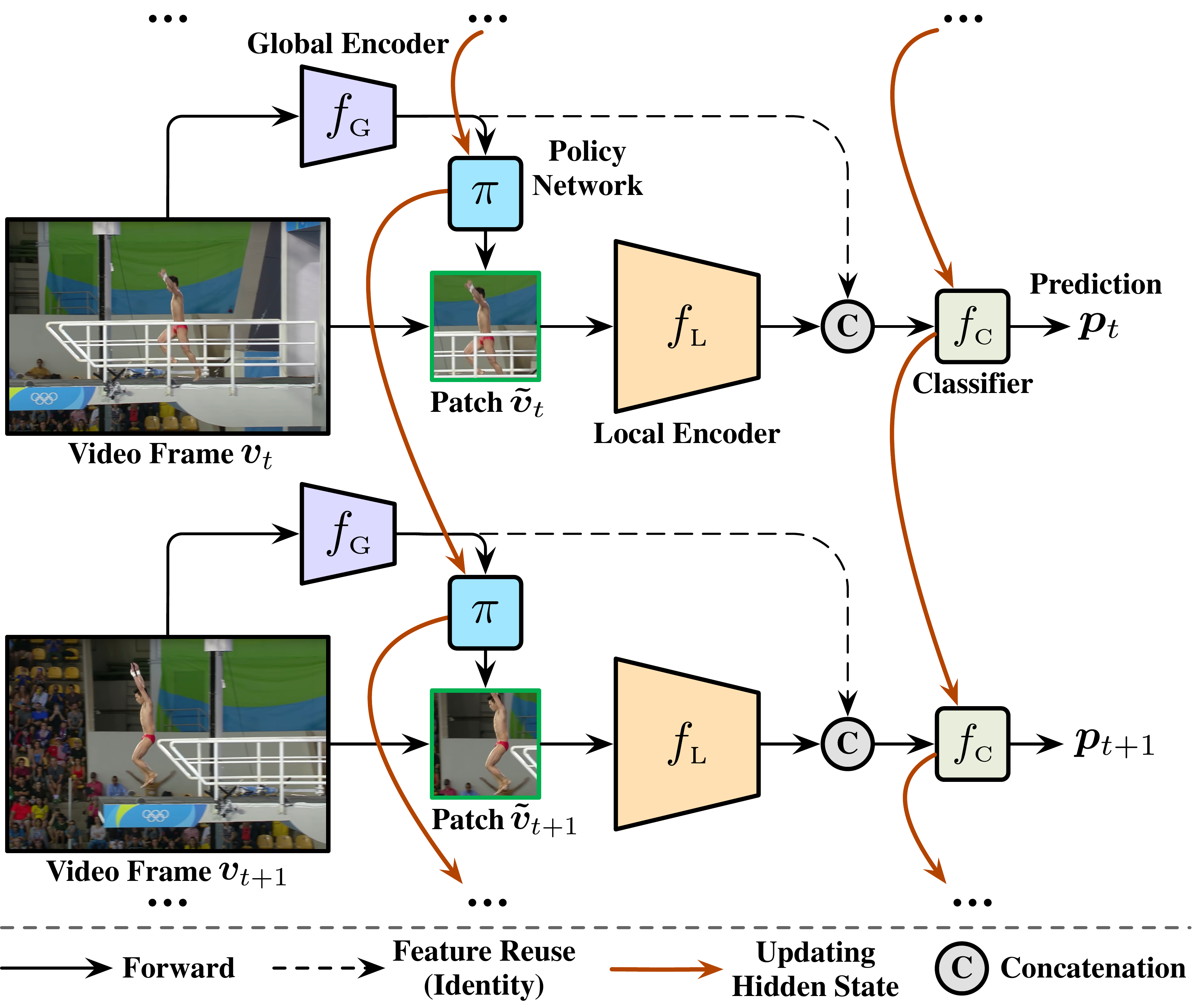}}
    \vskip -0.125in
    \caption{\textbf{Overview of AdaFocus}. It first takes a quick glance at each frame $\bm{v}_t$ using a lightweight global encoder $f_{\textnormal{G}}$. Then a policy network $\pi$ is built on top of $f_{\textnormal{G}}$ to select the most important image region $\tilde{\bm{v}}_t$ in terms of recognition. A high-capacity local encoder $f_{\textnormal{L}}$ is adopted to extract features from $\tilde{\bm{v}}_t$. Finally, a classifier aggregates the features across frames to obtain the prediction $\bm{p}_t$. \label{fig:overview}
    }
    \end{center}
    \vspace{-4ex}
\end{figure}

% \vspace{-1ex}
\section{Adaptive Focus Network (AdaFocus)}
% \vspace{-1ex}
\label{sec:method}

Different from most existing works that facilitate efficient video recognition by leveraging the \emph{temporal redundancy}, we seek to save the computation spent on the task-irrelevant regions of video frames, and thus improve the efficiency by reducing the \emph{spatial redundancy}. To attain this goal, we propose an AdaFocus framework to adaptively identify and attend to the most informative regions of each frame, such that the computational cost can be significantly reduced without sacrificing accuracy. In this section, we first introduce the basic formulation of AdaFocus and its network architecture (Section \ref{sec:arch}). Then we show that the straightforward optimization problem derived from this basic formulation can be solved by a reinforcement-learning-based three-stage algorithm (AdaFocusV1, Section \ref{sec:AdaFocusV1}). Built upon these discussions, we further establish the feasibility of reformulating the training of AdaFocus into an end-to-end algorithm, which consistently improves the accuracy with a simpler and more efficient training process (AdaFocusV2, Section \ref{sec:AdaFocusV2}).

\subsection{Network Architecture}
\label{sec:arch}

\textbf{Overview.}
We start by giving an overview of AdaFocus (Figure \ref{fig:overview}). Without loss of generality, we consider an online video recognition scenario, where a stream of frames come in sequentially while a prediction may be retrieved after processing any number of frames. At each time step, AdaFocus first takes a quick glance at the full frame with a lightweight deep network $f_{\textnormal{G}}$, obtaining cheap and coarse global features. Then the features are fed into a policy network $\pi$ to aggregate the information across frames and accordingly determine the location of an image patch to be focused on, under the goal of maximizing its contribution to video recognition. A high-capacity local encoder $f_{\textnormal{L}}$ is then adopted to process the selected patch for more accurate but computationally expensive representations. Finally, a classifier $f_{\textnormal{C}}$ integrates the features of all previous frames to produce a prediction. In the following, we describe the four components of our method in details.

\textbf{Global encoder $f_{\textnormal{G}}$ and local encoder $f_{\textnormal{L}}$}
are both backbone networks that extract deep features from the inputs, but with distinct aims. The former is designed to quickly catch a glimpse of each frame, providing necessary information for determining which region the local encoder $f_{\textnormal{L}}$ should attend to. Therefore, a lightweight network is adopted for $f_{\textnormal{G}}$. On the contrary, $f_{\textnormal{L}}$ is leveraged to take full advantage of the selected image regions for learning discriminative representations, and hence we deploy large and accurate models. Since $f_{\textnormal{L}}$ only needs to process a series of relatively small regions instead of the full images, this stage enjoys high efficiency as well. Importantly, the formulation of $f_{\textnormal{G}}$ and $f_{\textnormal{L}}$ is general and flexible, \emph{i.e.}, most state-of-the-art deep learning models can be conveniently deployed in AdaFocus to improve their computational efficiency for inference. Representative examples are given in Section \ref{sec:experiment}.

% We defer the details on the architectures of $f_{\textnormal{G}}$ and $f_{\textnormal{L}}$ to Section \ref{sec:experiment}.

Formally, given video frames $\{\bm{v}_1, \bm{v}_2, \ldots\}$ with size $H\!\times\!W$, $f_{\textnormal{G}}$ directly takes them as inputs and produces the coarse global feature maps $\bm{e}^{\textnormal{G}}_{t}$: 
\begin{equation}
    \bm{e}^{\textnormal{G}}_{t} = f_{\textnormal{G}}(\bm{v}_t),\ \ \  t=1,2,\ldots,
\end{equation}
where $t$ is the frame index. By contrast, $f_{\textnormal{L}}$ processes $P\!\times\!P$ ($P<H, W$) square image patches $\{\tilde{\bm{v}}_1, \tilde{\bm{v}}_2, \ldots\}$, which are cropped from $\{\bm{v}_1, \bm{v}_2, \ldots\}$ respectively, and we have 
\begin{equation}
    \bm{e}^{\textnormal{L}}_{t} = f_{\textnormal{L}}(\tilde{\bm{v}}_t),\ \ \  t=1,2,\ldots,
\end{equation}
where $\bm{e}^{\textnormal{L}}_{t}$ denotes the fine local feature maps. Importantly, the patch $\tilde{\bm{v}}_t$ is localized to capture the most informative regions for the given task, and this procedure is fulfilled by the policy network $\pi$, which is introduced in the following.

\textbf{Policy network $\pi$}
specifies which region the local encoder $f_{\textnormal{L}}$ should attend to for each frame, \emph{i.e.}, the locations of $\{\tilde{\bm{v}}_1, \tilde{\bm{v}}_2, \ldots\}$. This goal is attained by leveraging the coarse global features $\bm{e}^{\textnormal{G}}_{t}$ from the global encoder $f_{\textnormal{G}}$. Note that the features of both previous and current frames can be used, and hence $\pi$ should be designed as the architecture capable of encoding temporal information (\emph{e.g.}, via incorporating recurrent networks, 3D convolution or self-attention modules). The detailed formulations and training algorithms related to $\pi$ will be further discussed in Sections \ref{sec:AdaFocusV1} and \ref{sec:AdaFocusV2}.

% Formally, $\pi$ determines the locations of images patches $\{\tilde{\bm{v}}_1, \tilde{\bm{v}}_2, \ldots\}$ to be cropped from the frames. Given that this leads to a non-differentiable operation, we formalize $\pi$ as an agent and train it with reinforcement learning. In specific, the location of the patch $\tilde{\bm{v}}_t$ is drawn from the distribution:
% \begin{equation}
%     \label{eq:select_location}
%     \tilde{\bm{v}}_t \sim \pi(\cdot|\bm{e}^{\textnormal{G}}_{t}, \bm{h}^{\pi}_{t-1}),
% \end{equation}
% where $\bm{h}^{\pi}_{t-1}$ denotes the hidden states maintained in $\pi$ that are updated at $(t-1)^{\textnormal{th}}$ frame. In our implementation, we consider multiple candidates (e.g., 36 or 49) uniformly distributed across the images, and establish a categorical distribution on them, which is parameterized by the outputs of $\pi$. At test time, we simply adopt the candidate with maximum probability as $\tilde{\bm{v}}_t$ for a deterministic inference procedure. In addition, note that we do not perform any pooling on the features maps $\bm{e}^{\textnormal{G}}_{t}$ since pooling typically corrupts the useful spatial information for localizing $\tilde{\bm{v}}_t$. As an alternative, we compress the number of channels with $1\times1$ convolution to reduce the computational cost of $\pi$. An illustration of $\pi$ is shown in Figure \ref{fig:policy_net}.

\textbf{Classifier $f_{\textnormal{C}}$}
is a prediction network aiming to aggregate the information from all the frames that have been processed by the model, and output the current recognition result at each time step. To be specific, we perform global average pooling on the feature maps $\bm{e}^{\textnormal{G}}_{t}, \bm{e}^{\textnormal{L}}_{t}$ from the two aforementioned encoders to get feature vectors $\overline{\bm{e}}^{\textnormal{G}}_{t}, \overline{\bm{e}}^{\textnormal{L}}_{t}$, and concatenate them as the inputs of $f_{\textnormal{C}}$, namely
\begin{equation}
    \label{eq:classifier}
    \bm{p}_t = f_{\textnormal{C}}([\overline{\bm{e}}^{\textnormal{G}}_{1}, \overline{\bm{e}}^{\textnormal{L}}_{1}],\ldots,[\overline{\bm{e}}^{\textnormal{G}}_{t}, \overline{\bm{e}}^{\textnormal{L}}_{t}]),
\end{equation}
where $\bm{p}_t$ refers to the softmax prediction at $t^{\textnormal{th}}$ step. It is noteworthy that  $\bm{e}^{\textnormal{G}}_{t}$ is leveraged for both localizing the informative patches and recognition, under the goal of facilitating efficient feature reuse. This design is natural since it has been observed that deep networks (\emph{e.g.}, ConvNets and Vision Transformers) excel at learning representations for both recognition and localization simultaneously \cite{zhou2016learning, selvaraju2017grad, dosovitskiy2021an}. Many existing methods also adopt similar reusing mechanisms \cite{wu2019liteeval, wu2019adaframe, meng2020ar, gao2020listen, liang2022delving}. In addition, the architecture of $f_{\textnormal{C}}$ may have different choices, such as recurrent networks \cite{hochreiter1997long, cho-etal-2014-learning}, averaging the frame-wise predictions \cite{lin2019tsm, meng2020ar, meng2021adafuse}, and accumulated feature pooling \cite{ghodrati2021frameexit}.

% It is worth noting that we allow $\bm{e}^{\textnormal{G}}_{t}$ to be utilized for classification as well, with the aim of facilitating more efficient feature reusing. Such a design leverages previous observations \cite{zhou2016learning, selvaraju2017grad} revealing that deep networks are capable of achieving both remarkable localization and recognition performance at the same time. Many of existing methods also adopt similar reusing mechanisms \cite{wu2019liteeval, wu2019adaframe, meng2020ar, gao2020listen}. Besides, there are multiple possible architectures for $f_{\textnormal{C}}$. In addition to the choice of recurrent networks such as long short-term memory (LSTM) \cite{hochreiter1997long} or gated recurrent unit (GRU) \cite{cho-etal-2014-learning}, $f_{\textnormal{C}}$ can also be set as taking the average of frame-wise predictions, which are typically obtained with a common fully-connected layer, as done in \cite{lin2019tsm, meng2020ar, meng2021adafuse}. 

\subsection{AdaFocusV1: Three-stage Reinforcement Learning}
\label{sec:AdaFocusV1}

\textbf{Patch localization as a sequential discrete decision task.}
As aforementioned, the policy network $\pi$ localizes the task-relevant patches $\{\tilde{\bm{v}}_1, \tilde{\bm{v}}_2, \ldots\}$, which are cropped from video frames and processed by the local encoder $f_{\textnormal{L}}$. However, the cropping operation is inherently non-differentiable. To address this issue, a straightforward approach is to formulate $\pi$ as an agent that makes a series of discrete decisions, such that $\pi$ can be trained with reinforcement learning. 

% In this subsection, we demonstrate how this idea lead to an effective training algorithm for AdaFocus.

\begin{figure}[t]
    % \vskip -0.1in
    \begin{center}
    \centerline{\includegraphics[width=0.9\columnwidth]{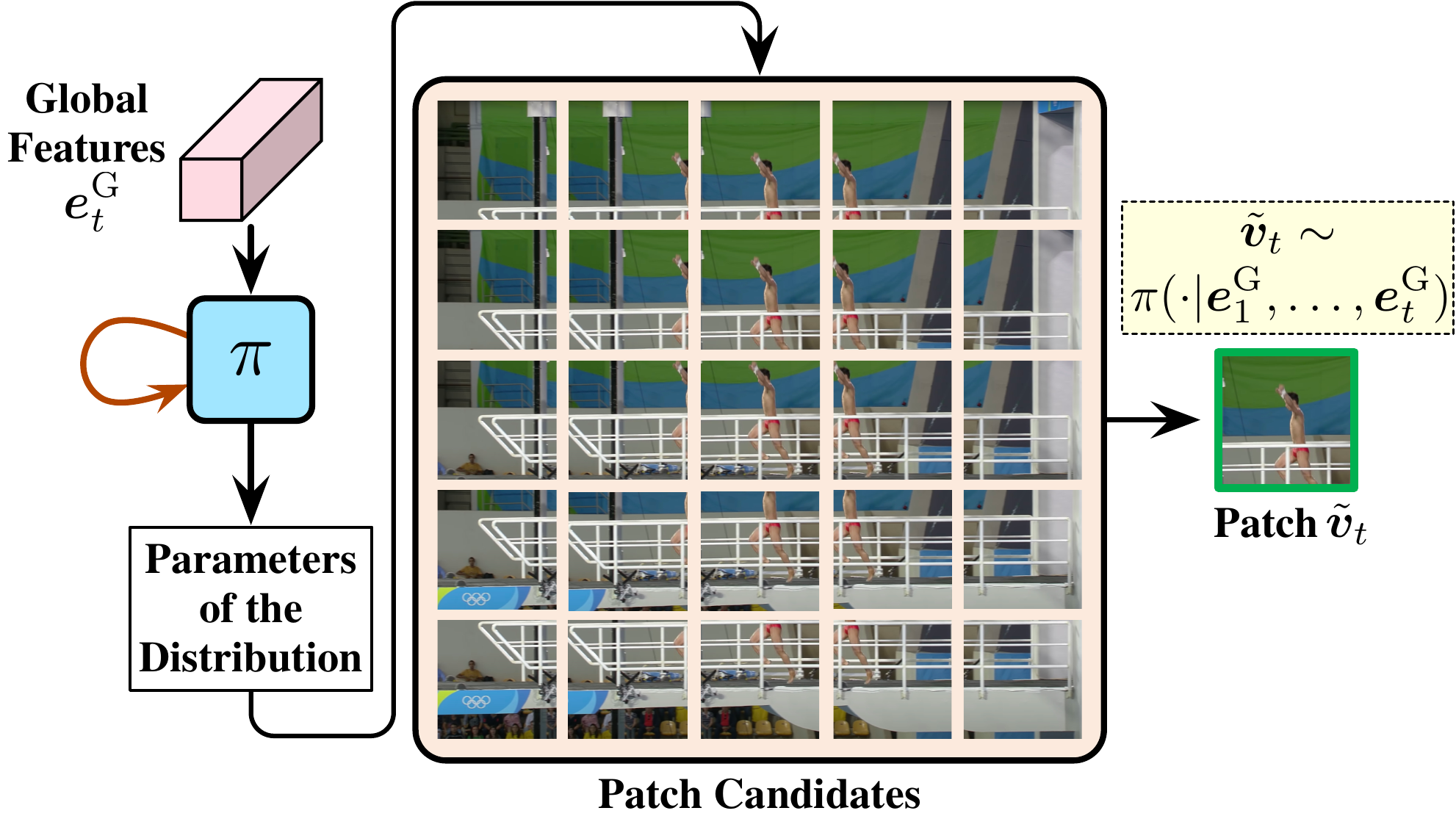}}
    \vskip -0.15in
    \caption{\textbf{Illustration of the policy network $\pi$ in AdaFocusV1.} The outputs of $\pi$ parameterize a categorical distribution $\pi(\cdot|{\bm{e}}^{\textnormal{G}}_{1},\ldots, {\bm{e}}^{\textnormal{G}}_{t})$ on multiple patch candidates (here we take 25 as an example). During training, we sample $\tilde{\bm{v}}_t$ from $\pi(\cdot|{\bm{e}}^{\textnormal{G}}_{1},\ldots, {\bm{e}}^{\textnormal{G}}_{t})$, while at test time, we directly select the patch with the largest softmax probability.  \label{fig:policy_net}
    }
    \end{center}
    \vspace{-3ex}
\end{figure}

As a basic assumption, we suppose that the location of the patch $\tilde{\bm{v}}_t$ is drawn from an action distribution parameterized by the outputs of $\pi$:
\begin{equation}
    \label{eq:select_location}
    \tilde{\bm{v}}_t \sim \pi(\cdot|{\bm{e}}^{\textnormal{G}}_{1},\ldots, {\bm{e}}^{\textnormal{G}}_{t}).
\end{equation}
Note that we do not perform any pooling on the features maps $\bm{e}^{\textnormal{G}}_{t}$ since pooling typically corrupts the useful spatial information for localizing $\tilde{\bm{v}}_t$. In our implementation, we consider multiple candidates (e.g., 36 or 49) uniformly distributed across the images, and establish a categorical distribution on them. At test time, we simply adopt the candidate with maximum probability as $\tilde{\bm{v}}_t$ for a deterministic inference procedure. An illustration is shown in Figure \ref{fig:policy_net}.

\textbf{Three-stage training.}
Since the formulation above includes both continuous (\emph{i.e.}, video recognition) and discrete (\emph{i.e.}, patch localization) optimization, the standard end-to-end training paradigm cannot be directly applied. Therefore, we introduce a three-stage training algorithm to solve the continuous and discrete optimization problems alternatively, which can be found in Appendix \ref{app:three_stage}, due to spatial limitations.

\subsection{AdaFocusV2: Differentiable End-to-end Training}
\label{sec:AdaFocusV2}

\textbf{Limitations of AdaFocusV1.}
The underlying logic behind the aforementioned three-stage training is straightforward. However, this procedure is unfriendly for practitioners. First, effectively deploying the reinforcement learning algorithm is nontrivial. It requires considerable efforts for properly designing the key components (\emph{e.g.}, the action space and the reward function), and implementing specialized optimization techniques (\emph{e.g.}, deep Q-Network \cite{mnih2013playing} or proximal policy optimization \cite{schulman2017proximal}). Second, the three-stage alternative algorithm is an indirect formulation for optimizing the recognition objective, which tends to be time-consuming, and may result in sub-optimal solutions. Third, the performance of AdaFocusV1 largely depends on a number of implementation configurations (\emph{e.g.}, performing pre-training, freezing some components in different stages, and stage-wise hyper-parameter searching) that need to be carefully tuned on a per-dataset or per-backbone basis.

In the following, we present an end-to-end trainable formulation for AdaFocus to address the issue of inefficient training. The proposed network, AdaFocusV2, can be conveniently implemented to achieve consistently better performance than AdaFocusV1 with reduced training cost. A comparison of AdaFocusV1 and V2 is given in Table \ref{tab:v1vsv2}.

\begin{table}[!h]
    \centering
    \begin{footnotesize}
    \vskip -0.1in
    \caption{\textbf{A comparison of training AdaFocusV1 (from \ding{182} to \ding{186}) and AdaFocusV2 (end-to-end).} Both procedures start from the same initial backbone network. Note that $f_{\textnormal{G}}$, $f_{\textnormal{L}}$, $f_{\textnormal{C}}$ and $\pi$ are the model components.}
    \vskip -0.22in
    \setlength{\tabcolsep}{3mm}{
    \vspace{5pt}
    \renewcommand\arraystretch{0.95}
    \resizebox{0.95\columnwidth}{!}{
    \begin{tabular}{cl|c}
    \toprule
     \multicolumn{2}{c|}{AdaFocusV1} & AdaFocusV2 \\
     \midrule
    %  \midrule
     \textbf{\emph{Proper}} & \ding{182} Pre-train $f_{\textnormal{G}}$ on the target dataset.& \multirow{8.5}{*}{\shortstack{\textcolor{blue}{{\textbf{\emph{End-to-end}}}}\\ \textbf{\emph{\textcolor{blue}{Training}}}\\ \textcolor{black}{{\emph{($f_{\textnormal{G}}$, $f_{\textnormal{L}}$, $f_{\textnormal{C}}$, $\pi$)}}}}} \\
     \textbf{\emph{Initialization}} & \ding{183} Pre-train $f_{\textnormal{L}}$ on the target dataset.& \\
    %  \cline{1-2}
    \cmidrule{1-2}
     \multirow{2}{*}{\textbf{\emph{Stage I}}} & \ding{184} Train $f_{\textnormal{G}}$, $f_{\textnormal{L}}$ and $f_{\textnormal{C}}$ using & \\
      &random patches.& \\
      \cmidrule{1-2}
      \multirow{2}{*}{\textbf{\emph{Stage II}}} & \ding{185} Train $\pi$ using & \\
      &reinforcement learning.& \\
      \cmidrule{1-2}
      {\textbf{\emph{Stage III}}} & \ding{186} Fine-tune $f_{\textnormal{L}}$ and $f_{\textnormal{C}}$. & \\
    \bottomrule
    \label{tab:v1vsv2}
    \end{tabular}}}
    \end{footnotesize}
    \vspace{-3.5ex}
\end{table}

\subsubsection{Interpolation-based Patch Selection}
\label{sec:interpolation}

% \footnotetext{In our implementation, the height/width/coordinates are correspondingly normalized using the linear projection $[0,H]\!\to\![0,1]$ and $[0,W]\!\to\![0,1]$. Here we use the original values for the ease of understanding.}
To enable end-to-end training, we propose a differentiable solution to obtain $\tilde{\bm{v}}_t$. Suppose that the size of the original frame ${\bm{v}}_t$ and the patch $\tilde{\bm{v}}_t$ is $H\!\times\!W$ and $P\!\times\!P\ (P\!<\!H, W)$, respectively. We assume that $\pi$ outputs the continuous centre coordinates $(\tilde{x}^t_{\textnormal{c}}, \tilde{y}^t_{\textnormal{c}})$ of $\tilde{\bm{v}}_t$, namely
\begin{equation}
    \label{eq:cetre_xy}
    \begin{split}
        &(\tilde{x}^t_{\textnormal{c}},\ \tilde{y}^t_{\textnormal{c}}) = \pi({\bm{e}}^{\textnormal{G}}_{1},\ldots, {\bm{e}}^{\textnormal{G}}_{t}), \\
        \tilde{x}^t_{\textnormal{c}} \in [&\frac{P}{2}, W-\frac{P}{2}],\ \ \ \tilde{y}^t_{\textnormal{c}} \in [\frac{P}{2}, H-\frac{P}{2}],
    \end{split}
\end{equation}
where ${\bm{e}}^{\textnormal{G}}_{1},\ldots, {\bm{e}}^{\textnormal{G}}_{t}$ are the global features of $1^{\textnormal{th}}-t^{\textnormal{th}}$ frames extracted by the global encoder $f_{\textnormal{G}}$. Notably, we refer to the coordinates of the top-left corner of the frame as $(0,0)$, and Eq. (\ref{eq:cetre_xy}) ensures that $\tilde{\bm{v}}_t$ will never go outside of ${\bm{v}}_t$. Our aim is to calculate the values of all pixels in $\tilde{\bm{v}}_t$, while allowing the gradients to be back-propagated through $(\tilde{x}^t_{\textnormal{c}}, \tilde{y}^t_{\textnormal{c}})$. 

\textbf{Feed-forward.}
We first introduce the feed-forward process of our method. Formally, the coordinates of a pixel in the patch $\tilde{\bm{v}}_t$ can be expressed as the addition of $(\tilde{x}^t_{\textnormal{c}}, \tilde{y}^t_{\textnormal{c}})$ and a fixed offset:
\begin{equation}
    \label{eq:c_plus_offset}
    \begin{split}
        &(\tilde{x}^t_{ij},\ \tilde{y}^t_{ij}) = (\tilde{x}^t_{\textnormal{c}},\ \tilde{y}^t_{\textnormal{c}}) + \bm{o}_{ij}, \\
        \bm{o}_{ij} &\in {\left\{ -\frac{P}{2}, -\frac{P}{2} + 1, \ldots, \frac{P}{2} \right\}}^2.
    \end{split}
\end{equation}
Herein, $(\tilde{x}^t_{ij}, \tilde{y}^t_{ij})$ denotes the horizontal and vertical coordinates in the original frame ${\bm{v}}_t$ corresponding to the pixel in the $i^{\textnormal{th}}$ row and $j^{\textnormal{th}}$ column of $\tilde{\bm{v}}_t$, while $\bm{o}_{ij}$ represents the vector from the patch centre $(\tilde{x}^t_{\textnormal{c}}, \tilde{y}^t_{\textnormal{c}})$ to $(\tilde{x}^t_{ij}, \tilde{y}^t_{ij})$. Given a fixed patch size, $\bm{o}_{ij}$ is a constant conditioned only on $i,j$, regardless of $t$ or the inputs of $\pi$.

\begin{figure}[!t]
    % \vskip -0.1in
    \begin{center}
    \centerline{\includegraphics[width=0.9\columnwidth]{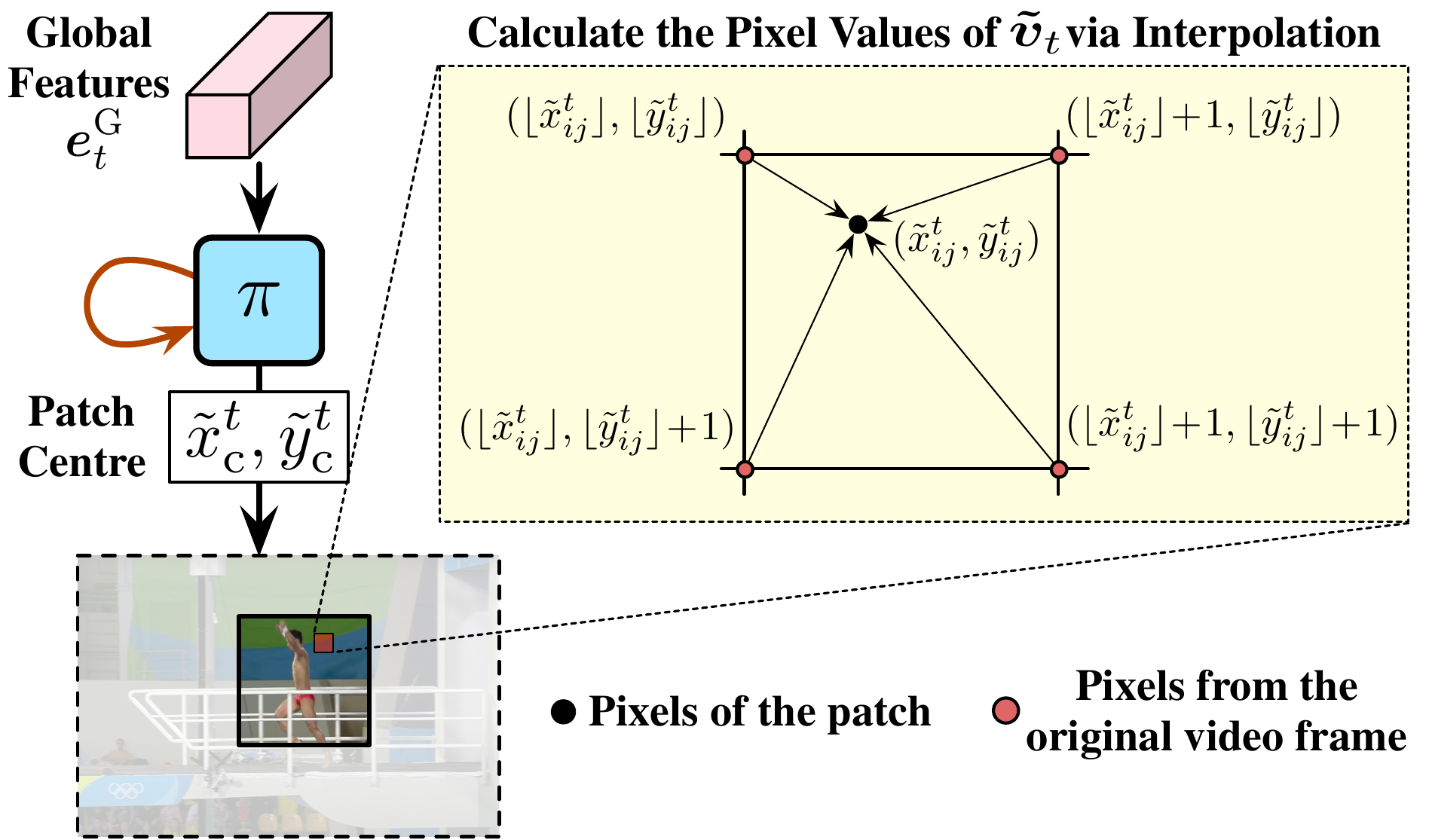}}
    \vskip -0.125in
    \caption{\textbf{Illustration of interpolation-based patch selection.} This operation is differentiable, \emph{i.e.}, the gradients can be directly back-propagated into the policy network $\pi$ through the selected image patch $\tilde{\bm{v}}_t$. Consequently, integrating the learning of $\pi$ into a unified end-to-end training paradigm turns out to be feasible. \label{fig:interpolation}
    }
    \end{center}
    \vspace{-2.5ex}
\end{figure}

Since the values of $(\tilde{x}^t_{\textnormal{c}}, \tilde{y}^t_{\textnormal{c}})$ are continuous, there does not exist a pixel of ${\bm{v}}_t$ exactly located at $(\tilde{x}^t_{ij}, \tilde{y}^t_{ij})$ to directly get the pixel value. Alternatively, as illustrated in Figure \ref{fig:interpolation}, we can always find that the location $(\tilde{x}^t_{ij}, \tilde{y}^t_{ij})$ is surrounded by four adjacent pixels of ${\bm{v}}_t$, forming a grid. The coordinates are $(\lfloor\tilde{x}^t_{ij}\rfloor, \lfloor\tilde{y}^t_{ij}\rfloor)$,  $(\lfloor\tilde{x}^t_{ij}\rfloor\!+\!1, \lfloor\tilde{y}^t_{ij}\rfloor)$,  $(\lfloor\tilde{x}^t_{ij}\rfloor, \lfloor\tilde{y}^t_{ij}\rfloor\!+\!1)$ and $(\lfloor\tilde{x}^t_{ij}\rfloor\!+\!1, \lfloor\tilde{y}^t_{ij}\rfloor\!+\!1)$, respectively, where $\lfloor\cdot\rfloor$ denotes the rounding-down operation. By assuming that the corresponding pixel values of these four pixels are $(m^t_{ij})_{00}$, $(m^t_{ij})_{01}$, $(m^t_{ij})_{10}$, and $(m^t_{ij})_{11}$, the pixel value at $(\tilde{x}^t_{ij}, \tilde{y}^t_{ij})$ (referred to as $\tilde{m}^t_{ij}$) can be obtained via interpolation algorithms. For example, we may simply adopt the differentiable bilinear interpolation:
\begin{equation}
    \label{eq:bilinear}
    \begin{split}
        \tilde{m}^t_{ij} &= (m^t_{ij})_{00}(\lfloor\tilde{x}^t_{ij}\rfloor\!-\!\tilde{x}^t_{ij}\!+\!1)(\lfloor\tilde{y}^t_{ij}\rfloor\!-\!\tilde{y}^t_{ij}\!+\!1) \\
        &+(m^t_{ij})_{01}(\tilde{x}^t_{ij}\!-\!\lfloor\tilde{x}^t_{ij}\rfloor)(\lfloor\tilde{y}^t_{ij}\rfloor\!-\!\tilde{y}^t_{ij}\!+\!1) \\
        &+(m^t_{ij})_{10}(\lfloor\tilde{x}^t_{ij}\rfloor\!-\!\tilde{x}^t_{ij}\!+\!1)(\tilde{y}^t_{ij}\!-\!\lfloor\tilde{y}^t_{ij}\rfloor) \\
        &+(m^t_{ij})_{11}(\tilde{x}^t_{ij}\!-\!\lfloor\tilde{x}^t_{ij}\rfloor)(\tilde{y}^t_{ij}\!-\!\lfloor\tilde{y}^t_{ij}\rfloor).
    \end{split}
\end{equation}
Consequently, we can obtain the image patch $\tilde{\bm{v}}_t$ by traversing all possible $i,j$ with Eq. (\ref{eq:bilinear}).

\textbf{Back-propagation.}
Give the training loss $\mathcal{L}$, it is easy to compute the gradient ${\partial\mathcal{L}}/{\partial\tilde{m}^t_{ij}}$ with standard back-propagation. Then, following on the chain rule, we have 
\begin{equation}
    % \begin{split}
        \label{eq:bp_1}
        \frac{\partial\mathcal{L}}{\partial\tilde{x}^t_{\textnormal{c}}} \!=\!\! \sum_{i,j}\! \frac{\partial\mathcal{L}}{\partial\tilde{m}^t_{ij}}
        \frac{\partial\tilde{m}^t_{ij}}{\partial\tilde{x}^t_{\textnormal{c}}}, \ \ \ 
        \frac{\partial\mathcal{L}}{\partial\tilde{y}^t_{\textnormal{c}}} \!=\!\! \sum_{i,j}\! \frac{\partial\mathcal{L}}{\partial\tilde{m}^t_{ij}}
        \frac{\partial\tilde{m}^t_{ij}}{\partial\tilde{y}^t_{\textnormal{c}}}.
    % \end{split}
\end{equation}
Combining Eq. (\ref{eq:c_plus_offset}) and Eq. (\ref{eq:bp_1}), we can further derive
\begin{equation}
    \label{eq:bp_2}
    \frac{\partial\tilde{m}^t_{ij}}{\partial\tilde{x}^t_{\textnormal{c}}}\!=\!\frac{\partial\tilde{m}^t_{ij}}{\partial\tilde{x}^t_{ij}},\ \ \ 
    \frac{\partial\tilde{m}^t_{ij}}{\partial\tilde{y}^t_{\textnormal{c}}}\!=\!\frac{\partial\tilde{m}^t_{ij}}{\partial\tilde{y}^t_{ij}}.
\end{equation}
Eq. (\ref{eq:bp_2}) can be solved by leveraging Eq. (\ref{eq:bilinear}), such that we can obtain the gradients ${\partial\mathcal{L}}/{\partial\tilde{x}^t_{\textnormal{c}}}$ and ${\partial\mathcal{L}}/{\partial\tilde{y}^t_{\textnormal{c}}}$. Given that $\tilde{x}^t_{\textnormal{c}}$ and $\tilde{y}^t_{\textnormal{c}}$ are the outputs of the policy network $\pi$, the regular back-propagation process is able to proceed.

% the back-propagation process is able to proceed in a regular way.

\subsubsection{Training Techniques}
\label{sec:AdaFocusV2_techs}

\textbf{Naive implementation.}
Thus far, we have enabled the gradients to be back-propagated throughout the whole AdaFocus network for updating all trainable parameters simultaneously. Consequently, end-to-end training has been feasible. For example, one can minimize the frame-wise cross-entropy loss $L_{\textnormal{CE}}(\cdot)$ in AdaFocusV1:
\begin{equation}
    \label{eq:naive_objective}
        \mathop{\textnormal{minimize}}_{f_{\textnormal{G}}, f_{\textnormal{L}}, f_{\textnormal{C}}, \pi}\ \ \mathcal{L}={\mathbb{E}}_{\{\bm{v}_1, \bm{v}_2, \ldots\}}
    \left[
        \frac{1}{T}\sum\nolimits_{t=1}^{T} L_{\textnormal{CE}}(\bm{p}_t, y)
    \right], 
\end{equation}
where $T$ and $y$ denote the length and the label of the video $\{\bm{v}_1, \bm{v}_2, \ldots\}$, and $\bm{p}_t$ is the softmax prediction at $t^{\textnormal{th}}$ frame.

% , as stated in Section \ref{sec:arch}.

However, importantly, such a straightforward implementation leads to the severely degraded performance (see Table \ref{tab:training_tech} for experimental evidence). We attribute this issue to the absence of some appealing optimization properties introduced by the three-stage training procedure, namely the lack of \emph{supervision}, \emph{input diversity} and \emph{training stability}. To solve these problems, we develop simple but effective training techniques, with which end-to-end training can significantly outperform the three-stage counterpart. These techniques do not introduce additional tunable hyper-parameters, while achieving consistent improvements across varying datasets, backbone architectures, and model configurations.

% In contrast, when solving problem (\ref{eq:naive_objective}), $f_{\textnormal{G}}$ and $f_{\textnormal{L}}$ are trained without pre-training, while they are only indirectly supervised by the gradients from the classifier $f_{\textnormal{C}}$.

\textbf{\emph{Lack of supervision}: auxiliary supervision.}
The effectiveness of three-stage training is largely ensured by a proper initialization. For example, AdaFocusV1 pre-trains the two encoders (\emph{i.e.}, $f_{\textnormal{G}}, f_{\textnormal{L}}$) separately using a direct frame-wise recognition loss (by simply appending a fully-connected layer) \cite{Wang_2021_ICCV}. However, when solving problem (\ref{eq:naive_objective}), we do not introduce such pre-training mechanisms, which hurts the overall training efficiency of our method. In other words, $f_{\textnormal{G}}$ and $f_{\textnormal{L}}$ are trained without specialized initialization, while they are only indirectly supervised by the gradients from the classifier $f_{\textnormal{C}}$. To this end, we find that explicitly introducing auxiliary supervision on $f_{\textnormal{G}}$ and $f_{\textnormal{L}}$ effectively facilitates the efficient end-to-end training of AdaFocus. In specific, we attach two linear classifiers, $\textnormal{FC}_{\textnormal{G}}(\cdot)$ and $\textnormal{FC}_{\textnormal{L}}(\cdot)$, to the outputs of $f_{\textnormal{G}}$ and $f_{\textnormal{L}}$, and replace the loss function $\mathcal{L}$ in (\ref{eq:naive_objective}) by $\mathcal{L}'$:
\begin{equation}
    \label{eq:l_prime}
    \begin{split}
    \mathcal{L}'={\mathbb{E}}_{\{\bm{v}_1, \bm{v}_2, \ldots\}}
    \left\{\right.
        &\frac{1}{T}\sum\nolimits_{t=1}^{T} 
        \left[
            L_{\textnormal{CE}}(\bm{p}_t, y)\right. \\ 
            &+ L_{\textnormal{CE}}(\textnormal{SoftMax}(\textnormal{FC}_{\textnormal{G}}(\overline{\bm{e}}^{\textnormal{G}}_{t})), y) \\
            &+ \left.L_{\textnormal{CE}}(\textnormal{SoftMax}(\textnormal{FC}_{\textnormal{L}}(\overline{\bm{e}}^{\textnormal{L}}_{t})), y) 
        \right]
    \left.\right\}, 
    \end{split}
\end{equation}
where $\overline{\bm{e}}^{\textnormal{G}}_{t}$ and $\overline{\bm{e}}^{\textnormal{L}}_{t}$ are the feature vectors after performing global average pooling on the feature maps ${\bm{e}}^{\textnormal{G}}_{t}$ and ${\bm{e}}^{\textnormal{L}}_{t}$ output by $f_{\textnormal{G}}$ and $f_{\textnormal{L}}$. Intuitively,
minimizing $\mathcal{L}'$ enforces the two encoders to learn linearized deep representations, which has been widely verified as an efficient approach for training deep networks \cite{he2016deep, huang2019convolutional, dosovitskiy2021an}. This paradigm benefits the learning of $f_{\textnormal{C}}$ as well, since its inputs are explicitly regularized to be linearly separable. 

\textbf{\emph{Lack of input diversity}: diversity augmentation.}
In the stage I for training AdaFocusV1, image patches are randomly cropped, yielding highly diversified inputs for learning well-generalized local encoder $f_{\textnormal{L}}$. However, the patch selection process presented in Section \ref{sec:interpolation} is deterministic. In Eq. (\ref{eq:l_prime}), given a video frame, the local encoder $f_{\textnormal{L}}$ only has access to the patch specified by the policy network $\pi$. This procedure leads to the limited diversity of training data for the inputs of $f_{\textnormal{L}}$. Empirically, we observe that it results in the inferior performance of $f_{\textnormal{L}}$. We address this issue by proposing a straightforward diversity augmentation approach. For each video, we first compute $\mathcal{L}'$ by activating $\pi$ as aforementioned. Then we infer $f_{\textnormal{L}}$ and the classifier $f_{\textnormal{C}}$ for a second time using randomly cropped patches, obtaining an additional loss $\mathcal{L}'_{\textnormal{random}}$, which follows Eq. (\ref{eq:l_prime}) as well. Our final optimization objective combines $\mathcal{L}'$ and $\mathcal{L}'_{\textnormal{random}}$:
\begin{equation}
    \label{eq:final_objective}
        \mathop{\textnormal{minimize}}_{f_{\textnormal{G}}, f_{\textnormal{L}}, f_{\textnormal{C}}, \pi}\ \ 
        \mathcal{L}=\frac{1}{2} (\mathcal{L}'+\mathcal{L}'_{\textnormal{random}}).
\end{equation}

\textbf{\emph{Lack of training stability}: stop-gradient.}
In AdaFocusV1, the policy network $\pi$ is learned on top of the fixed and completely trained global encoder $f_{\textnormal{G}}$. When it comes to end-to-end training, $\pi$ and $f_{\textnormal{G}}$ are simultaneously updated. In this case, we observe that the gradients back-propagated from $\pi$ interfere with the learning of $f_{\textnormal{G}}$, leading to an unstable training process with slow convergence speed. We find that this problem can be solved by simply stopping the gradients before the inputs of $\pi$. In other words, we propose to train $f_{\textnormal{G}}$ using the pure classification objective without any effect from $\pi$, as done in AdaFocusV1. This design is rational since previous works have revealed that the representations extracted by deep recognition networks can naturally be leveraged for localizing task-relevant regions \cite{zhou2016learning, selvaraju2017grad, dosovitskiy2021an}.

% \vspace{-0.5ex}
\section{Unified AdaFocus (Uni-AdaFocus)}
% \vspace{-0.5ex}
\label{sec:uni_adafocus}

% In this section, we further propose an improved unified AdaFocus (Uni-AdaFocus) framework. Uni-AdaFocus retains the merits of AdaFocusV2 in terms of efficient end-to-end training, but improves AdaFocusV2 in several important aspects. First, the spatial dynamic computation algorithm has been upgraded and advanced: 1) we present a deep-feature-interpolation-based mechanism for training the policy network $\pi$, which introduces negligible additional training cost, but contributes to a more stable learning process of $\pi$ and improves the final performance; and 2) we propose a deformable patch mechanism, which introduces the resizable rectangle image patches and thus is able to capture the task-relevant regions in video frames with varying shapes or sizes.

% In this section, we present an advanced Unified AdaFocus (Uni-AdaFocus) framework, which refines and extends AdaFocusV2 in several critical aspects. Firstly, the spatial dynamic computation algorithm is enhanced and developed (Section \ref{sec:improved_spatial}): 1) we introduce a deep-feature-interpolation-based technique for the effective training of policy network $\pi$, which incurs minimal additional training cost, yet facilitates a more stable learning process for $\pi$ and improves the final performance; and 2) we implement a deformable patch mechanism incorporating resizable rectangular image patches, allowing the capture of task-relevant regions with diverse shapes and sizes within each video frame.

In this section, we further introduce an enhanced Unified AdaFocus (Uni-AdaFocus) framework, which improves AdaFocusV2 in several important aspects. First, the spatial dynamic computation algorithm is refined and advanced (Section \ref{sec:improved_spatial}): 1) we introduce a deep-feature-interpolation-based technique for the effective training of the policy network $\pi$, which incurs minimal additional training cost, yet facilitates a more stable learning process for $\pi$ and improves the final performance; and 2) we develop a deformable patch mechanism incorporating resizable rectangular image patches, enabling the capture of task-relevant regions with diverse shapes and sizes within each video frame. 

% Additionally, we broaden our method by concurrently modeling spatial-temporal dynamic computation. This is accomplished by proposing a dynamic frame sampling algorithm specifically designed for AdaFocus (Section \ref{sec:frame_sample}), which enables our approach to predominantly allocate computation to the task-relevant spatial regions of the most informative frames. Finally, we illustrate that AdaFocus can be further optimized by minimizing sample-wise redundancy (Section \ref{sec:early_exit}), \emph{i.e.}, unevenly distributing computation across easier' and harder' videos, leading to a significant enhancement in overall efficiency.

Additionally, we extend our method by simultaneously modeling the spatial-temporal dynamic computation. This is accomplished by proposing a dynamic frame sampling algorithm tailored for AdaFocus (Section \ref{sec:frame_sample}), which enables our approach to allocate the majority of computation to the task-relevant spatial regions of the most informative frames. Lastly, we demonstrate that AdaFocus can be further optimized by reducing the sample-wise redundancy (Section \ref{sec:early_exit}), \emph{i.e.}, unevenly distributing computation across relatively ``easier'' and ``harder'' videos, leading to a considerable improvement of the overall efficiency.

% Built upon these technical innovations, Uni-AdaFocus is a highly general framework that unifies spatial-wise, temporal-wise, and sample-wise dynamic computation. It retains the advantages of AdaFocusV2 regarding efficient end-to-end training and being compatible with various backbone networks, while significantly improves the theoretical and practical computational efficiency during inference.

% By capitalizing on these technical advancements, Uni-AdaFocus establishes a holistic framework that seamlessly combines spatial-wise, temporal-wise, and sample-wise dynamic computation. It retains the benefits of AdaFocusV2 concerning efficient end-to-end training and compatibility with a wide range of backbone networks, while substantially improving both theoretical and practical computational efficiency during inference.

By leveraging these technical innovations, Uni-AdaFocus establishes a holistic framework that seamlessly combines spatial, temporal, and sample-wise dynamic computation. It retains the merits of AdaFocusV2 in efficient end-to-end training and being compatible with a wide range of backbone networks, while substantially improving both theoretical and practical computational efficiency during inference.

% \vspace{-0.4ex}
\subsection{Improved Spatial Dynamic Computation}
% \vspace{-0.4ex}
\label{sec:improved_spatial}

% Subsequently, we will examine two significant shortcomings of the spatial dynamic mechanism in AdaFocusV2, specifically the unsuitable supervision signals for the policy network $\pi$, and the suboptimal formulation stemming from the fixed shape and size of informative patches. Building upon our analysis, we will demonstrate how AdaFocusV2 can be refined to address these concerns. The ablation studies for these modifications are provided in Table XXX.

In the following, we examine two important shortcomings of the spatial dynamic computation mechanism in AdaFocusV2, namely the unsuitable supervision signals for the policy network $\pi$, and the sub-optimal formulation stemming from the fixed shape and size of informative patches. On top of our analysis, we will demonstrate how AdaFocusV2 can be refined to address these concerns. The ablation studies for these modifications are provided in Table \ref{tab:uni_techs}.

\subsubsection{Training the Policy Network with Feature Interpolation}
\label{sec:improved_spatial_sub1}

% While AdaFocusV2 facilitates end-to-end training of the policy network $\pi$, the gradients obtained by $\pi$ may be ambiguous and indirect with respect to learning an appropriate spatial dynamic computation policy. Ideally, during training, the model should recognize how the semantic content of $\tilde{\bm{v}}_t$ changes with different outputs of $\pi$, as the goal is to localize the most informative portions of each video frame. However, as illustrated in Eqs. (\ref{eq:bilinear} -- \ref{eq:bp_2}), the gradients of $\pi$ are acquired by modifying the pixel values of the patch $\tilde{\bm{v}}_t$. Each pixel value is determined solely by four neighboring pixels in the original video frame $\bm{v}_t$. This limited information source might be too localized to accurately represent the semantic-level changes when the location of $\tilde{\bm{v}}_t$ varies.

\textbf{Inappropriate supervision signals for $\pi$.}
AdaFocusV2 has enabled the end-to-end training of the policy network $\pi$. However, the gradients received by $\pi$ may be ambiguous and indirect with respect to learning a proper spatial dynamic computation policy. Ideally, during training, the model should be aware of how the semantic contents of the patch $\tilde{\bm{v}}_t$ will change with different outputs of $\pi$, as our goal is to localize the most informative parts of each video frame. Nevertheless, as shown in Eqs. (\ref{eq:bilinear} -- \ref{eq:bp_2}), the gradients of $\pi$ are acquired by varying the pixel values of $\tilde{\bm{v}}_t$. Each pixel value is determined solely by four neighboring pixels in the original video frame $\bm{v}_t$. Such a limited source of information might be too local to accurately represent the semantic-level changes when the location of $\tilde{\bm{v}}_t$ varies.

\begin{figure}[!h]
    \vskip -0.05in
    \begin{center}
    \centerline{\includegraphics[width=0.9\columnwidth]{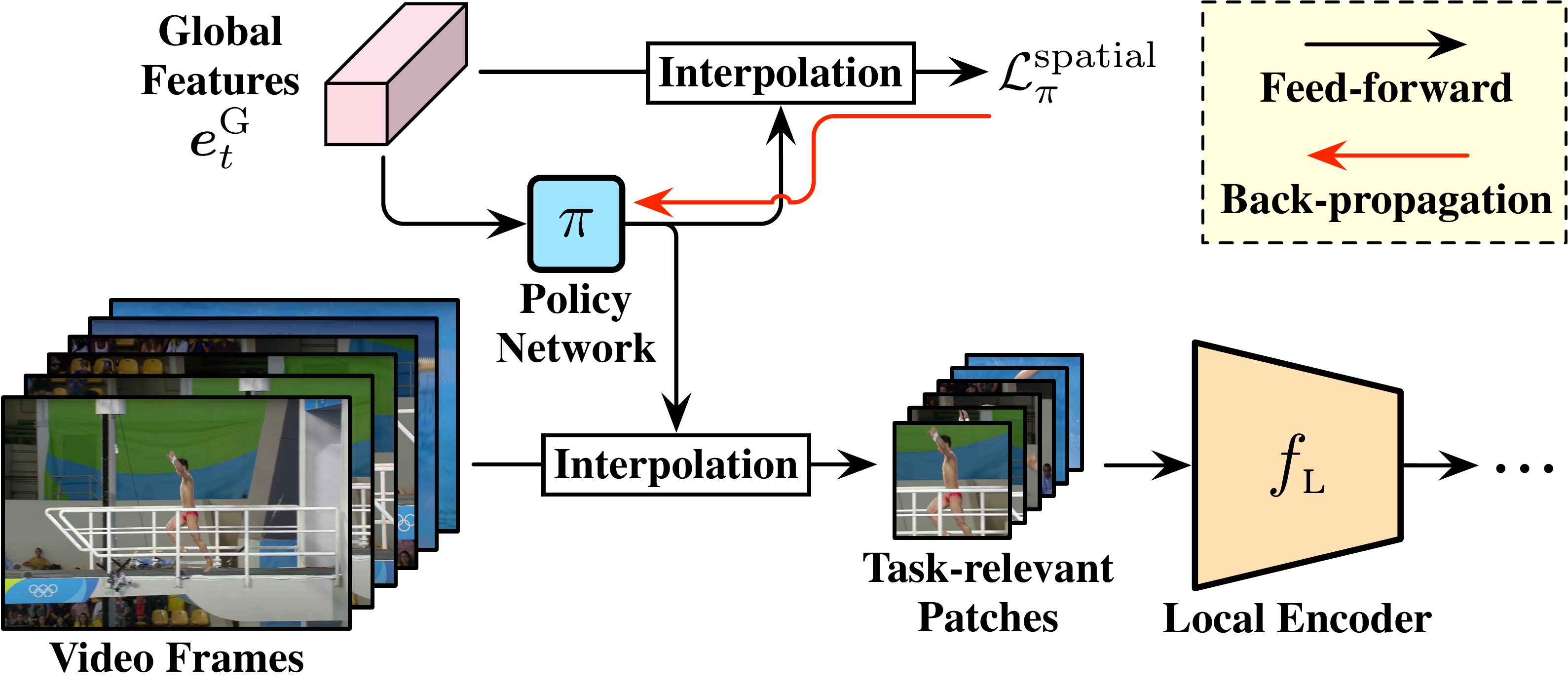}}
    \vskip -0.125in
    \caption{\textbf{Guiding the training of ${\pi}$ with deep features.} The gradients for $\pi$ are obtained by minimizing the deep-feature-based loss $\mathcal{L}_{\pi}^{\textnormal{spatial}}$, instead of being back propagated from the pixel space. With this design, the supervision signals for ${\pi}$ contain much more semantic-level information, which contributes to learning a more effective patch selection policy. \label{fig:train_pi_feature}
    }
    \end{center}
    \vspace{-1ex}
\end{figure}

% To mitigate the issue arising from pixel-space-based gradient computation for $\pi$, we suggest guiding the training of $\bm{\pi}$ by utilizing deep features. Our primary insight involves allowing the outputs of $\bm{\pi}$ to directly impact specific deep representations associated with $\tilde{\bm{v}}_t$. It has been established that features extracted by deep networks excel at capturing the task-relevant semantics of the inputs \cite{bengio2013better,upchurch2017deep,wang2019implicit}. By enabling the gradients to explicitly inform $\bm{\pi}$ about the effects of its outputs on $\tilde{\bm{v}}_t$ at the deep feature level, $\bm{\pi}$ will receive direct supervision signals regarding whether $\tilde{\bm{v}}_t$ contains valuable content for the recognition task.

\textbf{Gradients derived from interpolating deep features.}
To alleviate the problem arising from the pixel-space-based gradient computation for $\pi$, we propose to guide the training of ${\pi}$ by utilizing deep features. Our primary insight is to allow the outputs of ${\pi}$ to have a direct influence on certain deep representations associated with $\tilde{\bm{v}}_t$. The features extracted by deep networks are known to excel at capturing the task-relevant semantics of the inputs \cite{bengio2013better,wang2019implicit, li2020deep, xie2023sepico, xie2024adapting}. Once the gradients can explicitly inform ${\pi}$ about how its outputs will affect $\tilde{\bm{v}}_t$ at the deep feature level, ${\pi}$ will receive direct supervision signals regarding whether $\tilde{\bm{v}}_t$ contains valuable contents for the recognition task.

Driven by this motivation, we propose an elegant implementation by reusing the coarse global feature maps $\bm{e}^{\textnormal{G}}_{t}$. Assume that we want to obtain the $P\!\times\!P$ task-relevant patch $\tilde{\bm{v}}_t$ centred at $(\tilde{x}^t_{\textnormal{c}}, \tilde{y}^t_{\textnormal{c}})$ from the $H\!\times\!W$ video frame $\bm{v}_t$. We approximate this process on top of $\bm{e}^{\textnormal{G}}_{t}$. Specifically, suppose the size of $\bm{e}^{\textnormal{G}}_{t}$ is $H^{\textnormal{G}}\!\times\!W^{\textnormal{G}}$. We take the $\frac{H^{\textnormal{G}}}{H}P\!\times\!\frac{W^{\textnormal{G}}}{W}P$ feature patch centred at $(\frac{H^{\textnormal{G}}}{H}\tilde{x}^t_{\textnormal{c}}, \frac{W^{\textnormal{G}}}{W}\tilde{y}^t_{\textnormal{c}})$ from $\bm{e}^{\textnormal{G}}_{t}$, which we refer to as ${\bm{e}^{\textnormal{G}}_{t}}'$. Formally, we have
\begin{equation}
    \label{eq:interpolate_deep_feature_for_pi_0}
    {\bm{e}^{\textnormal{G}}_{t}}' = \textnormal{Interpolation}_{\frac{H^{\textnormal{G}}}{H}P\times\frac{W^{\textnormal{G}}}{W}P}(\bm{e}^{\textnormal{G}}_{t}, (\frac{H^{\textnormal{G}}}{H}\tilde{x}^t_{\textnormal{c}}, \frac{W^{\textnormal{G}}}{W}\tilde{y}^t_{\textnormal{c}})).
\end{equation}
Here $\textnormal{Interpolation}(\cdot)$ corresponds to the differentiable interpolation-based patch cropping operation from AdaFocusV2. During training, $\pi$ is updated to minimize the classification loss with respect to ${\bm{e}^{\textnormal{G}}_{t}}'$, namely
\begin{equation}
    \label{eq:interpolate_deep_feature_for_pi}
    \begin{split}
    \mathop{\textnormal{minimize}}_{\pi}\ \ &\mathcal{L}_{\pi}^{\textnormal{spatial}} \\[-0.75ex]
    =\ &{\mathbb{E}}_{\{\bm{v}_1, \bm{v}_2, \ldots\}}\!\left[
        L_{\textnormal{CE}}(\textnormal{SoftMax}(\textnormal{FC}_{\textnormal{Aux}}({\overline{\bm{e}}^{\textnormal{G}}_{t}}')), y)
    \right],
    \end{split}
    % \frac{\partial\mathcal{L}}{\partial\pi} = \frac{\partial L_{\textnormal{CE}}(\textnormal{SoftMax}(\textnormal{FC}_{\textnormal{Aux}}({\overline{\bm{e}}^{\textnormal{G}}_{t}}')), y)}{\partial\pi},
\end{equation}
where ${\overline{\bm{e}}^{\textnormal{G}}_{t}}'$ is derived by pooling ${\bm{e}^{\textnormal{G}}_{t}}'$, $y$ is the label and $\textnormal{FC}_{\textnormal{Aux}}(\cdot)$ denotes a linear classifier. The gradients ${\partial\mathcal{L}_{\pi}^{\textnormal{spatial}}}/{\partial\pi}$ can be solved by leveraging Eqs. (\ref{eq:bilinear} -- \ref{eq:bp_2}). It should be noted that only $\pi$ is updated following Eq. (\ref{eq:interpolate_deep_feature_for_pi}), while the training of all the other components remains unchanged. Furthermore, the process of obtaining $\tilde{\bm{v}}_t$ from $\bm{v}_t$ and feeding $\tilde{\bm{v}}_t$ into the local encoder $f_{\textnormal{L}}$ is not altered. The only difference is that the gradients of $\pi$ are calculated using (\ref{eq:interpolate_deep_feature_for_pi}) rather than being back-propagated from $\tilde{\bm{v}}_t$. 
See Figure \ref{fig:train_pi_feature} for an illustration.

% An illustration is provided in Figure \ref{fig:train_pi_feature}.

% Where ${\overline{\bm{e}}^{\textnormal{G}}{t}}'$ is derived by pooling ${\bm{e}^{\textnormal{G}}{t}}'$, $y$ represents the label, and $\textnormal{FC}{\textnormal{Aux}}(\cdot)$ denotes a linear classifier. The gradients ${\partial\mathcal{L}{\pi}^{\textnormal{spatial}}}/{\partial\pi}$ can be computed using Eqs. (\ref{eq:bilinear} -- \ref{eq:bp_2}). It should be noted that only $\pi$ is updated according to Eq. (\ref{eq:interpolate_deep_feature_for_pi}), while the training of all other components remains unchanged. Furthermore, the process of obtaining $\tilde{\bm{v}}_t$ from $\bm{v}_t$ and feeding $\tilde{\bm{v}}t$ into the local encoder $f{\textnormal{L}}$ is not altered. The sole distinction is that the gradients of $\pi$ are calculated following (\ref{eq:interpolate_deep_feature_for_pi}) rather than being back-propagated from $\tilde{\bm{v}}_t$.

The underlying logic behind Eq. (\ref{eq:interpolate_deep_feature_for_pi}) is straightforward. In the context of training $\pi$, we employ the change of ${\bm{e}^{\textnormal{G}}_{t}}'$ to estimate the change of the deep features corresponding to $\tilde{\bm{v}}_t$. This design is reasonable since the relative position of ${\bm{e}^{\textnormal{G}}_{t}}'$ on ${\bm{e}^{\textnormal{G}}_{t}}$ is identical to that of $\tilde{\bm{v}}_t$ on $\bm{v}_t$. That is to say, the location-preserving nature of deep features ensures that ${\bm{e}^{\textnormal{G}}_{t}}'$ represents the contents at the location of $\tilde{\bm{v}}_t$. 

Notably, Eq. (\ref{eq:interpolate_deep_feature_for_pi}) is conceptually relevant to AdaFocusV3 \cite{wang2022adafocusv3} in training $\pi$ using deep features. However, compared to \cite{wang2022adafocusv3}, our method provides more effective guidance for $\pi$ based on the global information within ${\bm{e}^{\textnormal{G}}_{t}}$, introduces negligible computational overhead during training (\emph{i.e.}, only an additional time of linear classification: $\textnormal{FC}_{\textnormal{Aux}}(\cdot)$), and does not leads to a discrepancy between training/test inputs for the local encoder $f_{\textnormal{L}}$. See Appendix \ref{app:additional_discussions} for more comparisons.

\subsubsection{Deformable Patches}
\label{sec:improved_spatial_sub2}

\textbf{Limitations of fixed-size patches.}
In both AdaFocusV1 and V2, the informative image patches $\{\tilde{\bm{v}}_1, \tilde{\bm{v}}_2, \ldots\}$ are assumed to have a fixed size (\emph{i.e.}, $P\!\times\!P$). The model is only trained to dynamically determine their locations to capture the most task-relevant spatial regions of each video frame. This design is straightforward, and it simplifies the model training process. However, in contrast, it compromises the flexibility of AdaFocus. As a matter of fact, the shapes and sizes of discriminative regions may vary across different videos and different frames. By design, the patches with a fixed size are inherently unable to accommodate the informative regions adaptively conditioned on each individual video frame. In other words, $\{\tilde{\bm{v}}_1, \tilde{\bm{v}}_2, \ldots\}$ may either omit crucial task-relevant details or include undesirable redundant contents.

\begin{figure}[!h]
    % \vskip -0.1in
    \begin{center}
    \centerline{\includegraphics[width=0.9\columnwidth]{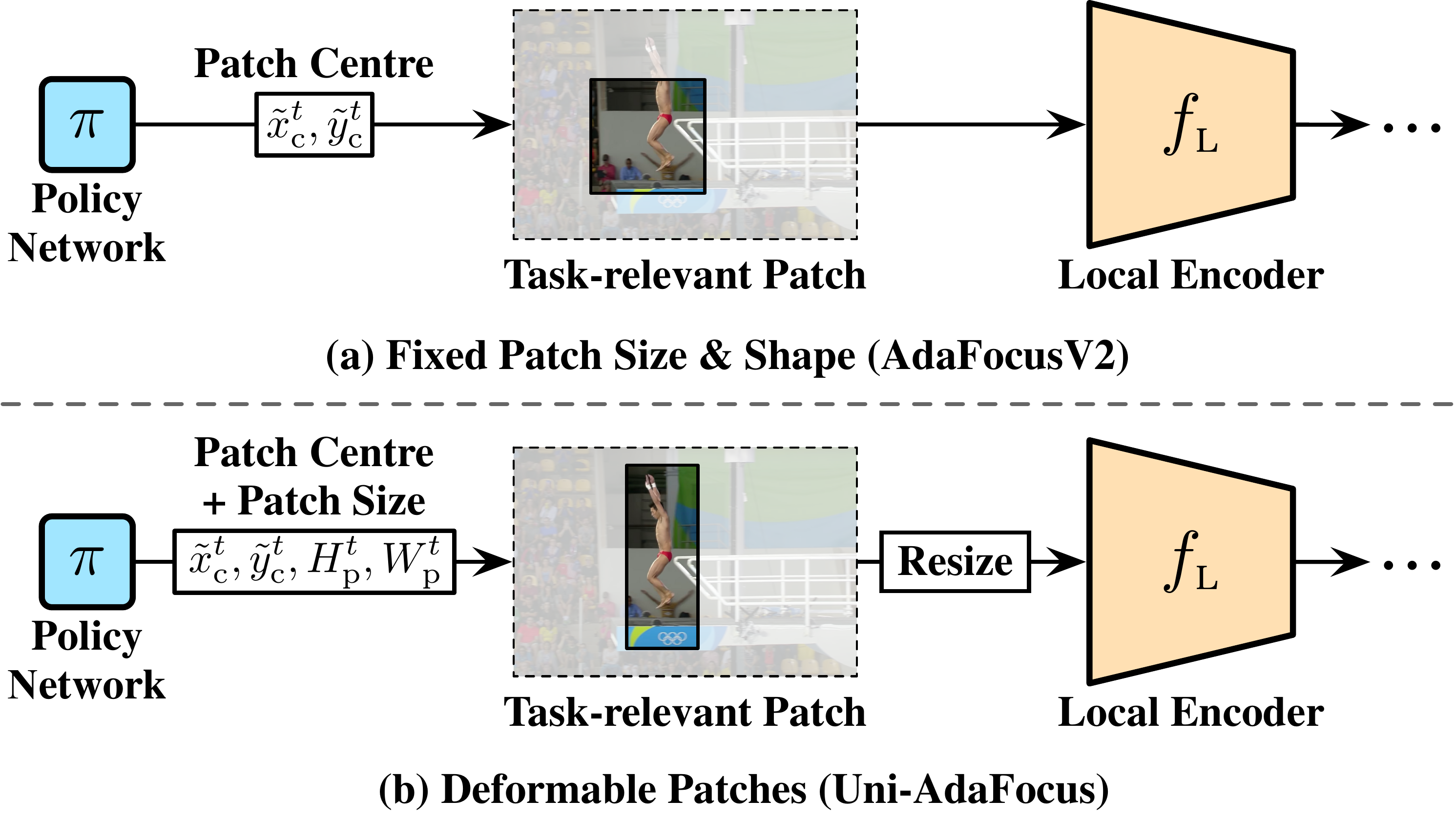}}
    \vskip -0.15in
    \caption{\textbf{Deformable patches}, which enable the patch selection policy to adapt flexibly to the task-relevant regions in various shapes, scales, and locations. The selected patches are resized to a common size $P^2$ to be processed efficiently by $f_{\textnormal{L}}$ on hardwares (\emph{e.g.}, GPUs).
    Notably, the flexibility of adapting to arbitrary task-relevant regions significantly outweighs the weakness of geometric distortion, yielding considerable accuracy improvements across diverse scenarios
    \label{fig:deformable_patch}
    }
    \end{center}
    \vspace{-2.3ex}
\end{figure}

\textbf{Incorporating deformable patches.}
To address this issue, we propose to allow the shape and size of the patches $\{\tilde{\bm{v}}_1, \tilde{\bm{v}}_2, \ldots\}$ to be dynamically configured according to the characteristics and semantics of each frame. To be specific, we let the policy network $\pi$ output a quadruple: $(\tilde{x}^t_{\textnormal{c}}, \tilde{y}^t_{\textnormal{c}}, H_{\textnormal{p}}^t, W_{\textnormal{p}}^t)$, where $\tilde{x}^t_{\textnormal{c}}, \tilde{y}^t_{\textnormal{c}}$ represent the centre coordinates of the patch $\tilde{\bm{v}}_t$, while $H_{\textnormal{p}}^t, W_{\textnormal{p}}^t$ denote its height and width. Then we directly take the $H_{\textnormal{p}}^t\!\times\!W_{\textnormal{p}}^t$ patch at $(\tilde{x}^t_{\textnormal{c}}, \tilde{y}^t_{\textnormal{c}})$ to obtain $\tilde{\bm{v}}_t$. It is worth noting that before feeding $\tilde{\bm{v}}_t$ into the local encoder $f_{\textnormal{L}}$, we resize all the patches to $P\!\times\!P$, such that they can be conveniently processed in parallel on hardwares. Although this implementation technique may result in slight geometric distortion, we observe that the flexibility of adapting to arbitrary task-relevant regions significantly outweighs this weakness, leading to considerable accuracy improvements. An illustration of this procedure is depicted in Figure \ref{fig:deformable_patch}.

% Given the aforementioned formulation, a crucial question arises concerning how to learn an appropriate $\pi$ that can assign a suitable patch height/width to each frame. A straightforward approach may involve replacing the $P$ in Eq. (\ref{eq:interpolate_deep_feature_for_pi_0}) with $H_{\textnormal{p}}^t$ or $W_{\textnormal{p}}^t$, and subsequently obtaining the gradients following Eq. (\ref{eq:interpolate_deep_feature_for_pi}). Nevertheless, Eq. (\ref{eq:interpolate_deep_feature_for_pi}) is not specifically designed for learning to determine $H_{\textnormal{p}}^t$ and $W_{\textnormal{p}}^t$, and simply utilizing it would lead to a trivial optimization solution, \emph{i.e.}, encouraging $H_{\textnormal{p}}^t, W_{\textnormal{p}}^t \to 0$ for all video frames. This outcome can be attributed to the nature of deep features: it is always possible to identify a point within any feature map at which the feature minimizes the classification loss. Consequently, Eq. (\ref{eq:interpolate_deep_feature_for_pi}) will push the resizable rectangular patches to fit this single point. We discover that this issue can be circumvented by slightly modifying Eq. (\ref{eq:interpolate_deep_feature_for_pi}), specifically

\textbf{Training algorithm.}
Given the aforementioned formulation, an important problem is how to learn a proper $\pi$ that can assign a suitable patch height/width to each frame. As a straightforward approach, one may replace the $P$ in Eq. (\ref{eq:interpolate_deep_feature_for_pi_0}) with $H_{\textnormal{p}}^t$ or $W_{\textnormal{p}}^t$, and subsequently obtain the gradients following Eq. (\ref{eq:interpolate_deep_feature_for_pi}). Nevertheless, Eq. (\ref{eq:interpolate_deep_feature_for_pi}) is not specifically designed for learning to determine $H_{\textnormal{p}}^t$ and $W_{\textnormal{p}}^t$, and simply utilizing it will lead to a trivial optimization solution, \emph{i.e.}, encouraging $H_{\textnormal{p}}^t, W_{\textnormal{p}}^t \to 0$ for all video frames. This outcome can be attributed to the characteristics of deep features: we can always find a point in any feature map, at which location the feature minimizes the recognition loss. As a consequence, Eq. (\ref{eq:interpolate_deep_feature_for_pi}) will push the resizable rectangular patches to fit this single point. We find this problem can be circumvented by slightly modifying Eq. (\ref{eq:interpolate_deep_feature_for_pi}), specifically,
\begin{equation}
    \label{eq:learning_deformable_patches}
    \begin{split}
        \mathop{\textnormal{minimize}}_{\pi}\ \ &\mathcal{L}_{\pi}^{\textnormal{spatial}} \\[-0.75ex]
        =\ &{\mathbb{E}}_{\{\bm{v}_1, \bm{v}_2, \ldots\}}\left[
        L_{\textnormal{CE}}(\textnormal{SoftMax}(\textnormal{FC}_{\textnormal{Aux}}({\overline{\bm{e}}^{\textnormal{G}}_{t}}')), y) \right.
         \\
    & + \alpha \!
        \left. \left( (H\!-\!H_{\textnormal{p}}^t)^2 + (W\!-\!W_{\textnormal{p}}^t)^2\right)\right],
    \end{split}
\end{equation}
where $H\!\times\!W$ is the size of the original video frames. In (\ref{eq:learning_deformable_patches}), we introduce a regularization term $\alpha[(H\!-\!H_{\textnormal{p}}^t)^2\!+\!(W\!-\!W_{\textnormal{p}}^t)^2]$ to explicitly promote larger $H_{\textnormal{p}}^t, W_{\textnormal{p}}^t$, and $\alpha$ is a pre-defined coefficient to control the strength of regularization. Despite its simplicity, Eq. (\ref{eq:learning_deformable_patches}) is capable of training flexible and adaptable patch selection policies that effectively capture task-relevant regions in various shapes, sizes, scales, and locations, resulting in significant accuracy improvements across diverse scenarios.

% is capable of training flexible and adaptable patch selection policies capable of capturing task-relevant regions in diverse shapes, sizes, scales, and locations, contributing to significant improvements in accuracy in various scenarios.
% In Eq. (\ref{eq:learning_deformable_patches}), we introduce a regularization term $\alpha[(H!-!H_{\textnormal{p}}^t)^2!+!(W!-!W_{\textnormal{p}}^t)^2]$ to explicitly promote larger $H_{\textnormal{p}}^t, W_{\textnormal{p}}^t$, and $\alpha$ is a pre-defined coefficient to govern the regularization strength. Despite its simplicity, Eq. (\ref{eq:learning_deformable_patches}) is capable of training flexible and adaptable patch selection policies that effectively capture task-relevant regions in various shapes, sizes, scales, and locations, resulting in significant accuracy improvements across diverse scenarios.

\begin{figure*}[!t]
    % \vskip -0.1in
    \begin{center}
    \centerline{\includegraphics[width=0.86\linewidth]{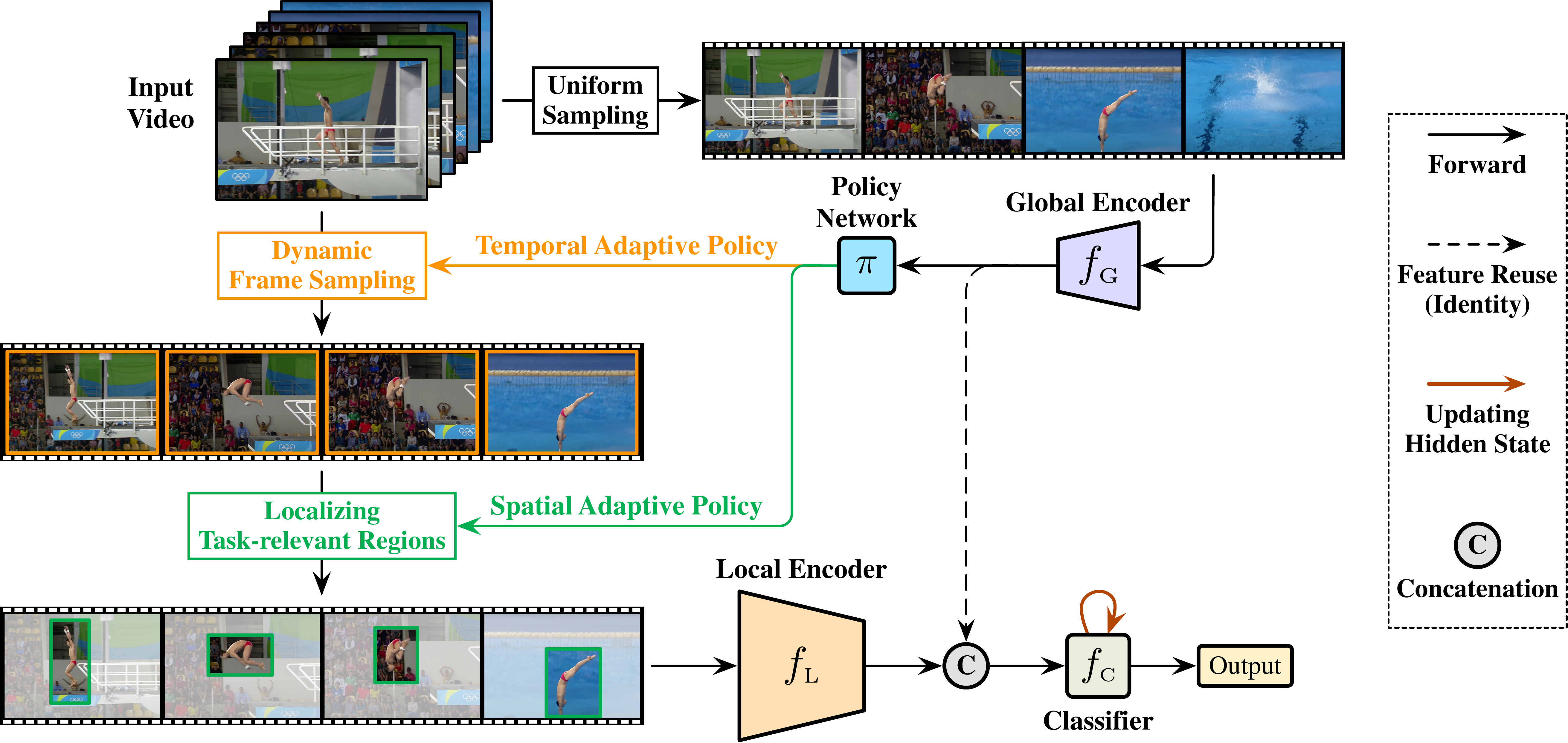}}
    \vskip -0.125in
    \caption{\textbf{Overview of Uni-AdaFocus}, a unified framework that models spatial-temporal dynamic computation concurrently, leading to a considerably improved computational efficiency for inference. Compared with Figure \ref{fig:overview}, here we let $\pi$ simultaneously determine which frames are more task-relevant, as well as which spatial regions within these frames contain more valuable information for the task. The high-capacity and computationally intensive local encoder $f_{\textnormal{L}}$ is only activated on the most informative regions of the most important frames. Notably, similar to AdaFocusV2, Uni-AdaFocus can be trained efficiently in an end-to-end fashion, and is compatible with off-the-shelf advanced backbone networks (\emph{e.g.}, TSM \cite{lin2019tsm} and X3D \cite{feichtenhofer2020x3d}).
    \label{fig:overview_uniada}
    }
    \end{center}
    \vspace{-1.5ex}
\end{figure*}

% consistently
% an arbitrary size/scale
% Particularly, before fed into the local encoder $f_{\textnormal{L}}$, we 
% the continuous centre coordinates $(\tilde{x}^t_{\textnormal{c}}, \tilde{y}^t_{\textnormal{c}})$ of $\tilde{\bm{v}}_t$
% The fixed size of patches may omit important 

% Up to this point, we have extensively discussed the learning process to establish a flexible spatial dynamic computation mechanism. However, our methodology thus far operates under a basic assumption, that is, the computation is uniformly distributed along the temporal dimension. In reality, AdaFocus can be further enhanced by minimizing temporal redundancy, specifically by adaptively identifying and focusing on the most task-relevant video frames. In this subsection, we will illustrate how this objective can be achieved with a dynamic frame sampling algorithm specifically designed for our approach.

\subsection{Reducing Temporal Redundancy -- Dynamic Frame Sampling}
\label{sec:frame_sample}

Up to this point, we have presented considerable discussions on learning to establish a flexible spatial dynamic computation mechanism. However, our methodology thus far follows a basic assumption, that is, the computation is uniformly distributed along the temporal dimension. In fact, AdaFocus can be further improved by reducing the temporal-wise redundancy, namely adaptively identifying and attending to the most task-relevant video frames. In this subsection, we will demonstrate that this objective can be achieved with a dynamic frame sampling algorithm tailored for our approach.

% Our objective is to devise a unified framework that addresses both spatial and temporal redundancy. Instead of processing videos on a frame-by-frame basis, we propose that the lightweight global encoder $f_{\textnormal{G}}$ initially scans the entire video. Specifically, given a video comprising $T_0$ frames in total, we uniformly sample $T_{\textnormal{G}}\ (T_{\textnormal{G}}!<!T_0)$ frames from the video and input them simultaneously into $f_{\textnormal{G}}$, directly obtaining the inexpensive and coarse global features corresponding to the entire video. Subsequently, we activate the policy network $\pi$ to concurrently determine: 1) the $T_{\textnormal{L}}\ (T_{\textnormal{L}}!<!T_0)$ task-relevant frames out of all the $T_0$ frames that the high-capacity local encoder $f_{\textnormal{L}}$ should process; and 2) the height, width, and location of the informative patch to be input into $f_{\textnormal{L}}$ for each frame identified in 1). We will first describe the process of obtaining 1) and then demonstrate how our previously discussed spatial dynamic computation techniques can be effortlessly adapted for acquiring 2). A visual representation of our unified framework is provided in Figure XXX.

\textbf{A unified framework.}
We seek to develop a unified framework that considers both spatial and temporal redundancies. Hence, instead of processing videos on a frame-by-frame basis, here we assume the lightweight global encoder $f_{\textnormal{G}}$ first takes a quick glance at the whole video. To be specific, suppose that a video comprises $T_0$ frames in total. We uniformly sample $T_{\textnormal{G}}\ (T_{\textnormal{G}}\!<\!T_0)$ frames from the video, and feed them into $f_{\textnormal{G}}$ at the same time, directly obtaining the inexpensive and coarse global features corresponding to the entire video. Then we activate the policy network $\pi$ to concurrently determine: 1) the $T_{\textnormal{L}}\ (T_{\textnormal{L}}\!<\!T_0)$ task-relevant frames out of all the $T_0$ frames that the high-capacity local encoder $f_{\textnormal{L}}$ should process; and 2) the height, width, and location of the informative patch to be fed into $f_{\textnormal{L}}$ for each frame identified in 1). In the following, we will first introduce how to acquire 1), and then demonstrate that our previously discussed spatial dynamic computation techniques can be effortlessly adapted for obtaining 2). An illustration of our unified framework is shown in Figure \ref{fig:overview_uniada}. 

% An example of the architecture of $\pi$ is shown in Figure XXX.

% To select $T_{\textnormal{L}}$ task-relevant frames from the original $T_0$ frames, we begin by examining a weighted sampling problem. We assume that $\pi$ generates a weight for each of the $T_0$ frames: ${w_1, \ldots, w_{T_0}}$, with $\sum_{i=1}^{T_0}!w_i!=!1$. Based on these weights, we execute weighted sampling without replacement over the $T_0$ frames, resulting in the selection of $T_{\textnormal{L}}$ frames. Formally, we denote the indices of these selected frames as $n_1, \ldots, n_{T_{\textnormal{L}}}$, which yields

\textbf{Mathematical formulation for dynamic frame sampling.}
To select $T_{\textnormal{L}}$ task-relevant frames from the original $T_0$ frames, we start by formulating a weighted sampling problem. We assume that $\pi$ generates a weight for each of the $T_0$ frames: $\{w_1, \ldots, w_{T_0}\}$, with $\sum_{j=1}^{T_0}\!w_j\!=\!1$. Based on these weights, we execute weighted sampling without replacement over $T_0$ frames, resulting in $T_{\textnormal{L}}$ selected frames. Formally, we refer to their indices as $n_1, \ldots, n_{T_{\textnormal{L}}}$, which yields
\begin{equation}
    \label{eq:weighted_sampling}
    \begin{split}
        n_{i} \sim\ &\textnormal{WeightedSample}(j \mid w^{(i)}_j,\ 1 \leq j \leq T_0), \\
        & w^{(i)}_j =
        \begin{cases}
        0,  & j \in \{n_1, \ldots, n_{i-1}\} \\
        w_j, & \textnormal{otherwise}
        \end{cases}.
    \end{split}
\end{equation}
By design, these $T_{\textnormal{L}}$ frames are expected to be more informative than the remaining $T_0\!-\!T_{\textnormal{L}}$ video frames in terms of the recognition task. This goal is attained by leveraging the training objective, which is expressed as
\begin{equation}
    \label{eq:temporal_training_objective}
    \begin{split}
    \mathop{\textnormal{minimize}}_{\pi}\ \ 
    &\mathcal{L}_{\pi}^{\textnormal{temporal}} \\[-1ex]
    = &\!\!\!\!\mathop{\mathbb{E}}_{\{\bm{v}_1, \bm{v}_2, \ldots\}}
    \!\!\left[
    \mathop{\mathbb{E}}_{\{n_1, n_2, \ldots\}}
    \!\!\left[
        \frac{1}{T_{\textnormal{L}}} \!\sum\nolimits_{i=1}^{T_{\textnormal{L}}} \!\!\mathcal{L}_{\textnormal{frame}}(\bm{v}_{n_i})
    \right]\right].
    \end{split}
\end{equation}
Herein, $\mathcal{L}_{\textnormal{frame}}(\bm{v}_{n_i})$ represents the loss corresponding to the $n_i^{\textnormal{th}}$ frame $\bm{v}_{n_i}$, which measures the quantity of the valuable task-relevant information encompassed by $\bm{v}_{n_i}$ or a short video clip centred at $\bm{v}_{n_i}$. The two expectations in $\mathcal{L}_{\pi}^{\textnormal{temporal}}$ are taken over the single-video frame-sampling distribution, and the different videos within training data, respectively. 

% The computation of $\mathcal{L}_{\textnormal{frame}}(\bm{v}_{n_i})$ will be introduced subsequently. 

Importantly, it is difficult to solve problem (\ref{eq:temporal_training_objective}) directly with gradient-based methods. Calculating the exact value of ${\mathbb{E}}_{\{n_1, n_2, \ldots\}}[\cdot]$ will lead to a complexity of $\mathcal{O}(P^{T_0}_{T_{\textnormal{L}}})$, where $P^{T_0}_{T_{\textnormal{L}}}$ denotes the number of $T_{\textnormal{L}}$-permutations of $T_0$. This inefficiency will make $\mathcal{L}_{\pi}^{\textnormal{temporal}}$ intractable when $T_{\textnormal{L}}$ is not very small (\emph{e.g.}, with $T_{\textnormal{L}}\!>\!8$). On the other hand, a more efficient approach is to estimate ${\mathbb{E}}_{\{n_1, n_2, \ldots\}}[\cdot]$ utilizing Monte Carlo sampling. However, this will eliminate the differentiability of $\mathcal{L}_{\pi}^{\textnormal{temporal}}$ with respect to $\pi$. As a consequence, $\pi$ cannot receive gradients from $\mathcal{L}_{\pi}^{\textnormal{temporal}}$ as supervision signals.

To address these challenges, we propose to further decompose $\mathcal{L}_{\pi}^{\textnormal{temporal}}$. In specific, we have
\begin{equation}
    \label{eq:decomposing_weighted_sampling}
    \begin{split}
        & \mathop{\mathbb{E}}_{\{n_1, n_2, \ldots\}}
        \!\!\left[
            \frac{1}{T_{\textnormal{L}}} \!\sum\nolimits_{i=1}^{T_{\textnormal{L}}} \!\!\mathcal{L}_{\textnormal{frame}}(\bm{v}_{n_i})
        \right] \\
        =&
        \frac{1}{T_{\textnormal{L}}}\!\sum\nolimits_{i=1}^{T_{\textnormal{L}}} 
        \mathop{\mathbb{E}}_{\{n_1, \ldots, n_i\}}\!\!
            \mathcal{L}_{\textnormal{frame}}(\bm{v}_{n_i})
        \\
        =&
        \frac{1}{T_{\textnormal{L}}}\!\sum\nolimits_{i=1}^{T_{\textnormal{L}}} \!
        \mathop{\mathbb{E}}_{\{n_1, \ldots, n_{i-1}\}}\!\!
        \left[
            \mathop{\mathbb{E}}_{
                p(n_i|n_1, \ldots, n_{i-1})
            }\!\!
            \mathcal{L}_{\textnormal{frame}}(\bm{v}_{n_i})
        \right] 
        \\
        =&
        \frac{1}{T_{\textnormal{L}}}\!\sum\nolimits_{i=1}^{T_{\textnormal{L}}} \!
        \mathop{\mathbb{E}}_{\{n_1, \ldots, n_{i-1}\}}\!\!
        \left[
            \sum\nolimits_{j=1}^{T_0}\!\!
            \frac{w_j^{(i)}}{\sum\nolimits_{j'=1}^{T_0} w_{j'}^{(i)}}
            \mathcal{L}_{\textnormal{frame}}(\bm{v}_{j})
        \right],
    \end{split}
\end{equation}
where $w_j^{(i)}$ is the sampling weights corresponding to $n_{i}$, as described in Eq. (\ref{eq:weighted_sampling}). Consider approximating Eq. (\ref{eq:decomposing_weighted_sampling}) using $M$ Monte Carlo samples, namely
\begin{equation}
    \label{eq:monte_carlo_1}
    \{n^m_1, \ldots, n^m_{T_{\textnormal{L}}}\} \sim p(\cdot|w_1, \ldots, w_{T_0}),
    \quad 1 \leq m \leq M,
\end{equation}
\begin{equation}
    \label{eq:monte_carlo_2}
    w^{m, (i)}_j =
        \begin{cases}
        0,  & j \in \{n^m_1, \ldots, n^m_{i-1}\} \\
        w_j, & \textnormal{otherwise}
        \end{cases}.
\end{equation}
By combining Eqs. (\ref{eq:temporal_training_objective} -- \ref{eq:monte_carlo_2}), we finally obtain
\begin{equation}
    \label{eq:final_temporal_obj}
    \begin{split}
        &\mathcal{L}_{\pi}^{\textnormal{temporal}} \\[-1ex]
        \approx & 
        \frac{1}{T_{\textnormal{L}}M}\!\!
        \sum\nolimits_{i=1}^{T_{\textnormal{L}}}\!
        \sum\nolimits_{m=1}^{M} \!\!
        \left[
            \sum\nolimits_{j=1}^{T_0}\!\!
            \frac{w_j^{m, (i)}}{\sum\nolimits_{j'=1}^{T_0} w_{j'}^{m, (i)}}
            \mathcal{L}_{\textnormal{frame}}(\bm{v}_{j})
        \right]\!\!.
    \end{split}
\end{equation}
As the complexity of generating a single Monte Carlo sample is $\mathcal{O}(T_0 + T_{\textnormal{L}} \log T_0)$, solving Eq. (\ref{eq:final_temporal_obj}) has a total complexity of $\mathcal{O}(T_{\textnormal{L}}(T_0 + T_{\textnormal{L}} \log T_0) \cdot M)$, which is dramatically more efficient than solving Eq. (\ref{eq:temporal_training_objective}). Additionally, Eq. (\ref{eq:final_temporal_obj}) is differentiable with respect to $\{w_1, \ldots, w_{T_0}\}$, allowing for the end-to-end training of $\pi$. In this paper, we fix $M\!=\!128$, which works reasonably well in various scenarios.

Similar to us, OCSampler \cite{lin2022ocsampler} also formulates frame selection as a video-conditional sequential weighted sampling problem. Our contribution over \cite{lin2022ocsampler} lies in developing a novel solving algorithm for this basic formulation. We propose to consider the expected loss over dynamic frame sampling as the training objective, and reveal that it can be decomposed into a differentiable form solved with the Monte Carlo method, yielding an efficient end-to-end optimization procedure. See Appendix \ref{app:additional_discussions} for more comparisons.

\textbf{Implementation.}
Our implementation of dynamic frame sampling minimizes Eq. (\ref{eq:final_temporal_obj}) without notable additional costs. See Appendix \ref{app:implementation_dynamic_frame_sampling} for more details (due to spatial limitations).

\textbf{Unified spatial-temporal dynamic computation.}
Notably, dynamic frame sampling is fully compatible with the previously described spatial-wise dynamic computation mechanism. For example, one may train $\pi$ to identify the informative frame patches for the $T_{\textnormal{G}}$ frames processed by $f_{\textnormal{G}}$ in accordance with (\ref{eq:learning_deformable_patches}), and interpolate between the outputs of $\pi$ of adjacent frames to acquire a deformable task-relevant patch for each of the $T_{\textnormal{L}}$ selected frames. In summary, the overall training objective of our method can be written as 
\begin{equation}
    \label{eq:uni_final_objective}
        \mathop{\textnormal{minimize}}_{f_{\textnormal{G}}, f_{\textnormal{L}}, f_{\textnormal{C}}, \pi}\ \ 
        \mathcal{L}_{f_{\textnormal{G}}, f_{\textnormal{L}}, f_{\textnormal{C}}} + \mathcal{L}_{\pi}^{\textnormal{spatial}} + \mathcal{L}_{\pi}^{\textnormal{temporal}}.
\end{equation}
Here $\mathcal{L}_{f_{\textnormal{G}}, f_{\textnormal{L}}, f_{\textnormal{C}}}$ represents the loss introduced in (\ref{eq:final_objective}), which is only leveraged to train the two encoders $f_{\textnormal{G}}$ and $f_{\textnormal{L}}$, and the classifier $f_{\textnormal{C}}$. The policy network $\pi$ is trained to perform spatial-temporal dynamic computation simultaneously by minimizing the combination of (\ref{eq:learning_deformable_patches}) and (\ref{eq:final_temporal_obj}).

% the combination of $\mathcal{L}_{\pi}^{\textnormal{spatial}}$ and $\mathcal{L}_{\pi}^{\textnormal{temporal}}$, as elaborated upon in (\ref{eq:learning_deformable_patches}) and (\ref{eq:final_temporal_obj}).

% In this context, $\mathcal{L}{f{\textnormal{G}}, f_{\textnormal{L}}, f_{\textnormal{C}}}$ represents the loss introduced in (\ref{eq:final_objective}) and is exclusively employed for training the two encoders, $f_{\textnormal{G}}$ and $f_{\textnormal{L}}$, as well as the classifier $f_{\textnormal{C}}$. Concurrently, the policy network $\pi$ is trained to conduct spatial and temporal dynamic computation by minimizing $\mathcal{L}{\pi}^{\textnormal{spatial}}$ and $\mathcal{L}{\pi}^{\textnormal{temporal}}$, respectively, as elaborated upon in (\ref{eq:learning_deformable_patches}) and (\ref{eq:temporal_training_objective}).

% Cumulative Distribution Function
% To realize this, we first approximate 
% and we will show in the following that it can be obtained with negligible additional cost in our AdaFocus framework.
% under the goal of 
% that are relatively more informative in terms of the recognition task
% As a common practice of video understanding \cite{wang2016temporal, lin2019tsm, wu2019adaframe, gao2020listen, feichtenhofer2020x3d, meng2020ar, sun2021dynamic, lin2022ocsampler}, we typically sample a pre-defined number of frames uniformly (\emph{e.g.}, 8 or 16) from each video for the model to process.

\subsection{Reducing Sample-wise Redundancy -- Conditional-exit}
\label{sec:early_exit}

% The basic formulation of AdaFocus processes each frame using the same amount of computation, and hence it can be improved by further reducing temporal redundancy. AdaFocusV1 achieves this via skipping less informative frames with reinforcement learning. In contrast, we propose a simple confidence-based early-exit algorithm that achieves competitive performance. Our approach can be directly deployed on AdaFocusV2 trained following the aforementioned paradigm, without any additional training process. We refer to this extension of AdaFocusV2 as AdaFocusV2+, as shown in Figure \ref{fig:AdaFocusV2_plus}. 

% In addition to the previously discussed spatial and temporal redundancies, another significant source of redundant computation arises from the uniform treatment of diverse samples concerning their computational cost. Indeed, numerous studies \cite{huang2017multi, yang2020resolution, wang2020glance, ghodrati2021frameexit, wang2021not, li2021ds} have identified a substantial number of "simpler" samples within datasets, which can be accurately classified with considerably reduced computation relative to other samples. To model this sample-wise redundancy, we propose a straightforward confidence-based early-exit algorithm. This approach can be seamlessly integrated into models trained using the previously outlined paradigm, obviating the need for an additional training phase.

Apart from the aforementioned spatial and temporal redundancies, an additional factor contributing to a significant amount of redundant computation can be attributed to the equivalent treatment of diverse samples concerning their computational cost. In fact, numerous studies \cite{huang2017multi, yang2020resolution, wang2020glance, ghodrati2021frameexit, wang2021not} have reported the existence of a significant number of ``easier'' samples within datasets, which can be accurately recognized with considerably less computation compared to other samples. We propose to model this sample-wise redundancy through a straightforward early-exit algorithm. This approach can be seamlessly integrated into the models trained using the previously outlined paradigm, obviating the need for an additional training phase.

% A visual representation of this concept is provided in Figure \ref{fig:AdaFocusV2_plus}.
% As a matter of fact, it has been widely observed that there exist a considerable number of ``easier'' samples in datasets \cite{huang2017multi, yang2020resolution, wang2020glance, ghodrati2021frameexit, wang2021not, li2021ds}, which can be accurately recognized with much smaller computation than others. We propose to model this sample-wise redundancy with a simple confidence-based early-exit algorithm. Our approach can be directly deployed for the model trained following the aforementioned paradigm, without any additional training process. An illustration is shown in Figure \ref{fig:AdaFocusV2_plus}. 

% To implement this idea, at test time, we propose to compare the largest entry of the softmax prediction $\bm{p}_t$ ({defined as confidence in previous works \cite{huang2017multi, yang2020resolution, wang2020glance, wang2021not}.}) at $t^{\textnormal{th}}$ frame with a pre-defined threshold $\eta_t$. Once $\max_j p_{tj} \geq \eta_t$, the prediction will be postulated to be reliable enough, and the inference will be terminated by outputting $\bm{p}_t$. 

\begin{figure}[!t]
    % \vskip -0.1in
    \begin{center}
    \centerline{\includegraphics[width=0.925\columnwidth]{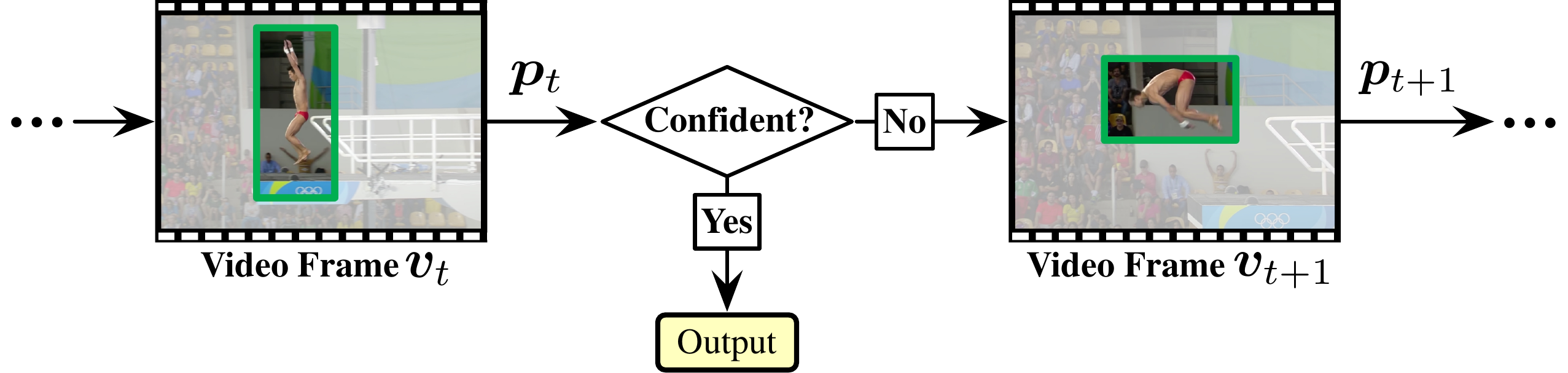}}
    \vskip -0.15in
    \caption{\textbf{Illustration of the conditional-exit algorithm.}  \label{fig:AdaFocusV2_plus}
    }
    \end{center}
    \vspace{-6.1ex}
\end{figure}

In the context of videos, we assume that activating the high-capacity local encoder $f_{\textnormal{L}}$ to process a subset of frames rather than all frames may be sufficient for the ``easier'' samples. To implement this idea, at test time, we propose to compare the entropy of the softmax prediction $\bm{p}_t$ at $t^{\textnormal{th}}$ frame (\emph{i.e.}, $-\sum\nolimits_{j}p_{tj}\log p_{tj}$) with a pre-defined threshold $\eta_t$. The inference will be terminated with $\bm{p}_t$ if $-\sum\nolimits_{j}p_{tj}\log p_{tj} \leq \eta_t$. We always adopt an infinite-threshold at the final frame. The concept is illustrated in Figure \ref{fig:AdaFocusV2_plus}. The values of $\{\eta_1, \eta_2, \ldots\}$ are solved on the validation set. Suppose that the model needs to classify a set of samples $\mathcal{D}_{\textnormal{val}}$ within a given computational budget $B>0$ \cite{huang2017multi, wang2021not}. One can obtain the thresholds through
\begin{equation}
    \label{eq:thres}
    \begin{split}
    \mathop{\textnormal{maximize}}_{\eta_1, \eta_2, \ldots}&\ \ \ \textnormal{Acc}(\eta_1, \eta_2, \ldots|\mathcal{D}_{\textnormal{val}})\\ 
    \textnormal{subject to}&\ \ \ \textnormal{FLOPs}(\eta_1, \eta_2, \ldots|\mathcal{D}_{\textnormal{val}})\leq B.
    \end{split}
\end{equation}

Here $\textnormal{Acc}(\eta_1, \eta_2, \ldots|\mathcal{D}_{\textnormal{val}})$ and $\textnormal{FLOPs}(\eta_1, \eta_2, \ldots|\mathcal{D}_{\textnormal{val}})$ refer to the accuracy and computational cost on $\mathcal{D}_{\textnormal{val}}$ using the thresholds $\{\eta_1, \eta_2, \ldots\}$. Notably, by changing $B$, one can obtain varying values of $\{\eta_1, \eta_2, \ldots\}$. The computational cost of our method can be flexibly adjusted without additional training by simply adjusting these thresholds. In our implementation, we solve problem (\ref{eq:thres}) following the method proposed in \cite{huang2017multi} on the training set, which we find performs on par with using cross-validation.

\begin{table*}[!t]
  \centering
  \begin{footnotesize}
  \caption{\textbf{Uni-AdaFocus v.s. state-of-the-art efficient video understanding approaches on ActivityNet, FCVID and Mini-Kinetics (\emph{i.e.}, Kinetics-200).} ``GFLOPs'' refers to the average computational cost for processing a single video. MN2/RN/ENB0 represents MobileNet-V2/ResNet/EfficientNet-B0. The best results are \textbf{bold-faced}. The \textcolor{blue}{blue} texts highlight the comparisons with the \underline{underlined} baselines.}
  \label{tab:actnet_main_table}
  \vskip -0.225in
  \setlength{\tabcolsep}{1mm}{
  \vspace{5pt}
  \renewcommand\arraystretch{1.075}
  \resizebox{1.75\columnwidth}{!}{
  \begin{tabular}{c|c|c|cccccc}
  \toprule
  \multirow{2}{*}{Methods} & \multirow{2}{*}{Published on}& \multirow{2}{*}{Backbones}  & \multicolumn{2}{c}{ActivityNet} &  \multicolumn{2}{c}{FCVID} & \multicolumn{2}{c}{Mini-Kinetics}  \\
  &&& \ mAP & GFLOPs & \ mAP &  GFLOPs & \ Top-1 Acc. &  GFLOPs \\
  % \midrule
  \midrule
  FrameGlimpse \cite{yeung2016end} & \emph{CVPR'16} & VGG-16        &  60.1\%  & 33.3   &  67.6\%  & 30.1    & -- & --  \\
  FastForward \cite{fan2018watching} & \emph{IJCAI'18} & Inception-V3 &  54.6\%  & {17.9}    &  71.2\%  & 66.1    & -- & --  \\
  LiteEval \cite{wu2019liteeval} & \emph{NeurIPS'19} & MN2+RN        &  72.7\%  & 95.1    &  80.0\% & 94.3     &  61.0\%  & 99.0  \\
  SCSampler \cite{korbar2019scsampler} & \emph{ICCV'19} & MN2+RN     &  72.9\%  & 42.0    &  81.0\% & 42.0     &  70.8\%  & 42.0  \\
  ListenToLook \cite{gao2020listen} & \emph{CVPR'20} & MN2+RN        &  72.3\%  & 81.4    & -- & --               & -- & --  \\
  AR-Net \cite{meng2020ar} & \emph{ECCV'20} & MN2+RN                 &  73.8\%  & 33.5    &  81.3\% & 35.1     &  71.7\%  & 32.0  \\
  AdaFrame \cite{wu2019adaframe} & \emph{T-PAMI'21} & MN2+RN         &  71.5\%  & 79.0    &  80.2\% & 75.1     & -- & --  \\
  AdaFuse \cite{meng2021adafuse} & \emph{ICLR'21} & RN                &  73.1\%  & 61.4    &  81.6\% & 45.0     &  72.3\%  & 23.0  \\
  FrameExit \cite{ghodrati2021frameexit} & \emph{CVPR'21} & RN        &  76.1\%  & 26.1    & -- & --               &  72.8\%  & 19.7  \\
  VideoIQ \cite{sun2021dynamic} & \emph{ICCV'21} & MN2+RN            &  74.8\%  & 28.1    &  82.7\% & 27.0     &  72.3\%  & 20.4  \\
  Dynamic-STE \cite{kim2021efficient} & \emph{ICCV'21} & RN           &  75.9\%  & 30.5    & -- & --               &  72.7\%  & 18.3  \\
  AdaMML \cite{panda2021adamml} & \emph{ICCV'21} & MN2+RN            &  73.9\%  & 94.0    &  {85.8\%}  & 93.9    & -- & --  \\
  SMART \cite{gowda2021smart}  & \emph{AAAI'21} & MN2+RN              & -- & --               &  82.1\% & --       & -- & --  \\
  OCSampler \cite{lin2022ocsampler}  & \emph{CVPR'22} & TSM-MN2+RN &  \underline{76.9\%} & \underline{21.7} &  {82.7\%} & {26.8}  &  72.9\% &  {17.5} \\
  AFNet \cite{zhang2022look}  & \emph{NeurIPS'22} & RN                &  75.6\% & 24.6     & -- & --               & 73.5\%  & 22.0 \\
  NSNet \cite{xia2022nsnet}  & \emph{ECCV'22} & MN2+RN               & {76.8\%} &  {26.0}      & \underline{83.9\%} & \underline{26.0}      & \underline{73.6\%} & \underline{18.1} \\
  TSQNet \cite{xia2022temporal}  & \emph{ECCV'22} & MN2+ENB0+RN           & 76.6\% &  26.1      & 83.5\% & 26.2      & 73.2\% & 19.7 \\
  \midrule
  \multirow{2}{*}{\shortstack{Uni-AdaFocus (128$^2$)\\[-0.7ex]w/o sample-wise dynamic}} & \multirow{2}{*}{--} & \multirow{2}{*}{MN2+RN}    &  \multirow{2}{*}{ \textbf{80.7\%}} &  \multirow{2}{*}{27.2}  & \multirow{2}{*}{ \textbf{86.4\%}} &  \multirow{2}{*}{27.2}  &   \multirow{2}{*}{ \textbf{75.8\%}} &  \multirow{2}{*}{27.2}   \\
    &  &       &&&&&& \\
  \midrule
  \multirow{2}{*}{Uni-AdaFocus (128$^2$)}  & \multirow{2}{*}{--} & \multirow{2}{*}{MN2+RN}    &  \ \ {80.4\%}\textcolor{blue}{$_{\uparrow 3.5\%}$} & 18.1 &  \ \ 86.2\%\textcolor{blue}{$_{\uparrow 2.3\%}$} & 18.1 &  \ \ 75.7\%\textcolor{blue}{$_{\uparrow 2.1\%}$}  & 18.1  \\
  &&&  \ \ {78.3\%}\textcolor{blue}{$_{\uparrow 1.4\%}$} & \textbf{9.1}\textcolor{blue}{$_{\downarrow 2.4\!\times}$} &  \ \ 85.4\%\textcolor{blue}{$_{\uparrow 1.5\%}$} &  \textbf{9.1}\textcolor{blue}{$_{\downarrow 2.9\!\times}$}  &  \ \ 74.3\%\textcolor{blue}{$_{\uparrow 0.7\%}$} &  \textbf{9.1}\textcolor{blue}{$_{\downarrow 2.0\!\times}$}  \\
  \bottomrule
  \end{tabular}}}
  \end{footnotesize}
  % \vskip -0.09in
  \vskip -0.15in
  % \vskip -0.1in
\end{table*}

% 17.887953920, 79.300
% 27.229979648, 80.665
% 39.241155584, 81.505

% \vspace{-0.3ex}
\section{Experiments}
\label{sec:experiment}

\textbf{Overview.}
This section presents extensive empirical results to validate the effectiveness of Uni-AdaFocus. In Section \ref{sec:uni_ada_mnv2_r50}, we compare Uni-AdaFocus with state-of-the-art efficient video understanding frameworks under the standard experimental settings in most of the literature. In Section \ref{sec:tsm_uniada}, we deploy Uni-AdaFocus on top of representative recently proposed lightweight deep networks (\emph{i.e.}, TSM and X3D), and demonstrate that our method is able to effectively improve the computational efficiency of these cutting-edge models. In Section \ref{sec:real_world_app}, we further evaluate Uni-AdaFocus in the context of three representative real-world application scenarios. In Section \ref{sec:abl}, comprehensive analytical results and visualizations are provided to give additional insights into our method.
Besides, in both Sections \ref{sec:uni_ada_mnv2_r50} \& \ref{sec:tsm_uniada}, we present the comparisons of Uni-AdaFocus v.s. AdaFocus V1/V2/V3.

\textbf{Datasets.}
Seven large-scale video understanding benchmark datasets are considered, \emph{i.e.}, ActivityNet \cite{caba2015activitynet}, FCVID \cite{TPAMI-fcvid}, Mini-Kinetics \cite{kay2017kinetics, meng2020ar}, Something-Something (Sth-Sth) V1\&V2 \cite{goyal2017something}, Jester \cite{materzynska2019jester} and Kinetics-400 \cite{kay2017kinetics}. We also consider three real-world application scenarios \emph{i.e.}, fine-grained diving action classification on Diving48 \cite{li2018resound}, Alzheimer's and Parkinson's diseases diagnosis with brain magnetic resonance images (MRI) on ADNI \cite{url_1}, OASIS \cite{url_2}, and PPMI \cite{url_3}, and violence recognition for online videos on RLVS \cite{soliman2019violence}. For data pre-processing, we adopt the same pipeline as \cite{lin2019tsm, meng2020ar, Wang_2021_ICCV}. More details are deferred to Appendix \ref{app:dataset}.

\begin{figure*}[!t]
  % \vskip -0.1in
  \begin{center}
  \centerline{\includegraphics[width=0.786\linewidth]{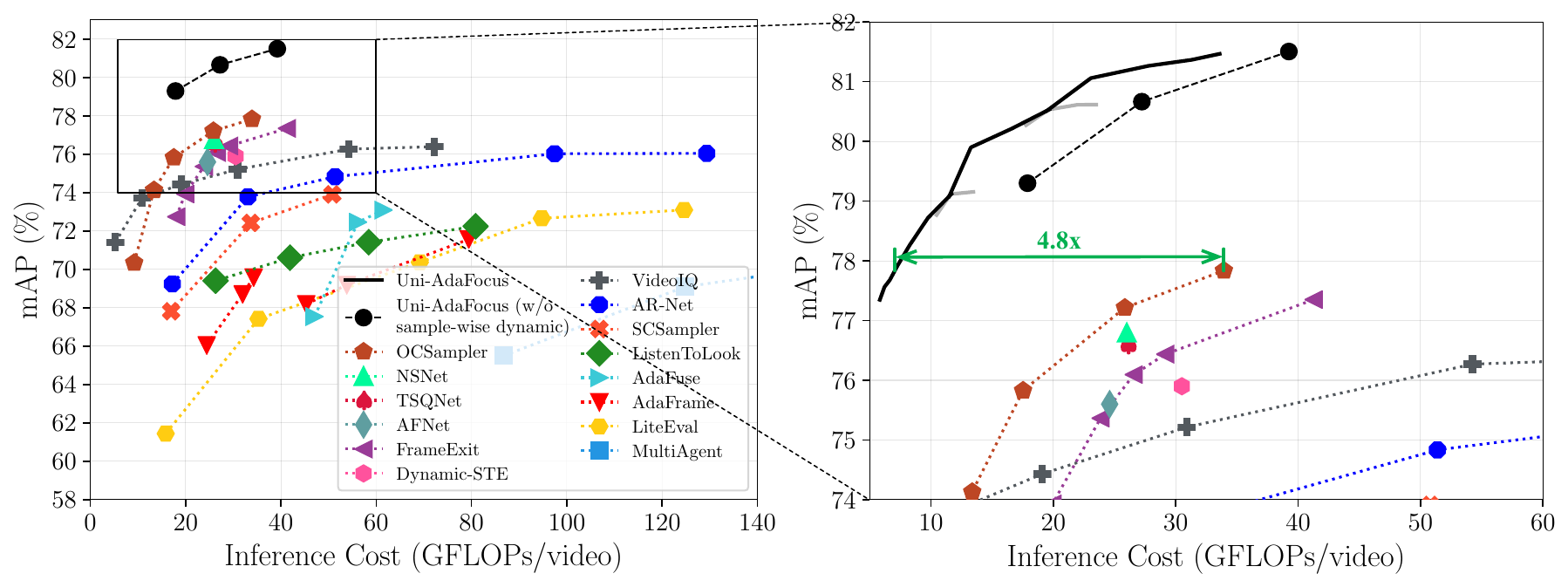}}
  \vskip -0.175in
  \caption{\textbf{Comparisons of Uni-AdaFocus and state-of-the-art efficient video understanding approaches on ActivityNet in terms of inference efficiency.}  Our method adopts $P^2\in$ \{96$^2$, 128$^2$, 160$^2$\}, corresponding to the three black curves. Notably, the inference cost of Uni-AdaFocus can switch within each black curve without additional training (by modifying the conditional-exit criterion, see: Section \ref{sec:early_exit}). \label{fig:actnet_vs_sota}}
  \end{center}
  \vspace{-2.5ex}
\end{figure*}

\begin{figure*}[b]
  % \vskip -0.1in
  \vspace{-2.75ex}
  \begin{center}
  \centerline{\includegraphics[width=1.0\linewidth]{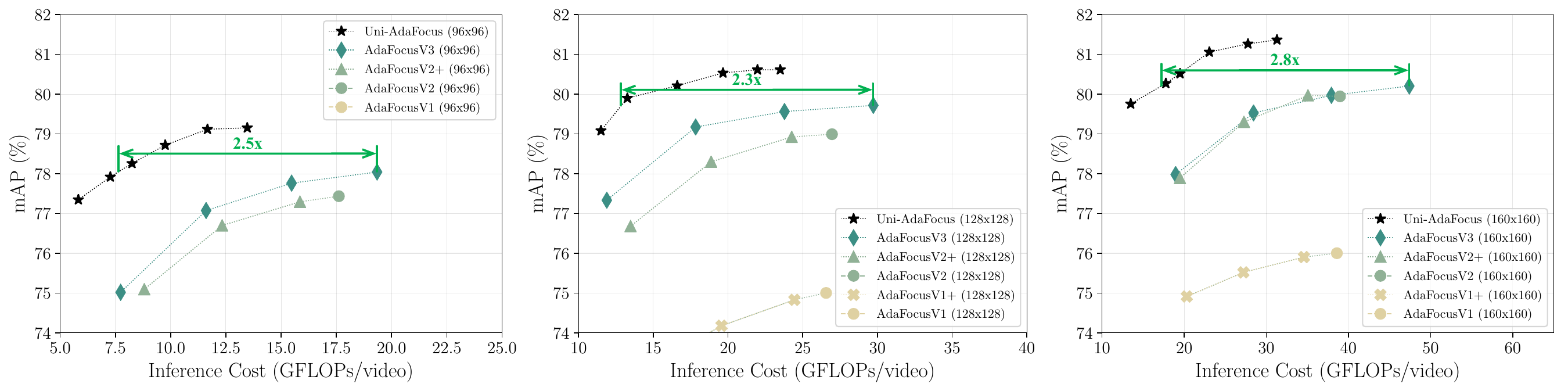}}
  \vskip -0.175in
  \caption{\textbf{Comparisons between Uni-AdaFocus and the preliminary versions of AdaFocus on ActivityNet in terms of inference efficiency.}  \label{fig:actnet_vs_v1_eval}}
  \end{center}
  \vspace{-2.5ex}
\end{figure*}

\subsection{Comparisons with State-of-the-art Efficient Video Understanding Approaches}
\label{sec:uni_ada_mnv2_r50}

\textbf{Setups.}
In this subsection, we comprehensively compare Uni-AdaFocus with state-of-the-art efficient video understanding frameworks on ActivityNet, FCVID and Mini-Kinetics. Our baselines include FrameGlimpse \cite{yeung2016end}, FastForward \cite{fan2018watching}, MultiAgent \cite{wu2019multi}, LiteEval \cite{wu2019liteeval}, SCSampler \cite{korbar2019scsampler}, ListenToLook \cite{gao2020listen}, AR-Net \cite{meng2020ar}, AdaFrame \cite{wu2019adaframe}, AdaFuse \cite{meng2021adafuse}, VideoIQ \cite{sun2021dynamic}, Dynamic-STE \cite{kim2021efficient}, AdaMML \cite{panda2021adamml}, SMART \cite{gowda2021smart}, FrameExit \cite{ghodrati2021frameexit}, OCSampler \cite{lin2022ocsampler}, NSNet \cite{xia2022nsnet}, TSQNet \cite{xia2022temporal}, AFNet \cite{zhang2022look}, and the preliminary versions of AdaFocus \cite{Wang_2021_ICCV, wang2021adafocus, wang2022adafocusv3}. Following the common practice of these baselines, we adopt MobileNet-V2 \cite{sandler2018mobilenetv2} and ResNet-50 \cite{he2016deep} as the global encoder $f_{\textnormal{G}}$ and local encoder $f_{\textnormal{L}}$ in Uni-AdaFocus. We uniformly sample $T_0\!=\!48$ frames from each video, and set $T_{\textnormal{G}}\!=\!T_{\textnormal{L}}\!=\!16$. In addition, note that Uni-AdaFocus resizes the selected informative patches with varying scales and shapes to a common size $P^2$, under the goal of processing these patches efficiently on hardwares like GPUs (see Section \ref{sec:improved_spatial_sub2} and Figure \ref{fig:deformable_patch} for details). We consider $P^2\in$ \{96$^2$, 128$^2$, 160$^2$\}. More implementation details can be found in Appendix \ref{app:implementation_details}.

\textbf{Comparisons with state-of-the-art baselines}
on ActivityNet, FCVID and mini-Kinetics are summarized in Table \ref{tab:actnet_main_table}. It is clear that Uni-AdaFocus (128$^2$) outperforms all the competitive efficient video understanding methods by large margins. For example, on ActivityNet, it achieves 3.5\% higher mean average precision (mAP) (80.4\% v.s. 76.9\%) than the strongest baseline, OCSampler \cite{lin2022ocsampler}, with smaller GFLOPs (18.1 v.s. 21.7). On FCVID and Mini-Kinetics, similar observations can be consistently obtained, \emph{i.e.}, Uni-AdaFocus outperforms the recently proposed NSNet \cite{xia2022nsnet} by 2.3\% and 2.1\% with similar or less computation. In Figure \ref{fig:actnet_vs_sota}, we further present the variants of the baselines and Uni-AdaFocus with different computational costs for a more comprehensive comparison. It can be observed that our method leads to a considerably better efficiency-accuracy trade-off. When achieving the same state-of-the-art level mAP, the number of the required GFLOPs/video for Uni-AdaFocus is approximately 4.8$\times$ less than OCSampler \cite{lin2022ocsampler}.

\begin{table}[t]
  \vskip -0.175in
  \centering
  \begin{footnotesize}
  \caption{\textbf{Comparisons of Uni-AdaFocus and AdaFocusV1/V2 in terms of training efficiency.} The mAP and wall-clock training time on ActivityNet are reported. The latter is obtained using 4 NVIDIA 3090 GPUs. The best results are \textbf{bold-faced}. E2E refers to End-to-End. $^\dagger$For fair comparison, here we do not perform sample-wise dynamic computation.}
  \label{tab:actnet_vs_v1_train}
  \vskip -0.22in
  \setlength{\tabcolsep}{0.8mm}{
  \vspace{5pt}
  \renewcommand\arraystretch{1.05}
  \resizebox{0.7\columnwidth}{!}{
  \begin{tabular}{ccccccc}
  \toprule
  & \multirow{2}{*}{Methods} &  E2E  & \multicolumn{3}{c}{$P^2$}\\
  && Training & 96$^2$ & 128$^2$ & 160$^2$ \\
  \midrule
  \multirow{3}{*}{\shortstack{mAP on\\ ActivityNet}} & AdaFocusV1 & \textcolor{red}{\xmark} & \ 71.9\% & 75.0\% & 76.0\% \\
  & AdaFocusV2 & \textcolor{_green}{\cmark} & \ {77.4\%} & {79.0\%} & {79.9\%} \\
  & Uni-AdaFocus$^\dagger$ &  \textcolor{_green}{\cmark} &\ \textbf{79.3\%} & \textbf{80.7\%} & \textbf{81.5\%}\\
  \midrule
  \multirow{3}{*}{\shortstack{Training\\ Wall-time}} & AdaFocusV1 & \textcolor{red}{\xmark} &\  6.4h & 7.2h & 8.6h \\
  & AdaFocusV2 & \textcolor{_green}{\cmark} &\  {3.4h} & {3.7h} & {4.3h}  \\
  & Uni-AdaFocus &  \textcolor{_green}{\cmark} &\ {3.6h} & {3.8h} & {4.1h}  \\
  \bottomrule
  \end{tabular}}}
  \end{footnotesize}
\vspace{-3.5ex}
\end{table}

\textbf{Effectiveness of sample-wise dynamic computation.}
Table \ref{tab:actnet_main_table} and Figure \ref{fig:actnet_vs_sota} evaluate the performance of Uni-AdaFocus both with and without the conditional-exit mechanism in Section \ref{sec:early_exit}. When this mechanism is activated, we vary the average computational budget, solve early-exit thresholds, and evaluate the corresponding mAP or Top-1 accuracy, as stated in Section \ref{sec:early_exit}. One can observe that sample-wise adaptive computation effectively improves the inference efficiency. Moreover, this mechanism enables Uni-AdaFocus to adjust its inference cost online without additional training (by simply changing the early-exit thresholds).

\textbf{Comparisons with the preliminary versions of AdaFocus.}
The comparisons of Uni-AdaFocus and AdaFocusV1/V2 in terms of mAP v.s. training/inference cost are presented in Table \ref{tab:actnet_vs_v1_train} and Figure \ref{fig:actnet_vs_v1_eval}, respectively. One can observe that both Uni-AdaFocus and AdaFocusV2 can be trained efficiently in an end-to-end fashion, which reduces the time consumption for training by $\sim\!2\times$ compared to the three-stage training algorithm of AdaFocusV1, while Uni-AdaFocus improves mAP by 1.6-1.9\% on top of AdaFocusV2.
Moreover, it can be observed from Figure \ref{fig:actnet_vs_v1_eval} that Uni-AdaFocus reduces the inference cost by 2.3-2.8x compared to AdaFocusV3 \cite{wang2022adafocusv3} without sacrificing the performance.

\begin{table*}[!t]
  \centering
  \begin{footnotesize}
  \caption{\textbf{Uni-AdaFocus-TSM v.s. representative efficient video understanding models on Sth-Sth V1\&V2 and Jester}. MN2, R18/R34/R50 and BN-Inc. denote MobileNet-V2, ResNet-18/34/50 and BN-Inception. TSM+ refers to the augmented TSM baseline with the same network architecture as our method except for the policy network $\pi$. The throughput is benchmarked on an NVIDIA 3090 GPU. The best results are \textbf{bold-faced}. The \textcolor{blue}{blue} texts highlight the comparisons with the \underline{underlined} baselines. $^\dagger$120-epoch training, which is 2.4x longer than our method and the baselines.}
  \label{tab:sthsth}
  \vskip -0.225in
  \setlength{\tabcolsep}{0mm}{
  \vspace{5pt}
  \renewcommand\arraystretch{1.075}
  \resizebox{1.9\columnwidth}{!}{
  \begin{tabular}{cccccccccc} 
  \toprule
  \multirow{2}{*}{{Method}} & \multirow{2}{*}{{Backbones}}  & \multirow{2}{*}{{\#Frames}}  & \multicolumn{2}{c}{{Sth-Sth V1}}  & \multicolumn{2}{c}{{Sth-Sth V2}} & \multicolumn{2}{c}{{Jester}} & Throughput\\ %\multicolumn{2}{c}{Practical Efficiency}\\
  % &&&&&&&& \\
  % \vspace{-0.1em}
  &&& \footnotesize{{\ \ Top-1 Acc.}} & \footnotesize{{GFLOPs}} &  \footnotesize{{Top-1 Acc.}} & \footnotesize{{GFLOPs}}  &  \footnotesize{{Top-1 Acc.}} & \footnotesize{{GFLOPs}}  & \scriptsize{(NVIDIA 3090 GPU)}\\  
  % \midrule
  \midrule
  I3D \cite{carreira2017quo} & 3D-R50 & 32$\times$2  & 41.6\% &306 & - & - & - &  - & -\\
  I3D+GCN+NL \cite{wang2018videos} & 3D-R50 & 32$\times$2 & 46.1\% & 606 & - & - & - &  - & -\\
  ECO\textsubscript{En}Lite \cite{zolfaghari2018eco} & BN-Inc.+3D-R18 & 92  & {46.4\%} & 267 & - & - & - &  - & - \\
  % \midrule
  % \hdashline\\[-2.3ex]
  TSN \cite{wang2016temporal} & R50 & 8 & 19.7\% & 33.2 &27.8\% &  33.2   & 82.6\% &  33.2 & - \\ 
  AR-Net \cite{meng2020ar} & MN2+R18/34/50 & 8 & 18.9\% & 41.4 & -  &  -  & 87.8\% &  21.2 & - \\ 
  TRN\textsubscript{RGB/Flow} \cite{zhou2018temporal} & BN-Inc. & 8/8  & 42.0\% & 32.0 & 55.5\% & 32.0  & - & - & - \\
  ECO \cite{zolfaghari2018eco} & BN-Inc.+3D-R18 & 8 & 39.6\% & 32.0 & - & - & - &  - & -\\
  SmallBig \cite{li2020smallbignet} & R50 & 8 & 47.0\% & 52.0 &  59.7\% & 52.0 & - &  - & -\\
  bLVNet-TAM \cite{fan2019more} & bL-R50 & 16 & 47.8\% & 35.1 & 60.2\% & 35.1 & - &  - & -\\
  SlowFast \cite{feichtenhofer2019slowfast} & R50+R50 & 8$\times$8 & - & - &  61.7\% & 66.6$\times$3 & -&  - & -  \\
  TANet \cite{liu2021tam} & R50 & 8 & 47.3\% & 33.0 &  60.5\% & 33.0  & - &  - & - \\
  TEINet \cite{liu2020teinet} & R50 & 8 & 47.4\% & 33.0 &  61.3\% & 33.0  & - &  - & - \\
  STM \cite{jiang2019stm} & R50 & 8 & 47.5\% & 33.3 &  - & -   & -&  - & - \\
  TEA \cite{li2020tea} & R50 & 8 & 48.9\% & 35.0 &  60.9\% & 35.0   & -&  - & - \\
  \midrule
  TSM \cite{lin2019tsm}  & R50 & 8 & 46.1\% & 32.7 &  59.1\% & 32.7 & 96.0\% & 32.7 &  - \\
  AdaFuse-TSM \cite{meng2021adafuse} & R50 & 8 & 46.8\% & 31.5 & 59.8\% & 31.3   & -&  - & - \\
  % \hdashline\\[-2.3ex]
  % \midrule
  TSM+ \cite{lin2019tsm}  & MN2+R50 & 8+8  & \underline{47.0\%} & \underline{35.1} & \underline{59.6\%} & \underline{35.1} & \underline{96.2\%} & \underline{35.1} &  \ \ \ \underline{ 218.2 Videos/s}   \\
  S2DNet-TSM \cite{liang2022delving}$^\dagger$  & MN2+R50 & 8+12  & 49.7\% & 22.0 & 62.5\% & 22.0 & - & - &  -   \\

  \multirow{2}{*}{\shortstack{Uni-AdaFocus-TSM (96$^2$)\\[-0.7ex]w/o sample-wise dynamic}}  & MN2+R50 & 8+8 & 47.5\% \scriptsize{(\textcolor{blue}{$\uparrow$0.5\%})} & \ \textbf{8.8} \scriptsize{(\textcolor{blue}{$\downarrow$3.99x})} & 60.8\% \scriptsize{(\textcolor{blue}{$\uparrow$1.2\%})}  & \ \ \ \textbf{8.8} \scriptsize{(\textcolor{blue}{$\downarrow$3.99x})} &  - & -  &  \ \ \ \textbf{ 569.4 Videos/s} \scriptsize{(\textcolor{blue}{$\uparrow$2.61x})}\ \  \\

  & MN2+R50 & 8+12 & 49.5\% & {11.8} & 62.6\%  & {11.8}  &  - & -  &  \ \ \ { 474.8 Videos/s}   \\

   \multirow{2}{*}{\shortstack{Uni-AdaFocus-TSM (128$^2$)\\[-0.7ex]w/o sample-wise dynamic}}  & MN2+R50 & 8+12 & 50.5\% & {18.8}& 63.2\%  & {18.8} &  \textbf{97.1\%} \scriptsize{(\textcolor{blue}{$\uparrow$0.9\%})} & \ \ \ \ \textbf{18.8} \scriptsize{(\textcolor{blue}{$\downarrow$1.87x})}  &  \ \ \ { 366.4 Videos/s}   \\

  & MN2+R50 & 8+16 & \textbf{51.5\%} \scriptsize{(\textcolor{blue}{$\uparrow$4.5\%})}  &  \ \ \ {24.2} \scriptsize{(\textcolor{blue}{$\downarrow$1.45x})}\ \ \ \  & \textbf{64.2\%} \scriptsize{(\textcolor{blue}{$\uparrow$4.6\%})} & \ \ \ 24.2 \scriptsize{(\textcolor{blue}{$\downarrow$1.45x})} &  - & -  &  \ \ \ { 305.4 Videos/s} \scriptsize{(\textcolor{blue}{$\uparrow$1.40x})}  \\

  \midrule

  Uni-AdaFocus-TSM (96$^2$)  & MN2+R50 & 8+12 & 48.9\% \scriptsize{(\textcolor{blue}{$\uparrow$1.9\%})} & \ \textbf{8.8} \scriptsize{(\textcolor{blue}{$\downarrow$3.99x})} & 62.5\% \scriptsize{(\textcolor{blue}{$\uparrow$2.9\%})}  & \ \ \ \textbf{8.8} \scriptsize{(\textcolor{blue}{$\downarrow$3.99x})} &  - & -  &  \ \ \ { 569.2 Videos/s} \scriptsize{(\textcolor{blue}{$\uparrow$2.61x})}\ \  \\

  Uni-AdaFocus-TSM (128$^2$)  & MN2+R50 & 8+16 & 51.0\% \scriptsize{(\textcolor{blue}{$\uparrow$4.0\%})} & \ {18.8} \scriptsize{(\textcolor{blue}{$\downarrow$1.87x})} & \textbf{64.2\%} \scriptsize{(\textcolor{blue}{$\uparrow$4.6\%})}  & \ \ \ {18.8} \scriptsize{(\textcolor{blue}{$\downarrow$1.87x})} &  - & -  &  \ \ \ { 366.8 Videos/s} \scriptsize{(\textcolor{blue}{$\uparrow$1.68x})}\ \  \\

  \bottomrule
  \end{tabular}}}
  \end{footnotesize}
  \vskip -0.125in
\end{table*}

\begin{figure*}[!t]
  % \vskip -0.1in
  \begin{center}
  \centerline{\includegraphics[width=0.73\linewidth]{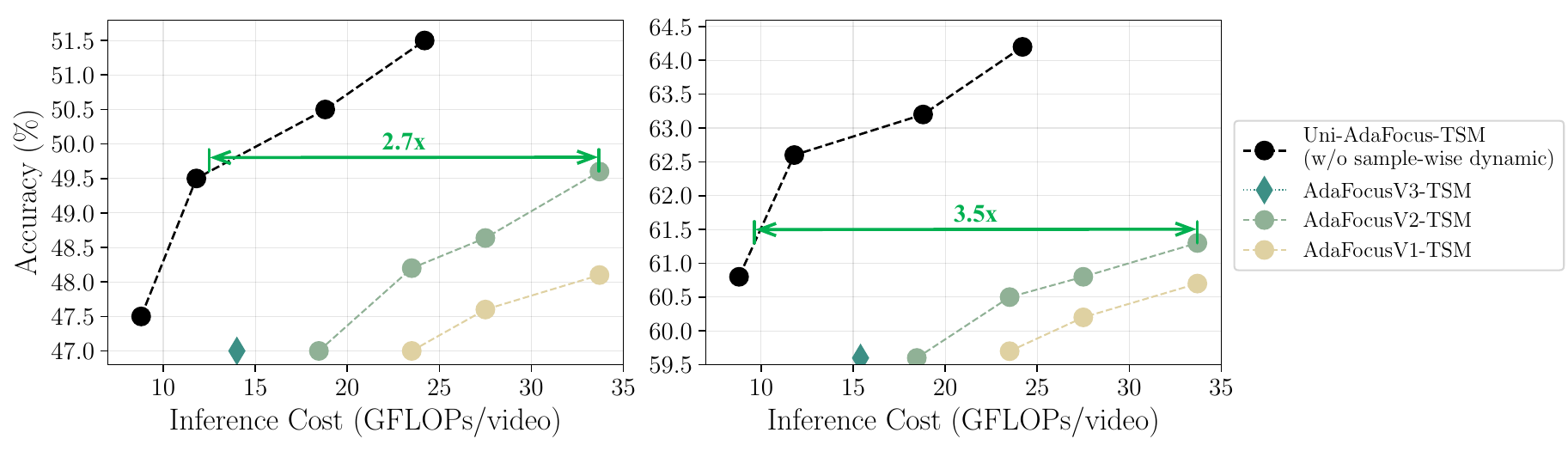}}
  \vskip -0.2in
  \caption{\textbf{Uni-AdaFocus v.s. the preliminary versions of AdaFocus on Sth-Sth V1 (\emph{left}) and V2 (\emph{right}) in terms of inference efficiency.} For fair comparison, we do not perform sample-wise dynamic computation for Uni-AdaFocus. \label{fig:sth_flops_acc}
  }
  \end{center}
  \vspace{-4.6ex}
\end{figure*}

\vspace{-0.5ex}
\subsection{Building on Top of Existing Efficient Backbones}
\label{sec:tsm_uniada}

\textbf{Setups.}
In this subsection, we implement Uni-AdaFocus on top of the recently proposed efficient network architectures, ConvNets with temporal shift module (TSM) \cite{lin2019tsm} and X3D networks \cite{feichtenhofer2020x3d}, to demonstrate that our method can further improve the efficiency of such state-of-the-art lightweight models. For TSM, we still use MobileNet-V2 and ResNet-50 as $f_{\textnormal{G}}$ and $f_{\textnormal{L}}$, but add TSM to them. Following the original design of TSM \cite{lin2019tsm}, a fully-connected layer is deployed as the classifier $f_{\textnormal{C}}$, and we average the frame-wise predictions as the outputs. For a fair comparison, we augment the vanilla TSM by introducing the same two backbone networks as ours (named as TSM+), where their output features are also concatenated to be fed into a linear classifier. In other words, TSM+ differentiates itself from our method only in that it feeds the uniformly-sampled whole video frames into ResNet-50, while we feed the dynamically-selected informative image patches from task-relative frames. For X3D, we simply replace MobileNet-V2/ResNet-50 by X3D-S/X3D-L on top of the settings of TSM. Here we directly compare the performance of our method with X3D, since both our two backbones come from the family of X3D networks. Following the experimental settings in the original papers of TSM \cite{lin2019tsm} and X3D \cite{feichtenhofer2020x3d}, the video recognition task on Sth-Sth V1\&V2, Jester and Kinetics-400 is considered here. We uniformly sample $T_0\!=\!24/36/48/96$ frames from each video, and set $T_{\textnormal{L}}\!=\!8/12/16/32$ correspondingly, with $T_{\textnormal{G}}\!=\!8/16$. See Appendix \ref{app:implementation_details} for more implementation details.

\textbf{Results on Sth-Sth V1\&V2 and Jester}
are reported in Table \ref{tab:sthsth}. Uni-AdaFocus enables TSM to concentrate the majority of computation on the most task-relevant video frames and image regions, while it allows allocating computation unevenly across ``easier'' and ``more difficult'' videos. As a consequence, the overall computational efficiency during inference is dramatically improved. For example, our method reduces GFLOPs by 3.99x (8.8 v.s. 35.1) on top of TSM+, but outperforms it by 2.9\% (62.5\% v.s. 59.6\%) on Sth-Sth V2 in terms of Top-1 accuracy. Compared with the recently proposed S2DNet framework, Uni-AdaFocus is able to achieve considerably better performance (\emph{e.g.}, enhancing the accuracy by 1.3-1.7\% using less inference cost) with 2.4x fewer training epochs. 
In Table \ref{tab:sthsth}, we also report the actual inference speed of our method on an NVIDIA 3090 GPU, which is benchmarked using a batch size of 128. 
At each mini-batch, when the inference procedure reaches each exit, the samples that meet the early-termination criterion will be output, with the remaining videos continuing to be processed.
It can be observed that our practical speedup is significant as well, with a slight drop compared with the theoretical results. We tentatively attribute this to the inadequate hardware-oriented optimization in our implementation.

% One can observe that by the computation our method is able to
% One can observe that by reducing the input size of the relatively expensive ResNet-50 network, our method enables TSM to  in the task-relevant region of each video using the same computation, leading to a significantly improved efficiency. For example, Uni-AdaFocus achieves the same performance as TSM+ with 1.5x less GFLOPs on Something-Something V1.

\textbf{Comparisons with the preliminary versions of AdaFocus on Sth-Sth}
are shown in Figure \ref{fig:sth_flops_acc}. When achieving the same accuracy, Uni-AdaFocus reduces the computational cost by 2.7-3.5x. Furthermore, when leveraging the same amount of computation at test time, Uni-AdaFocus improves the test accuracy by 2-3.5\% compared with AdaFocusV2.

\textbf{Results with X3D on Kinetics-400.}
In Table \ref{tab:x3d_k400}, we implement our method on top of the lightweight X3D \cite{feichtenhofer2020x3d} backbones, and report its performance on the large scale Kinetics-400 benchmark. For a comprehensive comparison, we also present the results of representative state-of-the-art efficient video understanding models on Kinetics-400. Note that we mainly compare Uni-AdaFocus-X3D with these baselines under similar validation accuracies or inference costs. One can clearly observe that our method contributes to a significantly improved accuracy-efficiency trade-off. For example, compared to X3D-L, Uni-AdaFocus-X3D reduces the computational cost by 3.76x (4.9 v.s. 18.4), while improving the accuracy by 0.5\% (76.2\% v.s. 75.7\%) in the meantime. Furthermore, compared to the best baseline, MoViNet-A2, Uni-AdaFocus-X3D has 2.10x less GLOPs/video (4.9 v.s. 10.3) when achieving the same accuracy of 75.0\%.

% AdaFocusV2 reduces the time consumption for training by $\sim\!2\times$, while dramatically improving the mAP using the same parch size (by 3-5\%). 

\vspace{-0.5ex}
\subsection{Real-world Application Scenarios}
\label{sec:real_world_app}

We also evaluate Uni-AdaFocus in three representative realistic application scenarios: 1) fine-grained diving action classification; 2) Alzheimer's and Parkinson's diseases diagnosis with brain magnetic resonance images (MRI); and 3) violence recognition for online videos. Due to spatial limitations, these results can be found in Appendix \ref{app:real_app_results}. In all scenarios, Uni-AdaFocus outperforms existing works by large margins, yielding a new state-of-the-art performance.

\begin{table}[t]
  \centering
  \begin{footnotesize}
  \caption{\label{tab:x3d_k400}\textbf{Performance of Uni-AdaFocus-X3D and representative efficient video understanding models on Kinetics-400.} MN2/R18/R34/R50/EN denotes MobileNet-V2/ResNet-18/34/50/EfficientNet. The \textcolor{blue}{blue} texts highlight the comparisons with the \underline{underlined} baselines.
  }
  \vskip -0.15in
  \setlength{\tabcolsep}{1mm}{
  \renewcommand\arraystretch{1.075}
  \resizebox{0.95\columnwidth}{!}{
  \begin{tabular}{c|cc|cc} 
  \toprule
  Method & Backbones  & \#Frames   &  \footnotesize{{\ Top-1 Acc.}} & \footnotesize{{GFLOPs}}\\  
  % \midrule
  \midrule
  ARTNet \cite{wang2018appearance} & {{R18}} & 16  & \  70.7\% & 23.5 $\times$ 250 \\
  TSN \cite{wang2016temporal} & {{InceptionV3}} & 25  & \  72.5\% & 3.2 $\times$ 250 \\
  S3D-G \cite{xie2018rethinking} & {{InceptionV1}} & 64  & \  74.7\% & 71.4 $\times$ 30 \\
  R(2+1)D \cite{tran2018closer} & {{R34}} & 32  & \  74.3\% & 152 $\times$ 10 \\
  I3D \cite{chen2021deep} & {{R50}} & 32  & \  \textbf{76.6\%} & 335 $\times$ 30 \\
  NL I3D \cite{wang2018non} & {{R50}} & 128  & \  76.5\% & 282 $\times$ 30 \\
  bLVNet-TAM \cite{fan2019more} & bL-R50  & 48  & \   73.5\% & 93.4 $\times$ 9 \\
  SlowFast \cite{feichtenhofer2019slowfast} & {{R50}} & 4/32  & \   75.6\% & 36 $\times$ 30 \\
  TSM \cite{lin2019tsm} & {{R50}} & 16  & \  74.7\% & 65 $\times$ 30 \\
  TEINet \cite{liu2020teinet} & {{R50}} & 16  & \   76.2\% & 66 $\times$ 30 \\
  TEA \cite{li2020tea} & {{R50}} & 16  & \   76.1\% & 70 $\times$ 30 \\
  SmallBig \cite{li2020smallbignet} & {{R50}} & 8  & \   76.3\% & 57 $\times$ 30 \\
  TANet \cite{liu2021tam} & {{R50}} & 8  & \   76.3\% & 43 $\times$ 30 \\
  TDN \cite{wang2021tdn}  & {{R50}} & 8  & \   \textbf{76.6\%} & 36 $\times$ 30 \\
  EfficientNet3D-B3 \cite{tan2019efficientnet}  & {{ENB3}} & 16  & \   72.4\% & 6.9 $\times$ 10 \\
  EfficientNet3D-B4 \cite{tan2019efficientnet}  & {{ENB4}} &  16 & \   74.5\% & 23.8 $\times$ 10 \\
  MoViNet-A2 \cite{kondratyuk2021movinets} & {{MoViNet-A2}} &  50 & \   75.0\% & 10.3 $\times$ 1 \  \\
  \midrule
  \multirow{1}{*}{X3D-S \cite{feichtenhofer2020x3d}}  & \multirow{1}{*}{{X3D-S}} & \multirow{1}{*}{13}  & \ 71.4\% & 2.0 $\times$ 3 \\
  \multirow{1}{*}{X3D-M \cite{feichtenhofer2020x3d}}  & \multirow{1}{*}{{X3D-M}} & \multirow{1}{*}{16} & \ 73.4\% & 4.7 $\times$ 3 \\ % XL: 77.3
  \multirow{1}{*}{X3D-L \cite{feichtenhofer2020x3d}}  & \multirow{1}{*}{{X3D-L}} & \multirow{1}{*}{16}  & \ \underline{75.7\%} & \underline{18.4 $\times$ 3} \\
  \multirow{2}{*}{\shortstack{Uni-AdaFocus-X3D (128$^2$)\\[-0.7ex]w/o sample-wise dynamic}}  & \multirow{2}{*}{\shortstack{X3D-S\\[-0.2ex]+X3D-L}} & 16+32 & \   75.3\% & 8.8 $\times$ 1 \\
  && 16+32 & \multirow{1}{*}{\ \textbf{76.6\%} \scriptsize{(\textcolor{blue}{$\uparrow$0.9\%})}} & \multirow{1}{*}{{8.8 \!\!$\times$\!\! 3} \scriptsize{(\textcolor{blue}{$\downarrow$2.09x})}} \\
  \midrule
  \multirow{2}{*}{\shortstack{Uni-AdaFocus-X3D (128$^2$)}}  & \multirow{2}{*}{\shortstack{X3D-S\\[-0.2ex]+X3D-L}} & 16+32 & \ 75.0\% & \textbf{4.9 $\times$ 1} \\
  && 16+32 & \ {76.2\%} \scriptsize{(\textcolor{blue}{$\uparrow$0.5\%})} & \multirow{1}{*}{\textbf{4.9 \!\!$\times$\!\! 3} \scriptsize{(\textcolor{blue}{$\downarrow$3.76x})}} \\
  \bottomrule
  \end{tabular}}}
  \end{footnotesize}
  \vspace{-4ex}
  \end{table}

\begin{figure*}[!b]
  \vspace{-4ex}
  \begin{center}
  \begin{minipage}{0.8525\columnwidth}
  \centering
  \begin{footnotesize}
  \makeatletter\def\@captype{table}\makeatother
  \caption{\textbf{Ablation studies of the training techniques proposed in Section \ref{sec:AdaFocusV2_techs}.} These techniques enable our method to be trained efficiently in an end-to-end fashion. The results of Uni-AdaFocus (w/o sample-wise dynamic) are reported. On Sth-Sth, we adopt \#Frames=8+12.}
  \label{tab:training_tech}
  \vskip -0.175in
  \setlength{\tabcolsep}{0.75mm}{
  \vspace{5pt}
  \renewcommand\arraystretch{1.075}
  \resizebox{\columnwidth}{!}{
  \begin{tabular}{ccc|cc|cc}
  \toprule
  Auxiliary & Diversity &  Stop-  & \multicolumn{2}{c|}{ActivityNet} &  \multicolumn{2}{c}{Sth-Sth V1} \\
  supervision & augmentation & gradient  & 128$^2$ & ${\Delta}$ &  128$^2$ & ${\Delta}$ \\
  % \midrule
  \midrule
  \textcolor{red}{\xmark} & \textcolor{red}{\xmark} & \textcolor{red}{\xmark} & \ 69.4\% & -- & 38.2\% & -- \\
  \textcolor{_green}{\cmark}& \textcolor{red}{\xmark} & \textcolor{red}{\xmark} & \ 72.5\% & +3.1\% & 42.9\% & +4.7\%  \\
  \textcolor{_green}{\cmark}&\textcolor{_green}{\cmark} & \textcolor{red}{\xmark} & \ 74.8\%& +2.3\% & 46.1\% & +3.2\%   \\
  \textcolor{_green}{\cmark}&\textcolor{_green}{\cmark} &\textcolor{_green}{\cmark}  & \ \textbf{78.7\%} & +3.9\% & \textbf{47.0\%} & +0.9\%  \\
  
  \bottomrule
  \end{tabular}}}
  \end{footnotesize}
      \end{minipage}    
  \hspace{0.005in}
  \begin{minipage}{1.15\columnwidth}
    \centering
    \begin{footnotesize}
    \makeatletter\def\@captype{table}\makeatother
    \caption{\textbf{Ablation studies of the techniques proposed in Sections \ref{sec:improved_spatial} and \ref{sec:frame_sample}.}  Representative conditions with different backbone networks, datasets and varying $P^2$ (input size for $f_{\textnormal{L}}$) are considered. The results of Uni-AdaFocus (w/o sample-wise dynamic) are reported. On Sth-Sth, we adopt \#Frames=8+12. The first line of the table corresponds to the last line of Table \ref{tab:training_tech}.}
    \label{tab:uni_techs}
    \vskip -0.175in
    \setlength{\tabcolsep}{0.75mm}{
    \vspace{5pt}
    \renewcommand\arraystretch{1.075}
    \resizebox{\columnwidth}{!}{
    \begin{tabular}{ccc|cc|cc|cc}
    \toprule
    Training $\pi$ with  & Deformable &  Dynamic  & \multicolumn{2}{c|}{ActivityNet} &  \multicolumn{4}{c}{Sth-Sth V1} \\
    deep features & patches & frame sampling  & 128$^2$ & ${\Delta}$ &  96$^2$ & ${\Delta}$ &  128$^2$ & ${\Delta}$ \\
    % \midrule
    \midrule
    \textcolor{red}{\xmark} & \textcolor{red}{\xmark} & \textcolor{red}{\xmark} & \ 78.7\% & -- & 45.1\% & --  & 47.0\% & -- \\
    \textcolor{_green}{\cmark}& \textcolor{red}{\xmark} & \textcolor{red}{\xmark} & \ 79.5\% & +0.8\% & 46.1\% & +1.0\%  & 48.1\% & +1.1\%  \\
    \textcolor{_green}{\cmark}&\textcolor{_green}{\cmark} & \textcolor{red}{\xmark} & \ 79.9\%& +0.4\% & 48.9\% & +2.8\%  & 49.4\% & +1.3\%   \\
    \textcolor{_green}{\cmark}&\textcolor{_green}{\cmark} &\textcolor{_green}{\cmark}  & \ \textbf{80.7\%} & +0.8\% & \textbf{49.5\%} & +0.6\%  &  \textbf{50.5\%} & +1.1\%  \\
    
    \bottomrule
    \end{tabular}}}
    \end{footnotesize}
        \end{minipage}  
  \end{center}
  % \vspace{-4ex}
\end{figure*}

\vspace{-0.5ex}
\subsection{Analytical Results}
\label{sec:abl}

\subsubsection{Ablation Studies}
\label{sec:abl_abl}

\textbf{Effectiveness of the techniques for end-to-end training} (introduced in Section \ref{sec:AdaFocusV2_techs})
is validated in Table \ref{tab:training_tech}. One can observe that all of the three techniques significantly improve the performance across different experimental settings. The combination of them allows our method to be trained effectively in an end-to-end fashion. Importantly, these techniques do not introduce additional tunable hyper-parameters. In all our experiments, we simply implement them as fixed components of the training algorithm.

\textbf{Effectiveness of the improved techniques for spatial-temporal dynamic computation} (introduced in Sections \ref{sec:improved_spatial} and \ref{sec:frame_sample}).
The ablation study results are provided in Table \ref{tab:uni_techs}. It is clear that guiding the training of $\pi$ with deep features consistently improves the performance by $\sim$1\% via obtaining a better patch selection policy. Built upon it, introducing deformable patches further yields significant gains by flexibly adapting to the task-relevant regions with diverse scales and shapes. Notably, this mechanism is particularly effective when the input size of the local encoder $f_{\textnormal{L}}$ is relatively small (\emph{e.g.}, 96$^2$). In other words, it is important to properly configure the actual contents of the inputs when the input size is small. Moreover, dynamic frame sampling is compatible with the spatial dynamic computation, and is able to further enhance the accuracy.

\begin{figure}[!t]
  % \vskip -0.015in
  % \vskip -0.175in
  \begin{center}
  \centerline{\includegraphics[width=0.71\columnwidth]{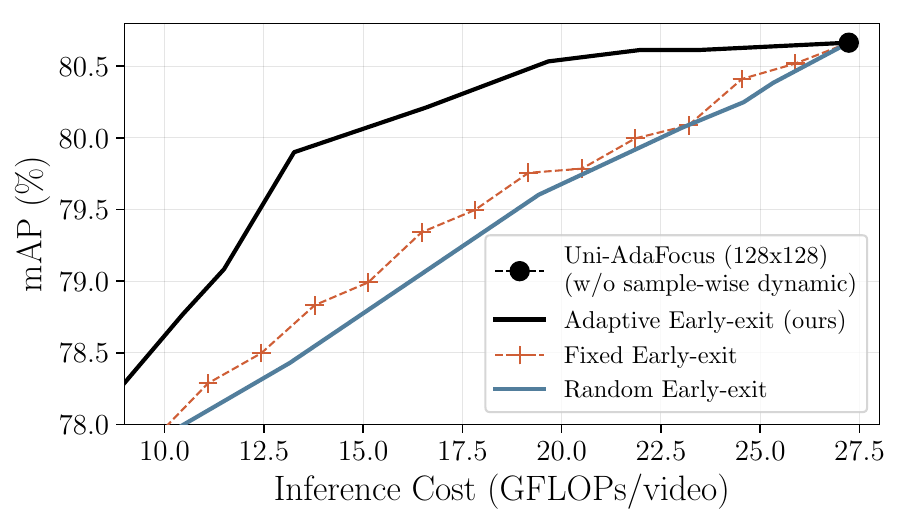}}
  \vskip -0.2in
  \caption{\textbf{Ablation studies of the conditional-exit algorithm.} The results on ActivityNet are provided. For a fair comparison, different variants are deployed on top of the same base model. Our method significantly reduces the computational cost when achieving the same mAP. \label{fig:abl_early_exit}
  }
  \end{center}
  \vspace{-3.5ex}
\end{figure}

\textbf{Design of the sample-wise dynamic computation mechanism} (introduced in Section \ref{sec:early_exit})
is studied in Figure \ref{fig:abl_early_exit}. We compare our design with two baselines, \emph{i.e.}, 1) early-exit with fixed frame length; and 2) random early-exit with the same exit proportion as Uni-AdaFocus. It can be clearly observed that our adaptive conditional-exit algorithm outperforms both of them. When achieving the same mAP, our method is able to effectively save the required amount of computation.

\textbf{Effectiveness of reusing $\bm{e}^{\textnormal{G}}_{t}$ for recognition.}
As aforementioned, our method leverages the coarse global feature $\bm{e}^{\textnormal{G}}_{t}$ for both activating the policy network $\pi$ and recognition, aiming to facilitate efficient feature reusing (see Section \ref{sec:arch} for details). The effectiveness of this design is studied in Table \ref{tab:abl_reuse_feat}. One can observe that this mechanism is able to improve the accuracy by 0.8-2.8\%. It is worth noting that this feature reusing introduces negligible computational cost, and hence the improvement of performance is almost a free lunch.

\begin{table}[t]
  \centering
  \begin{footnotesize}
  \makeatletter\def\@captype{table}\makeatother
  \caption{\textbf{Effects of Reusing $\bm{e}^{\textnormal{G}}_{t}$ for Recognition.}  Representative conditions with different backbone networks, datasets and varying $P^2$ (input size for $f_{\textnormal{L}}$) are considered. The results of Uni-AdaFocus (w/o sample-wise dynamic) are reported. On Sth-Sth, we adopt \#Frames=8+12. The last line of the table corresponds to the last line of Table \ref{tab:uni_techs}.}
  \label{tab:abl_reuse_feat}
  \vskip -0.2in
  \setlength{\tabcolsep}{0.5mm}{
  \vspace{5pt}
  \renewcommand\arraystretch{1.075}
  \resizebox{0.95\columnwidth}{!}{
  \begin{tabular}{c|cc|cc|cc|cc|cc}
  \toprule
  Reusing $\bm{e}^{\textnormal{G}}_{t}$ & \multicolumn{2}{c|}{ActivityNet} &  \multicolumn{4}{c|}{Sth-Sth V1} &  \multicolumn{4}{c}{Sth-Sth V2} \\
  for Recognition  & 128$^2$ & ${\Delta}$ &  96$^2$ & ${\Delta}$ &  128$^2$ & ${\Delta}$  &  96$^2$ & ${\Delta}$ &  128$^2$ & ${\Delta}$  \\
  % \midrule
  \midrule
  \textcolor{red}{\xmark} & \ 79.6\% & -- & 46.7\% & --  & 48.9\% & -- & 61.0\% & --  & 62.4\% & -- \\
  \textcolor{_green}{\cmark} & \ \textbf{80.7\%} & +1.1\% & \textbf{49.5\%} & +2.8\%  &  \textbf{50.5\%} & +1.6\%  & \textbf{62.6\%} & +1.6\%  &  \textbf{63.2\%} & +0.8\%  \\
  \bottomrule
  \end{tabular}}}
  \end{footnotesize}
  \vspace{-3ex}
\end{table}

\begin{table}[!t]
  \centering
  \begin{footnotesize}
  \caption{\textbf{Ablation studies of the learned policy for spatial-temporal dynamic computation.} All the three components of the policy are investigated, \emph{i.e.}, 1) how to localize informative patches, 2) how to determine the sizes/shapes of patches, and 3) how to sample task-relevant frames to process. We replace each of them with possible alternatives. \label{tab:abl_policy_net}}
  \vskip -0.2in
  \setlength{\tabcolsep}{0.6mm}{
  \vspace{5pt}
  \renewcommand\arraystretch{1.1}
  \resizebox{\columnwidth}{!}{
  \begin{tabular}{l|c|ccccc}
  \toprule
  \multicolumn{1}{c|}{\multirow{3}{*}{\shortstack{Experimental\\[-0.5ex]Configurations\\[-0.5ex]($P^2$=128$^2$)}}} & \multirow{3}{*}{\shortstack{Variants of Policy}} & \multicolumn{5}{c}{ActivityNet mAP after Processing} \\[-0.5ex]
  && \multicolumn{5}{c}{$t$ Frames (\emph{i.e.}, corresponding to $\bm{p}_t$)} \\
  && $t$=1 & $t$=2 & $t$=4 & $t$=8 & $t$=16\\
  % \midrule
  \midrule
  % \midrule
  \multirow{4}{*}{\!{\footnotesize{\shortstack[l]{\textbf{Patch localization: To study}\\[-0.5ex]\textcolor{gray}{Deformable-patch: None}\\[-0.5ex]\textcolor{gray}{Frame sampling: Uniform}}}}\!}  & Random Sampling & \ 26.1\% & 39.3\% & 53.2\% & 66.6\% & 74.3\% \\
  & Central Patch & \ 27.5\% & 39.2\% & 52.8\% & 65.8\% & 73.4\% \\
  & Gaussian Sampling & \ 23.8\% & 36.6\% & 52.2\% & 66.2\% & 74.5\% \\
  & Dynamic (ours)  & \ \textbf{30.4\%} & \textbf{43.4\%} & \textbf{57.5\%} & \textbf{70.1\%} & \textbf{77.2\%} \\
  \midrule
  \multirow{5}{*}{\!{\footnotesize{\shortstack[l]{\textit{Patch localization: Dynamic}\\[-0.5ex]\textbf{Deformable-patch: To study}\\[-0.5ex]\textcolor{gray}{Frame sampling: Uniform}}}}\!} 
  & \textcolor{gray}{Fixed Scale\&Shape}  & \ \textcolor{gray}{30.4\%} & \textcolor{gray}{43.4\%} & \textcolor{gray}{57.5\%} & \textcolor{gray}{70.1\%} & \textcolor{gray}{77.2\%} \\
  & Random Scale\&Shape & \ 17.8\% & 29.9\% & 44.6\% & 61.0\% & 70.3\% \\
  & Down-sampling & \ 32.7\% & 45.9\% & 58.3\% & 69.5\% & 75.9\% \\
  & \!(\ref{eq:learning_deformable_patches}) w/o regularization\! & \ 27.1\% & 39.5\% & 53.4\% & 66.8\% & 74.4\% \\
  & Dynamic (ours) & \ \textbf{33.5\%} & \textbf{46.7\%} & \textbf{59.6\%} & \textbf{71.7\%} & \textbf{78.3\%} \\
  \midrule
  \multirow{6}{*}{{\footnotesize{\shortstack[l]{\textit{Patch localization: Dynamic}\\[-0.5ex]\textit{Deformable-patch: Dynamic}\\[-0.5ex]\textbf{Frame sampling: To study}}}}} 
  & \textcolor{gray}{Uniform Sampling} & \ \textcolor{gray}{33.5\%} & \textcolor{gray}{46.7\%} & \textcolor{gray}{59.6\%} & \textcolor{gray}{71.7\%} & \textcolor{gray}{78.3\%} \\
  & Random Sampling & \ 36.8\% & 47.1\% & 59.6\% & 70.5\% & 77.2\% \\
  & Central Video Clip & \ \textbf{46.6\%} & 54.3\% & 61.9\% & 68.9\% & 75.0\% \\
  & Gaussian Sampling & \ 41.9\% & 52.8\% & 62.8\% & 71.1\% & 77.2\% \\
  & MG Sampler \cite{zhi2021mgsampler} & \ 34.1\% & 45.5\% & 60.1\% & 71.8\% & 78.7\% \\
  & Dynamic (ours) & \ 45.7\% & \textbf{58.3\%} & \textbf{67.4\%} & \textbf{74.8\%} & \textbf{79.6\%} \\
  \bottomrule
  \end{tabular}}}
  \end{footnotesize}
  \vspace{-3ex}
\end{table}

\textbf{Effectiveness of the learned spatial-temporal dynamic computation policy}
is investigated in Table \ref{tab:abl_policy_net}. The three major components of our policy are one by one replaced with several corresponding alternatives, aiming to validate whether our design is proper. In specific, our baselines include 1) patch localization policy: 1-a) randomly sampling patches, 1-b) cropping patches from the centres of the frames, 1-c) sampling patches from a gaussian distribution centred at the frame centre; 2) policy to determine the sizes/shapes of patches: 2-a) fixed patch size, 2-b) randomly determining patch sizes, 2-c) down-sampling the frames as patches, 2-d) removing the regularization term in Eq. (\ref{eq:learning_deformable_patches}) when training our policy; and 3) frame sampling policy: 3-a) uniformly sampling frames from each video, 3-b) randomly sampling frames, 3-c) directly taking the central clips of videos, 3-d) sampling frames from a gaussian distribution centred at the video centre, 3-e) adopting MG Sampler \cite{zhi2021mgsampler}.
For an isolated and focused reflection of the effects of our policy with clear comparisons, here we do not reuse the global feature $\bm{e}^{\textnormal{G}}_{t}$ for recognition (its effects have been studied in Table \ref{tab:abl_reuse_feat}). We assume that Uni-AdaFocus processes a fixed number of frames for all videos, and report the corresponding mAP on ActivityNet.
One can observe that our learned policies have considerably better performance compared with all the baselines. For 1), an interesting phenomenon is that random policy appears strong and outperforms the central policy, which may be attributed to the spatial similarity between frames. That is to say, adjacent central patches might have repetitive contents, while randomly sampling is likely to collect more comprehensive information. For 2), we observe that introducing the regularization term in Eq. (\ref{eq:learning_deformable_patches}) is an important technique to learn effective deformable-patch policies.
For 3), although the recently proposed MG Sampler \cite{zhi2021mgsampler} is also able to improve the accuracy, our dynamic frame sampling algorithm outperforms it by large margins.

% although,  we show that our dynamic frame sampling algorithm outperforms the recently proposed MG Sampler \cite{zhi2021mgsampler} by large margins. 

% find the recently proposed MG Sampler \cite{zhi2021mgsampler} is also able to improve the performance of our network, but our dynamic frame sampling algorithm outperforms it significantly. 

\begin{figure}[!t]
  % \vskip -0.015in
  % \hskip -0.1in
  \begin{center}
  \centerline{\includegraphics[width=0.95\columnwidth]{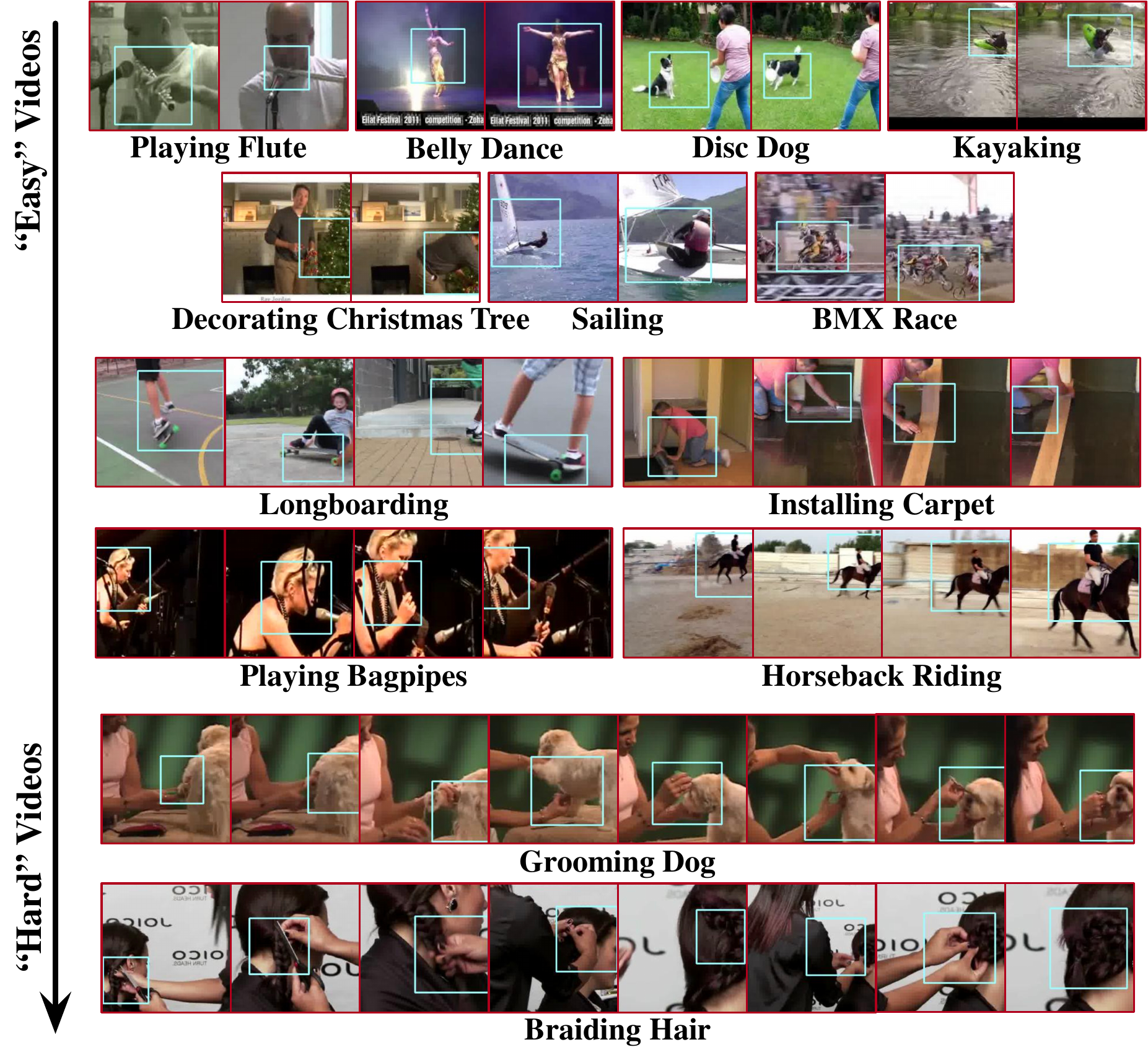}}
  \vskip -0.175in
  \caption{\textbf{Examples of the task-relevant patches and informative videos frames selected by Uni-AdaFocus (zoom in for details).} We present a variety of representative input videos, where ``easy/hard'' videos refer to the samples to which Uni-AdaFocus allocates a relatively smaller or larger number of computation resources. \label{fig:visualize}
  }
  \end{center}
  \vspace{-2ex}
\end{figure}

\begin{table*}[!t]
  \centering
  \begin{footnotesize}
  \caption{\textbf{Effects of spatial, temporal, and sample-wise dynamic computation.} All results are based on identical experimental setups except for eliminating several specified types of dynamic computation strategies. We compare the performance of different variants under the same inference cost.}
  \label{tab:_discuss_effects_three_dynamic}
  \vskip -0.15in
  \setlength{\tabcolsep}{2.3mm}{
  \vspace{5pt}
  \renewcommand\arraystretch{1.15}
  \resizebox{1.52\columnwidth}{!}{
  \begin{tabular}{ccc|llllll}
  \toprule
  \multicolumn{3}{c|}{Strategies of Dynamic Computation} & 
  \multicolumn{6}{c}{
    ActivityNet mAP with Different Average Computational Costs (in GFLOPs/video)
    } \\[-0.1ex]
  \!\!\!Spatial {\tiny ($P^2$=128$^2$)}\!\!\! & \!\!\!Temporal\!\!\! & \!\!\!Sample-wise\!\!\! &
  \multicolumn{1}{c}{11.0} & 
  \multicolumn{1}{c}{13.7} & 
  \multicolumn{1}{c}{16.4} & 
  \multicolumn{1}{c}{19.1} & 
  \multicolumn{1}{c}{21.8} & 
  \multicolumn{1}{c}{24.4} \\
  \midrule
  \textcolor{red}{\xmark} &  \textcolor{red}{\xmark} & \textcolor{red}{\xmark}
  & 41.8\%  \tiny{\textcolor{gray}{($\downarrow$27.8\%)}} 
  & 47.8\%  \tiny{\textcolor{gray}{($\downarrow$26.5\%)}}
  & 53.6\%  \tiny{\textcolor{gray}{($\downarrow$23.4\%)}}
  & 58.0\%  \tiny{\textcolor{gray}{($\downarrow$20.5\%)}}
  & 62.4\%  \tiny{\textcolor{gray}{($\downarrow$16.6\%)}} 
  & 65.8\%  \tiny{\textcolor{gray}{($\downarrow$13.7\%)}} 
  \\
  \textcolor{red}{\xmark} &  \textcolor{red}{\xmark} & \textcolor{_green}{\cmark}
  & 46.4\%  \tiny{\textcolor{gray}{($\downarrow$23.2\%)}} 
  & 54.1\%  \tiny{\textcolor{gray}{($\downarrow$20.2\%)}}
  & 59.7\%  \tiny{\textcolor{gray}{($\downarrow$17.3\%)}}
  & 64.2\%  \tiny{\textcolor{gray}{($\downarrow$14.3\%)}}
  & 68.0\%  \tiny{\textcolor{gray}{($\downarrow$11.0\%)}} 
  & 70.7\%  \tiny{\textcolor{gray}{($\downarrow$8.8\%)}} 
  \\
  \textcolor{red}{\xmark} &  \textcolor{_green}{\cmark} & \textcolor{red}{\xmark}
  & 55.6\%  \tiny{\textcolor{gray}{($\downarrow$14.0\%)}} 
  & 60.4\%  \tiny{\textcolor{gray}{($\downarrow$13.9\%)}}
  & 64.3\%  \tiny{\textcolor{gray}{($\downarrow$12.7\%)}}
  & 67.6\%  \tiny{\textcolor{gray}{($\downarrow$10.9\%)}}
  & 70.4\%  \tiny{\textcolor{gray}{($\downarrow$8.6\%)}} 
  & 72.3\%  \tiny{\textcolor{gray}{($\downarrow$7.2\%)}} 
  \\
  \textcolor{_green}{\cmark} &  \textcolor{red}{\xmark} & \textcolor{red}{\xmark}
  & 54.2\%  \tiny{\textcolor{gray}{($\downarrow$15.4\%)}} 
  & 61.2\%  \tiny{\textcolor{gray}{($\downarrow$13.1\%)}}
  & 66.6\%  \tiny{\textcolor{gray}{($\downarrow$10.4\%)}}
  & 70.7\%  \tiny{\textcolor{gray}{($\downarrow$7.8\%)}}
  & 74.4\%  \tiny{\textcolor{gray}{($\downarrow$4.6\%)}} 
  & 76.9\%  \tiny{\textcolor{gray}{($\downarrow$2.6\%)}} 
  \\
  \textcolor{red}{\xmark} &  \textcolor{_green}{\cmark} & \textcolor{_green}{\cmark} 
  & 59.2\%  \tiny{\textcolor{gray}{($\downarrow$10.4\%)}} 
  & 65.1\%  \tiny{\textcolor{gray}{($\downarrow$9.2\%)}}
  & 69.0\%  \tiny{\textcolor{gray}{($\downarrow$8.0\%)}}
  & 71.5\%  \tiny{\textcolor{gray}{($\downarrow$7.0\%)}}
  & 73.6\%  \tiny{\textcolor{gray}{($\downarrow$5.4\%)}} 
  & 75.1\%  \tiny{\textcolor{gray}{($\downarrow$4.4\%)}} 
  \\
  \textcolor{_green}{\cmark} & \textcolor{red}{\xmark} & \textcolor{_green}{\cmark} 
  & 63.0\%  \tiny{\textcolor{gray}{($\downarrow$6.6\%)}} 
  & 70.7\%  \tiny{\textcolor{gray}{($\downarrow$3.6\%)}}
  & 74.9\%  \tiny{\textcolor{gray}{($\downarrow$2.1\%)}}
  & 77.0\%  \tiny{\textcolor{gray}{($\downarrow$1.5\%)}}
  & 77.8\%  \tiny{\textcolor{gray}{($\downarrow$1.2\%)}} 
  & 78.1\%  \tiny{\textcolor{gray}{($\downarrow$1.4\%)}} 
  \\
  \textcolor{_green}{\cmark} &  \textcolor{_green}{\cmark} & \textcolor{red}{\xmark}
  & 62.6\%  \tiny{\textcolor{gray}{($\downarrow$7.0\%)}} 
  & 67.5\%  \tiny{\textcolor{gray}{($\downarrow$6.8\%)}}
  & 71.2\%  \tiny{\textcolor{gray}{($\downarrow$5.8\%)}}
  & 74.4\%  \tiny{\textcolor{gray}{($\downarrow$4.1\%)}}
  & 76.8\%  \tiny{\textcolor{gray}{($\downarrow$2.2\%)}} 
  & 78.5\%  \tiny{\textcolor{gray}{($\downarrow$1.0\%)}} 
  \\
  \textcolor{_green}{\cmark}  & \textcolor{_green}{\cmark}  & \textcolor{_green}{\cmark}  
  & \textbf{69.6\%} 
  & \textbf{74.3\%}  
  & \textbf{77.0\%}  
  & \textbf{78.5\%} 
  & \textbf{79.0\%} 
  & \textbf{79.5\%} \\
  \bottomrule
  \end{tabular}}}
  \end{footnotesize}
  % \vskip -0.09in
  \vskip -0.1in
  % \vskip -0.1in
\end{table*}

\begin{table*}[!t]
  \centering
  \begin{footnotesize}
  \caption{
    \textbf{Analysis of the design of our spatial dynamic computation mechanism.} 
    We investigate eliminating its three major components (two of which are introduced by the deformable patch mechanism proposed in Section \ref{sec:improved_spatial_sub2}), or further incorporating more flexible dynamic transformations.
    }
  \label{tab:_discuss_patch_selection}
  \vskip -0.15in
  \setlength{\tabcolsep}{2mm}{
  \vspace{5pt}
  \renewcommand\arraystretch{1.15}
  \resizebox{2.0\columnwidth}{!}{
  \begin{tabular}{ccc|cccc|lllll}
  \toprule
  \multicolumn{7}{c|}{Characteristics of the Patch Selection Mechanism within Uni-AdaFocus ($P^2$=128$^2$)} & 
  \multicolumn{5}{c}{
    \multirow{3}{*}{
    \shortstack{
      ActivityNet mAP after Processing $t$ Frames\\[-0.25ex]
      (\emph{i.e.}, corresponding to $\bm{p}_t$)
    }
    }} \\[-0.1ex]
  {\emph{(basic spatial dynamic)}} &  \multicolumn{2}{c|}{{\emph{(deformable-patch)}}}  & 
  \multicolumn{4}{c|}{
    {
    \emph{(introducing \textbf{more dynamic transformations})}}
    } &&& \\[-0.1ex]
  \multirow{2}{*}{\ {\shortstack{Dynamic Patch\\[-0.5ex]Localization}}\ } & 
  \multirow{2}{*}{\ {\shortstack{Dynamic\\[-0.5ex]Patch Scale}}\ } & 
  \multirow{2}{*}{\ {\shortstack{Dynamic\\[-0.5ex]Patch Shape}}\ } &    
  \multirow{2}{*}{{\shortstack{Rotation}}} & 
  \multirow{2}{*}{{\shortstack{Affine}}} &  
  \multirow{2}{*}{{\shortstack{Homo-\\[-0.5ex]graphy}}} & 
  \multirow{2}{*}{{\shortstack{Thin-plate\\[-0.5ex]Spline}}} 
  &&& \\[-0.1ex]
  &&&&&&& 
  \multicolumn{1}{c}{$t$=1} & \multicolumn{1}{c}{$t$=2} & \multicolumn{1}{c}{$t$=4} & \multicolumn{1}{c}{$t$=8} & \multicolumn{1}{c}{$t$=16} \\
  \midrule
  \textcolor{red}{\xmark} \emph{(random sampling)} & \textcolor{red}{\xmark} \emph{(fixed)} & \textcolor{red}{\xmark} \emph{(fixed)} & \textcolor{red}{\xmark} & \textcolor{red}{\xmark} & \textcolor{red}{\xmark} &  \textcolor{red}{\xmark}
  & 36.7\% \tiny{\textcolor{gray}{($\downarrow$9.0\%)}}
  & 51.1\% \tiny{\textcolor{gray}{($\downarrow$7.2\%)}}
  & 62.4\% \tiny{\textcolor{gray}{($\downarrow$5.0\%)}}
  & 71.0\% \tiny{\textcolor{gray}{($\downarrow$3.8\%)}}
  & 76.5\% \tiny{\textcolor{gray}{($\downarrow$3.1\%)}} 
  \\
  \textcolor{red}{\xmark} \emph{(central patch)} & \textcolor{red}{\xmark} \emph{(fixed)} & \textcolor{red}{\xmark} \emph{(fixed)} & \textcolor{red}{\xmark} & \textcolor{red}{\xmark} & \textcolor{red}{\xmark} &  \textcolor{red}{\xmark}
  & 38.1\% \tiny{\textcolor{gray}{($\downarrow$7.6\%)}}
  & 50.1\% \tiny{\textcolor{gray}{($\downarrow$8.2\%)}}
  & 60.3\% \tiny{\textcolor{gray}{($\downarrow$7.1\%)}}
  & 68.9\% \tiny{\textcolor{gray}{($\downarrow$5.9\%)}}
  & 75.1\% \tiny{\textcolor{gray}{($\downarrow$4.5\%)}} 
  \\
  \textcolor{_green}{\cmark} & \textcolor{red}{\xmark} \emph{(fixed)} & \textcolor{red}{\xmark} \emph{(fixed)} & \textcolor{red}{\xmark} & \textcolor{red}{\xmark} & \textcolor{red}{\xmark} &  \textcolor{red}{\xmark}
  & 42.5\% \tiny{\textcolor{gray}{($\downarrow$3.2\%)}} 
  & 55.7\% \tiny{\textcolor{gray}{($\downarrow$2.6\%)}} 
  & 65.1\% \tiny{\textcolor{gray}{($\downarrow$2.3\%)}} 
  & 73.0\% \tiny{\textcolor{gray}{($\downarrow$1.8\%)}} 
  & 78.5\% \tiny{\textcolor{gray}{($\downarrow$1.1\%)}} 
  \\
  \textcolor{_green}{\cmark} &  \textcolor{_green}{\cmark} & \textcolor{red}{\xmark} \emph{(fixed)} & \textcolor{red}{\xmark} & \textcolor{red}{\xmark} & \textcolor{red}{\xmark} &  \textcolor{red}{\xmark}
  & 45.3\%  \tiny{\textcolor{gray}{($\downarrow$0.4\%)}} 
  & 57.9\%  \tiny{\textcolor{gray}{($\downarrow$0.4\%)}}
  & 66.6\%  \tiny{\textcolor{gray}{($\downarrow$0.8\%)}}
  & 73.8\%  \tiny{\textcolor{gray}{($\downarrow$1.0\%)}}
  & 79.1\%  \tiny{\textcolor{gray}{($\downarrow$0.5\%)}} 
  \\
  \textcolor{_green}{\cmark}  & \textcolor{_green}{\cmark}  & \textcolor{_green}{\cmark}  & \textcolor{red}{\xmark} & \textcolor{red}{\xmark} & \textcolor{red}{\xmark} &  \textcolor{red}{\xmark}
  & \textbf{45.7\%} 
  & \textbf{58.3\%}  
  & \textbf{67.4\%}  
  & \textbf{74.8\%} 
  & \textbf{79.6\%} \\
  \midrule
  \textcolor{_green}{\cmark}  & \textcolor{_green}{\cmark}  & \textcolor{_green}{\cmark}  &  \textcolor{_green}{\cmark} & \textcolor{red}{\xmark} & \textcolor{red}{\xmark} & \textcolor{red}{\xmark} 
  & 45.6\%  \tiny{\textcolor{gray}{($\downarrow$0.1\%)}}
  & 57.8\%  \tiny{\textcolor{gray}{($\downarrow$0.5\%)}}
  & 67.3\%  \tiny{\textcolor{gray}{($\downarrow$0.1\%)}}
  & 74.2\%  \tiny{\textcolor{gray}{($\downarrow$0.6\%)}}
  & 79.3\%  \tiny{\textcolor{gray}{($\downarrow$0.3\%)}}
  \\
  \textcolor{_green}{\cmark}  & \textcolor{_green}{\cmark}  & \textcolor{_green}{\cmark}  & \textcolor{red}{\xmark} &  \textcolor{_green}{\cmark} & \textcolor{red}{\xmark} &  \textcolor{red}{\xmark}
  & 44.9\%  \tiny{\textcolor{gray}{($\downarrow$0.8\%)}}
  & 57.5\%  \tiny{\textcolor{gray}{($\downarrow$0.8\%)}}
  & 66.3\%  \tiny{\textcolor{gray}{($\downarrow$1.1\%)}}
  & 74.0\%  \tiny{\textcolor{gray}{($\downarrow$0.8\%)}}
  & 78.7\%  \tiny{\textcolor{gray}{($\downarrow$0.9\%)}}
  \\
  \textcolor{_green}{\cmark}  & \textcolor{_green}{\cmark}  & \textcolor{_green}{\cmark}  & \textcolor{red}{\xmark} & \textcolor{red}{\xmark} &  \textcolor{_green}{\cmark} &  \textcolor{red}{\xmark}
  & 43.0\%  \tiny{\textcolor{gray}{($\downarrow$2.7\%)}}
  & 56.2\%  \tiny{\textcolor{gray}{($\downarrow$2.1\%)}}
  & 65.4\%  \tiny{\textcolor{gray}{($\downarrow$2.0\%)}}
  & 72.6\%  \tiny{\textcolor{gray}{($\downarrow$2.2\%)}}
  & 78.6\%  \tiny{\textcolor{gray}{($\downarrow$1.0\%)}}
  \\
  \textcolor{_green}{\cmark}  & \textcolor{_green}{\cmark}  & \textcolor{_green}{\cmark}  & \textcolor{red}{\xmark} & \textcolor{red}{\xmark} & \textcolor{red}{\xmark} &   \textcolor{_green}{\cmark}
  & 41.1\%  \tiny{\textcolor{gray}{($\downarrow$4.6\%)}}
  & 53.4\%  \tiny{\textcolor{gray}{($\downarrow$4.9\%)}}
  & 62.8\%  \tiny{\textcolor{gray}{($\downarrow$4.6\%)}}
  & 70.8\%  \tiny{\textcolor{gray}{($\downarrow$4.0\%)}}
  & 76.6\%  \tiny{\textcolor{gray}{($\downarrow$3.0\%)}}
  \\
  \bottomrule
  \end{tabular}}}
  \end{footnotesize}
  % \vskip -0.09in
  \vskip -0.1in
  % \vskip -0.1in
\end{table*}

\subsubsection{Visualization Results}

% \textbf{Visualization results}
Representative visualization results are shown in Figure \ref{fig:visualize} (blue boxes indicate the patches selected by Uni-AdaFocus). Our method can adaptively attend to the informative regions of some task-relevant video frames, such as the dancer, the dog, the skateboard, and the actions of human hands. Due to spatial limitations, more visualizations (including an analysis of failure cases) can be found in Appendix \ref{app:additional_visualization}.

% \clearpage

\subsubsection{Discussions}

Section \ref{sec:abl_abl} has comprehensively investigated whether each component of Uni-AdaFocus works as we expect. On top of those studies, here we further design a series of more in-depth analyses, aiming to highlight the important findings of this paper, and clarify our novel contributions over existing works. The performance of Uni-AdaFocus in Tables \ref{tab:_discuss_effects_three_dynamic}-\ref{tab:_discuss_vs_oc} is built upon the experimental setups of the last line of Table \ref{tab:abl_policy_net}.

% Similar to Table \ref{tab:abl_policy_net}, for an isolated and focused reflection of the effects of our dynamic computation strategies with clear comparisons, we do not reuse the global feature $\bm{e}^{\textnormal{G}}_{t}$ for recognition (its effects have been studied in Table \ref{tab:abl_reuse_feat}).

% "In Table \ref{tab:_discuss_effects_three_dynamic}, we examine the impact of removing specific facets of spatial, temporal, and sample-wise dynamic computation, presenting the performance outcomes of the ablated variants. Concretely, upon eliminating each respective component, our model undergoes processing using full video frames, uniformly sampled frames, or randomly executed early-exit mechanisms. This ensures equitable distribution of computational resources across distinct spatial regions, temporal instances, or diverse samples. Notably, the integration of any single dynamic computation type yields a marked enhancement in the computational efficiency of Uni-AdaFocus, demonstrating their compatibility and combinability. Particularly noteworthy is the significant performance boost observed with the inclusion of spatial adaptiveness. Furthermore, the combination of 'spatial+sample-wise' exhibits a more pronounced improvement in 'spatial' over 'spatial+temporal', albeit both combinations yield inferior mean Average Precision (mAP) scores compared to 'spatial+temporal+sample-wise'."

\textbf{Importance of different dynamic computation strategies.}
Table \ref{tab:_discuss_effects_three_dynamic} examines the impact of eliminating certain aspects of spatial/temporal/sample-wise dynamic computation. Specifically, when removing each respective component, our model processes full video frames, uniformly samples frames, or randomly performs early-exit, yielding an equivalent computation allocation across different spatial regions, temporal locations, or diverse samples. It can be observed that modeling any one within the three types of dynamic computation significantly improves the computational efficiency of Uni-AdaFocus, while they are compatible with each other in an arbitrary form of combination. In general, incorporating spatial adaptiveness results in the most significant gains. 

% Moreover, ``spatial+sample-wise'' seems to further enhance ``spatial'' more effectively than ``spatial+temporal'', yet both of them attain notably inferior mAP compared to ``spatial+temporal+sample-wise''.

% the most significant gains usually come from considering spatial dynamic computation. Moreover, sample-wise adaptiveness

% \textbf{Development of the Spatial Dynamic Computation Mechanism.}
% The fundamental attributes of visual content, including diverse variations in locations, scales, and shapes, necessitate an adaptive approach to region selection. As delineated in Table \ref{tab:discuss_patch_selection}, explicit consideration of these variations significantly enhances the performance of Uni-AdaFocus. Intriguingly, the introduction of more flexible dynamic transformations, such as rotation, affine, homography, and thin-plate spline, yields a marginal decrease in accuracy. This phenomenon may stem from the limited general applicability of these image warping and rectification techniques to our specific task, potentially distorting inputs to the local encoder $f{\textnormal{L}}$ from their natural distribution and augmenting the complexity of training the policy network $\pi$.

\textbf{Design of the spatial dynamic computation mechanism.}
Diverse variations in locations, scales, and shapes are three fundamental characteristics intrinsic to visual elements. As shown in Table \ref{tab:_discuss_patch_selection}, the performance of Uni-AdaFocus can be markedly improved by explicitly considering adapting to every type of these variations when selecting the task-relevant regions of different video frames. Intriguingly, introducing more flexible dynamic transformations (\emph{e.g.}, affine, homography, and thin-plate spline) slightly degrades the accuracy. This phenomenon may be caused by the lack of sufficiently general demands for utilizing these image warping/rectification techniques to process the task-relevant regions in our problem. On the contrary, they may distort the inputs of the local encoder $f_{\textnormal{L}}$ from the natural distribution, and increase the complexity of training the policy network $\pi$.

\section{Conclusion}
% \vspace{-0.75ex}
\label{sec:conclusion}

This paper presented Uni-AdaFocus, an approach that enables deep networks to allocate computation dynamically across three dimensions: spatial, temporal, and different samples. The major motivation behind it is to concentrate the computational resources on the most task-relevant image regions, informative video frames, and relatively difficult test data, and thus attain a superior overall accuracy with a minimal total computational cost. We delved deep into the model design, training, and efficient implementation of Uni-AdaFocus. With our proposed techniques, Uni-AdaFocus can be easily deployed on top of popular lightweight backbones (\emph{e.g.}, TSM and X3D) and considerably improves their inference efficiency through adaptive computation. Moreover, it can be trained efficiently in an end-to-end fashion. Empirical results based on seven large-scale benchmark datasets and three real-world application scenarios demonstrate that Uni-AdaFocus achieves state-of-the-art performance in terms of generalization performance, theoretical computational efficiency, and practical inference speed.

In the future, it would be interesting to further explore how our proposed dynamic computation techniques can be employed to improve the inference efficiency of multi-modal large language models. For example, one may consider adaptively identifying task-relevant video frames and image regions conditioned on the input text or audio prompts. Given the expensive cost of training large models, it would also be important to investigate how pre-trained multi-modal models can be fine-tuned to acquire such capabilities of dynamic computation.

% effectively enhance the overall computational efficiency. 
% We delved deep into the network design, training algorithm, and efficient implementation of Uni-AdaFocus. With our proposed techniques, Uni-AdaFocus can be efficiently trained in an end-to-end fashion, while it can be deployed on top of existing lightweight backbones and significantly improves their 
% In this paper, we enabled the end-to-end training of adaptive focus video recognition networks (AdaFocus). We first proposed a differentiable interpolation-based operation for selecting patches, allowing the gradient back-propagation throughout the whole model. Then we present three tailored training techniques to address the optimization issues introduced by end-to-end training. Experimental results on six benchmarks demonstrated that our AdaFocusV2 network is considerably more efficient to train than the original AdaFocus model, while achieving state-of-the-art performance. 

\ifCLASSOPTIONcompsoc
  \section*{Acknowledgments}
\else
  \section*{Acknowledgment}
\fi

% This work is supported in part by National Key R\&D Program of China (2021ZD0140407), the National Natural Science Foundation of China under Grant 62022048, Tsinghua University-China Mobile Communications Group Co.,Ltd. Joint Institute and Guoqiang Institute of Tsinghua University.

% This work is supported in part by the National Key R\&D Program of China under Grant 2021ZD0140407, the National Natural Science Foundation of China under Grants 42327901 and 62321005, and Tsinghua University-China Mobile Communications Group Co., Ltd. Joint Institute.
% This work is supported in part by the National Key R\&D Program of China under Grant 2021ZD0140407, the National Natural Science Foundation of China under Grants 42327901 and 62321005, the BNRist project (No. BNR2024TD03003), and Tsinghua University-China Mobile Communications Group Co., Ltd. Joint Institute.

This work was supported in part by the National Natural Science Foundation of China under Grant 42327901 and Grant 62321005, in part by the BNRist project under Grant BNR2024TD03003, and in part by Tsinghua University-China Mobile Communications Group Company, Ltd. Joint Institute.

% and Guoqiang Institute of Tsinghua University.

% This work was supported in part by the National Key R&D Program of China under Grant 2021ZD0140407, and in part by the National Natural Science Foundation of China under Grants 42327901.
\ifCLASSOPTIONcaptionsoff
  \newpage
\fi



%\bibliographystyle{IEEEtran}

% H.~Kopka and P.~W. Daly, \emph{A Guide to {\LaTeX}}, 3rd~ed.\hskip 1em plus

% \end{thebibliography}

\bibliographystyle{IEEEtran}
\bibliography{IEEEabrv,IEEEtran}

\clearpage

% \newpage
% \onecolumn

\clearpage
\appendices

% \section*{Appendix}

% \onecolumn

% This paper extends conference papers that introduced the basic AdaFocus framework \cite{Wang_2021_ICCV} and preliminarily discussed its end-to-end training \cite{wang2021adafocus}. We have improved these earlier works substantially in several important aspects. Moreover, our deep-feature-based approach for training the patch selection strategy (Section \ref{sec:improved_spatial_sub1}) is conceptually relevant to, and improved upon \cite{wang2022adafocusv3} (see Table \ref{tab:_discuss_vs_adaV3} for a detailed comparison). See Appendix \ref{app:summary_of_change} for a summary of changes.

\section{Summary of Changes}
\label{app:summary_of_change}

This paper extends two previous conference papers that introduced the basic AdaFocus network \cite{Wang_2021_ICCV} and preliminarily discussed its end-to-end training \cite{wang2021adafocus}. 
Moreover, our deep-feature-based approach for training the patch selection policy (Section \ref{sec:improved_spatial_sub1}) is conceptually relevant to, and improved upon \cite{wang2022adafocusv3} (see Table \ref{tab:_discuss_vs_adaV3} for a detailed comparison). We have improved these earlier works substantially in several important aspects, as summarized in the following.
\vspace{-0.4ex}
\begin{itemize}
    \item The spatial dynamic computation algorithm has been improved. First, we introduce a deep-feature-based approach for training the patch selection policy effectively, which incurs negligible additional cost while considerably improving the final accuracy (Section \ref{sec:improved_spatial_sub1}). Second, we develop a deformable patch mechanism, enabling AdaFocus to flexibly capture the task-relevant spatial regions within each video frame with diverse shapes, sizes, and scales (Section \ref{sec:improved_spatial_sub2}).
    \item We propose a dynamic frame sampling algorithm tailored for AdaFocus (Section \ref{sec:frame_sample}), and extend our method by simultaneously modeling the spatial, temporal, and sample-wise dynamic computation. The resulting Uni-AdaFocus framework (Figure \ref{fig:overview_uniada}) achieves a remarkable performance across various settings in terms of computational efficiency.
    \item The experimental results on six benchmark datasets (\emph{i.e.}, ActivityNet, FCVID, Mini-Kinetics, Something-Something V1\&V2, and Jester) have been updated. We report new state-of-the-art results with Uni-AdaFocus (Tables \ref{tab:actnet_main_table} and \ref{tab:sthsth}, Figure \ref{fig:actnet_vs_sota}), and comprehensively compare Uni-AdaFocus with the works in \cite{Wang_2021_ICCV, wang2021adafocus, wang2022adafocusv3} (Figures \ref{fig:actnet_vs_v1_eval} and \ref{fig:sth_flops_acc}, Table \ref{tab:actnet_vs_v1_train}). Moreover, we conduct more experiments to evaluate our method on top of the efficient X3D \cite{feichtenhofer2020x3d} backbone networks on the large-scale Kinetics-400 benchmark (Table \ref{tab:x3d_k400}).
    \item We further apply our approach to three real-world application scenarios (\emph{i.e.}, fine-grained diving action classification, Alzheimer's and Parkinson's diseases diagnosis using brain magnetic resonance images (MRI), and violence recognition for online videos), and report encouraging results (Section \ref{sec:real_world_app}).
    \item Additional analytical results are provided, including the comprehensive ablation studies of the new components (Tables \ref{tab:uni_techs}, \ref{tab:abl_reuse_feat}, and \ref{tab:abl_policy_net}, Figure \ref{fig:abl_early_exit}), in-depth discussions of our important findings (Tables \ref{tab:_discuss_effects_three_dynamic} and \ref{tab:_discuss_patch_selection}), thorough comparisons with existing works (Tables \ref{tab:_discuss_vs_adaV3} and \ref{tab:_discuss_vs_oc}), and new visualization results (Figures \ref{fig:visualize}, \ref{fig:fail_case}, and \ref{fig:multi_person}).
\end{itemize}

\section{Three Stage Training of AdaFocusV1}
\label{app:three_stage}

Since the formulation of AdaFocusV1 includes both continuous (\emph{i.e.}, video recognition) and discrete (\emph{i.e.}, patch localization) optimization, the standard end-to-end training paradigm cannot be directly applied. Therefore, we introduce a three-stage training algorithm to solve continuous and discrete optimization problems alternatively.

\textbf{Stage I: Warm-up.}
We first initialize $f_{\textnormal{G}}$, $f_{\textnormal{L}}$ and $f_{\textnormal{C}}$, but leave the policy network $\pi$ out at this stage. Then we randomly sample the image patches $\tilde{\bm{v}}_t$ to minimize the cross-entropy loss $L_{\textnormal{CE}}(\cdot)$ over the training set $\mathcal{D}_{\textnormal{train}}$:
\begin{equation}
    \label{eq:stage_1}
    \begin{split}
        \mathop{\textnormal{minimize}}_{f_{\textnormal{G}}, f_{\textnormal{L}}, f_{\textnormal{C}}}\ \ \  \mathbb{E}&_{\{\bm{v}_1, \bm{v}_2, \ldots\} \in \mathcal{D}_{\textnormal{train}}}
    \left[
        \frac{1}{T}\sum\nolimits_{t=1}^{T} L_{\textnormal{CE}}(\bm{p}_t, y)
    \right], \\ &\tilde{\bm{v}}_t\sim\textnormal{RandomCrop}(\bm{v}_t).
    % \frac{1}{\mathcal{D}_{\textnormal{train}}}\sum_{\{\bm{v}_1, \bm{v}_2, \ldots\} \in \mathcal{D}_{\textnormal{train}}}
    \end{split}
\end{equation}
Here $T$ and $y$ refer to the length and the label corresponding to the video $\{\bm{v}_1, \bm{v}_2, \ldots\}$, respectively. In this stage, the model learns to extract task-relevant information from an arbitrary sequence of frame patches, laying the basis for training the policy network $\pi$.

\textbf{Stage II: Learning to select informative patches.}
At this stage, we fix the two encoders ($f_{\textnormal{G}}$ and $f_{\textnormal{L}}$) and the classifier $f_{\textnormal{C}}$ obtained from stage I, and evoke a randomly initialized policy network $\pi$ to be trained with reinforcement learning. Specifically, after sampling a location of $\tilde{\bm{v}}_t$ from $\pi(\cdot|{\bm{e}}^{\textnormal{G}}_{1},\ldots, {\bm{e}}^{\textnormal{G}}_{t})$ for the frame $\bm{v}_t$ (see Eq. (\ref{eq:select_location})), $\pi$ will receive a reward $r_t$ indicating whether this action is beneficial. We train $\pi$ to maximize the sum of discounted rewards:
% The training objective of $\pi$ is to maximize the discounted sum of rewards:
\begin{equation}
    \label{eq:stage_2}
    \mathop{\textnormal{maximize}}_{\pi}\ \ \  \mathbb{E}_{\tilde{\bm{v}}_t\sim\pi(\cdot|{\bm{e}}^{\textnormal{G}}_{1},\ldots, {\bm{e}}^{\textnormal{G}}_{t})}
    \left[
        \sum\nolimits_{t=1}^{T} \gamma^{t-1} r_t
    \right],
\end{equation}
where $\gamma\!\in\!(0, 1)$ is a discount factor for long-term rewards. In our implementation, we fix $\gamma\!=\!0.7$ and solve Eq. (\ref{eq:stage_2}) using the off-the-shelf proximal policy optimization (PPO) algorithm \cite{schulman2017proximal}. Notably, here we directly train $\pi$ on the basis of the features extracted by $f_{\textnormal{G}}$, since previous works \cite{zhou2016learning, selvaraju2017grad} have demonstrated that the vision backbones learned for recognition generally excel at localizing task-relevant regions with their deep representations.

% , and it also performs well empirically.

Ideally, the reward $r_t$ is expected to measure the value of selecting $\tilde{\bm{v}}_t$ in terms of video recognition. With this aim, we define $r_t$ as:
\begin{equation}
    \label{eq:reward}
    \begin{split}
        &r_t(\tilde{\bm{v}}_t|\tilde{\bm{v}}_1, \ldots, \tilde{\bm{v}}_{t-1}) \\=\  &p_{ty}(\tilde{\bm{v}}_t|\tilde{\bm{v}}_1, \ldots, \tilde{\bm{v}}_{t-1}) \\  &- \ {\mathbb{E}}_{\tilde{\bm{v}}_t\sim\textnormal{RandomCrop}(\bm{v}_t)}\left[p_{ty}(\tilde{\bm{v}}_t|\tilde{\bm{v}}_1, \ldots, \tilde{\bm{v}}_{t-1})\right],
    \end{split}
\end{equation}
where $p_{ty}$ refers to the softmax prediction on $y$ (\emph{i.e.}, confidence on the ground truth label). When computing $r_t$, we assume all previous patches $\{\tilde{\bm{v}}_1, \ldots, \tilde{\bm{v}}_{t-1}\}$ have been determined, while only $\tilde{\bm{v}}_t$ can be changed. The second term in Eq. (\ref{eq:reward}) refers to the expected confidence achieved by the randomly sampled $\tilde{\bm{v}}_t$. By introducing it we ensures $\mathbb{E}_{\tilde{\bm{v}}_t}[r_t]=0$, which is empirically found to yield a more stable training procedure. In our experiments, we estimate this term with a single time of Monte Carlo sampling. Intuitively, Eq. (\ref{eq:reward}) encourages the model to select the patches that are capable of producing confident predictions on the correct labels with as fewer frames as possible.

\textbf{Stage III: Fine-tuning.}
Finally, we fine-tune $f_{\textnormal{L}}$ and $f_{\textnormal{C}}$ (or only $f_{\textnormal{C}}$) with the learned policy network $\pi$ from stage II, namely minimizing Eq. (\ref{eq:stage_1}) with $\tilde{\bm{v}}_t\sim\pi(\cdot|{\bm{e}}^{\textnormal{G}}_{1},\ldots, {\bm{e}}^{\textnormal{G}}_{t})$. This stage further improves the performance of our method.

\section{Implementation of Dynamic Frame Sampling}
\label{app:implementation_dynamic_frame_sampling}

Our implementation of dynamic frame sampling minimizes Eq. (\ref{eq:final_temporal_obj}) without notable additional cost. This is achieved by utilizing the features of the $T_{\textnormal{G}}$ uniformly sampled frames processed by the global encoder $f_{\textnormal{G}}$. We approximate the $T_0 \!\Rightarrow\! T_{\textnormal{L}}$ sampling process with $T_{\textnormal{G}} \!\Rightarrow\! \lfloor \frac{T_{\textnormal{L}}}{T_0}T_{\textnormal{G}} \rfloor$ weighted sampling, where the weights are able to be acquired by down-sampling $\{w_1, \ldots, w_{T_0}\}$. Subsequently, $\mathcal{L}_{\textnormal{frame}}(\bm{v}_{n_i})$ can be effortlessly obtained with the off-the-shelf outputs of the auxiliary linear classifier, \emph{i.e.}, $L_{\textnormal{CE}}(\textnormal{SoftMax}(\textnormal{FC}_{\textnormal{G}}(\overline{\bm{e}}^{\textnormal{G}}_{n_i})), y)$ (see: Eq. (\ref{eq:l_prime})). Note that when $f_{\textnormal{G}}$ is implemented as video backbones with temporal fusion operations (\emph{e.g.}, 3D convolution or temporal self-attention), $\mathcal{L}_{\textnormal{frame}}(\bm{v}_{n_i})$ actually represents the loss of a short video clip centred at $\bm{v}_{n_i}$. Moreover, at test time, to establish a deterministic inference procedure, we simply compute the cumulative distribution function of a single-time weighted sampling distribution parameterized by $\{w_1, \ldots, w_{T_0}\}$, and take $T_{\textnormal{L}}$ uniform quantile points on the vertical axis to find the corresponding positions on the horizontal axis.

\section{Results in Real-world Application Scenarios}
\label{app:real_app_results}

\textbf{Setups.}
This subsection evaluates Uni-AdaFocus in three representative realistic application scenarios. The details of our experimental setups can be found in Appendix \ref{app:real_app}.

\textbf{Fine-grained diving action classification}
 on Diving48 \cite{li2018resound}. The results are reported in Table \ref{tab:diving48}. Our method is compared with both the basic baselines (\emph{i.e.}, TSM/TSM+) and representative recently proposed methods that achieve current state-of-the-art performance. Uni-AdaFocus attains the best test accuracy with a minimal computational cost. In particular, it improves the accuracy by 7.7\% (88.1\% v.s. 80.4\%) with 1.87x less computation on top of TSM+, which has a static computational graph. This may be explained by that Uni-AdaFocus is more effective for less biased scenarios like Diving48, where the task-relevant information is intensive, and it is lossless or even beneficial to filter out the less informative image regions and redundant video frames.

\textbf{Alzheimer's and Parkinson's diseases diagnosis with brain magnetic resonance images (MRI).}
Alzheimer's and Parkinson's diseases are two of the most common neurodegenerative disorders among the elderly \cite{scheltens2021alzheimer, bloem2021parkinson, dauer2003parkinson}, characterized by the irreversible loss of neurons as well as the impairment of cognitive and motor functions. The accurate diagnosis of them and early intervention at the prodromal stage is of great importance, with which the onset of the diseases can be significantly delayed. To address this issue, analyzing the MRI of potential patients is a widely used technique, where MRI depict the 3D structure of human brains in a non-invasive fashion. Our proposed method can be deployed as an effective approach for the data-driven automatic analysis of 3D MRI data, under the goal of improving the diagnosis precision and reducing the reliance on experienced clinicians, which are typically scarce. As shown in Table \ref{tab:mri}, Uni-AdaFocus is able to outperform the state-of-the-art medical imaging analysis frameworks on ADNI \cite{url_1}, OASIS \cite{url_2}, and PPMI \cite{url_3} in terms of the accuracy of detecting Alzheimer's and Parkinson's diseases. Notably, many of the baselines leverage generative models, Transformer networks, or 3D convolution, which are more complicated and computationally intensive than our method.

\begin{table}[!t]
  \centering
  \begin{footnotesize}
  \caption{\label{tab:diving48}\textbf{Fine-grained diving action classification on Diving48 \cite{li2018resound}}. MN2/R50/R101 denotes MobileNet-V2/ResNet-50/ResNet-101. The \textcolor{blue}{blue} texts highlight the comparisons with the \underline{underlined} baselines.
  }
  \vskip -0.15in
  \setlength{\tabcolsep}{1mm}{
  \renewcommand\arraystretch{1.075}
  \resizebox{0.95\columnwidth}{!}{
  \begin{tabular}{c|cc|cc} 
  \toprule
  Method & Backbones  & \#Frames   &  \footnotesize{{\ Top-1 Acc.}} & \footnotesize{{GFLOPs}}\\  
  % \midrule
  \midrule
  SlowFast-R101 \cite{feichtenhofer2019slowfast} & {{R101}} & -  & \ 77.6\% & 213 $\times$ 3 \\
  TimeSformer \cite{bertasius2021space} & TimeSformer & -  & \ 75.0\% & 196 $\times$ 3 \\
  TimeSformer-HR \cite{bertasius2021space} & TimeSformer & -  & \ 78.0\% & 1703 $\times$ 3 \\
  TimeSformer-L \cite{bertasius2021space} & TimeSformer & -  & \ 81.0\% & 2380 $\times$ 3 \\
  RSANet-R50 \cite{kim2021relational} & R50 & 16 & \ 84.2\% & 72 $\times$ 2 \\
  \midrule
  TSM \cite{lin2019tsm}  & {{R50}} & 8  & \ 77.4\% & 32.7 \\
  TSM+ \cite{lin2019tsm}   & {{MN2+R50}} & 8+8 & \ \underline{80.4\%}  & \underline{35.1}  \\
  \multirow{2}{*}{\shortstack{Uni-AdaFocus-TSM (128$^2$)\\[-0.7ex]w/o sample-wise dynamic}}  & \multirow{2}{*}{{{MN2+R50}}} & \multirow{2}{*}{8+12} & \multirow{2}{*}{\ \textbf{88.1\%} \scriptsize{(\textcolor{blue}{$\uparrow$7.7\%})}} & \multirow{2}{*}{\textbf{18.8} \scriptsize{(\textcolor{blue}{$\downarrow$1.87x})}} \\
  &&&&\\
  \bottomrule
  \end{tabular}}}
  \end{footnotesize}
  \vspace{-2ex}
  \end{table}

\begin{table}[!t]
  \centering
  \begin{footnotesize}
  \caption{\label{tab:mri}\textbf{Alzheimer's and Parkinson's diseases diagnosis utilizing MRI on ADNI \cite{url_1}, OASIS \cite{url_2}, and PPMI \cite{url_3}}. For backbones, MN2/R/RX/D denotes MobileNet-V2/ResNet/ResNeXt/DenseNet. The \textcolor{blue}{blue} texts highlight the comparisons with the \underline{underlined} baselines.
  }
  \vskip -0.15in
  \setlength{\tabcolsep}{1.5mm}{
  \renewcommand\arraystretch{1.075}
  \resizebox{0.875\columnwidth}{!}{
  \begin{tabular}{c|c|ccc} 
  \toprule
  \multirow{2}{*}{{Method}} & \multirow{2}{*}{{Backbones}}  &  \multicolumn{3}{c}{{Top-1 Acc.}} \\  
  && ADNI & OASIS & PPMI \\
  % \midrule
  \midrule
  wH-FCN \cite{lian2018hierarchical} & 2D-CNN & 90.3\% & - & - \\
  MRNet \cite{cao2023end} & R50 & 90.1\% & 80.0\% & - \\
  DenseNet \cite{huang2019convolutional} & D121 & - & - & 71.0\% \\
  DBV \cite{zhang2019deep} & WGAN+RX50 & - & - & \underline{76.5\%} \\
  Med3D \cite{chen2019med3d} & 3D-R50 & 90.8\% & 82.6\% & - \\
  \multirow{2}{*}{\shortstack{3D ResNet-101\\[-0.2ex]+Transformer \cite{jang2022m3t}}}  & \multirow{2}{*}{\shortstack{3D-R101\\[-0.2ex]+Transformer}} & \multirow{2}{*}{91.5\%} & \multirow{2}{*}{81.9\%} & \multirow{2}{*}{-}  \\
  &&&&\\
  \multirow{2}{*}{\shortstack{3D DenseNet-201\\[-0.2ex]+Transformer \cite{jang2022m3t}}}  & \multirow{2}{*}{\shortstack{3D-D201\\[-0.2ex]+Transformer}} & \multirow{2}{*}{90.4\%} & \multirow{2}{*}{82.2\%} & \multirow{2}{*}{-}  \\
  &&&&\\
  \multirow{2}{*}{\shortstack{M3T \cite{jang2022m3t}}}  & \multirow{2}{*}{\shortstack{3D-CNN+R50\\[-0.2ex]+Transformer}} & \multirow{2}{*}{93.2\%} & \multirow{2}{*}{\underline{85.3\%}} & \multirow{2}{*}{-}  \\
  &&&&\\
  AutoLoc \cite{cao2023end} & MN2+R18 & 93.4\% & - & - \\
  LEAR \cite{9854196} & R18+3D-CNN & \underline{94.9\%} & - & - \\
  \midrule
  \multirow{2}{*}{\shortstack{Uni-AdaFocus (128$^2$)\\[-0.7ex]w/o sample-wise dynamic}}  & \multirow{2}{*}{{{MN2+R50}}}  & \multirow{2}{*}{\ \shortstack{\textbf{97.6\%}\\[-0.2ex]\scriptsize{(\textcolor{blue}{$\uparrow$2.7\%})}}}  & \multirow{2}{*}{\ \shortstack{\textbf{88.5\%}\\[-0.2ex]\scriptsize{(\textcolor{blue}{$\uparrow$3.2\%})}}}& \multirow{2}{*}{\ \shortstack{\textbf{82.0\%}\\[-0.2ex]\scriptsize{(\textcolor{blue}{$\uparrow$5.5\%})}}}   \\
  &&&&\\
  \bottomrule
  \end{tabular}}}
  \end{footnotesize}
  \vspace{-2ex}
\end{table}

\begin{table}[!t]
  \centering
  \begin{footnotesize}
  \caption{\label{tab:rlvs}\textbf{Violence recognition for online videos based on the RLVS dataset \cite{soliman2019violence}}. MN2/R50 denotes MobileNet-V2/ResNet-50. The \textcolor{blue}{blue} texts highlight the comparisons with the \underline{underlined} baselines.
  }
  \vskip -0.15in
  \setlength{\tabcolsep}{1.1mm}{
  \renewcommand\arraystretch{1.075}
  \resizebox{0.86\columnwidth}{!}{
  \begin{tabular}{c|c|cc} 
  \toprule
  Method & Backbones    &  \footnotesize{{\ Top-1 Acc.}} & \footnotesize{{GFLOPs}}\\  
  % \midrule
  \midrule
  TimeSformer \cite{bertasius2021space} & TimeSformer & \ 79.0\% & - \\
  2D BiGRU-CNN \cite{traore20202d} & VGG & \ 90.3\%  & - \\
  CNN+LSTM \cite{soliman2019violence} & VGG & \ 88.2\%  & - \\
  CNN+LSTM \cite{de2021temporal} & VGG & \ 91.0\%  & - \\
  \multirow{2}{*}{\shortstack{Data Efficient\\[-0.2ex]Transformer \cite{abdali2021data} }}  & \multirow{2}{*}{\shortstack{VGG\\[-0.2ex]+Transformer}} & \multirow{2}{*}{\ \underline{96.3\%}} & \multirow{2}{*}{\underline{625.2}}  \\
  &&& \\
  \midrule
  \multirow{2}{*}{\shortstack{Uni-AdaFocus (128$^2$)\\[-0.7ex]w/o sample-wise dynamic}}  & \multirow{2}{*}{{{MN2+R50}}}  & \multirow{2}{*}{\ \textbf{98.5\%} \scriptsize{(\textcolor{blue}{$\uparrow$2.2\%})}}  & \multirow{2}{*}{\textbf{27.2} \scriptsize{(\textcolor{blue}{$\downarrow$23.0x})}} \\
  &&\\
  \bottomrule
  \end{tabular}}}
  \end{footnotesize}
  \vspace{-2ex}
\end{table}

\textbf{Violence recognition for online videos.}
Realizing highly accurate automatic recognition of violent behaviors is a practically important task, \emph{e.g.}, to detect harmful visual contents on the Internet in real-time or to build surveillance systems in the real world to increase safety. The gains of Uni-AdaFocus are also significant for such realistic applications. As shown in Table \ref{tab:rlvs}, our method achieves a test accuracy of 98.5\% on RLVS \cite{soliman2019violence}, which outperforms all the competitive baselines, and attains a new state-of-the-art performance. Moreover, the computational cost of our method is 23.0x less than the previous best baseline in \cite{abdali2021data}, which is beneficial for developing real-time video processing applications.

\section{Additional Discussions}
\label{app:additional_discussions}

\textbf{Comparisons with AdaFocusV3 \cite{wang2022adafocusv3}.}
Our work is related to \cite{wang2022adafocusv3} in the concept of training $\pi$ to localize informative patches using deep-feature-based supervision signals. Nevertheless, our simple but elegant design addresses several critical limitations of \cite{wang2022adafocusv3}, as discussed in Table \ref{tab:_discuss_vs_adaV3}. The method in \cite{wang2022adafocusv3} increases the input size of $f_{\textnormal{L}}$ during training to probe the context information of the patch $\tilde{\bm{v}}_t$, and leverages the resulting enlarged local feature $\bm{e}^{\textnormal{L}}_{t}$ to estimate the gradients of $\pi$. In contrast, the size of training/test inputs of $f_{\textnormal{L}}$ is consistent in Uni-AdaFocus, which not only enhances the performance by eliminating the training/test discrepancy for $f_{\textnormal{L}}$, but also avoids additional training cost (with the same epochs). Besides, the cleverness of Uni-AdaFocus is also manifested in employing the off-the-shelf global feature $\bm{e}^{\textnormal{G}}_{t}$ from $f_{\textnormal{G}}$ to produce gradients. It can be obtained without extra cost, while the global-level semantic information encompassed by $\bm{e}^{\textnormal{G}}_{t}$ demonstrates superior efficacy than the local feature $\bm{e}^{\textnormal{L}}_{t}$ in guiding the training of $\pi$.

\textbf{Comparisons with OCSampler \cite{lin2022ocsampler}.}
Both Uni-AdaFocus and \cite{lin2022ocsampler} formulate frame selection as a sequential weighted sampling problem without replacement, where the distribution is dynamically parameterized conditioned on each video. Compared to \cite{lin2022ocsampler}, our contribution lies in developing a novel solving algorithm for this basic formulation. Our theoretical analyses (Section \ref{sec:frame_sample}) propose to consider the expected loss over dynamic frame sampling as the training objective, and reveal that it can be decomposed into a differentiable form solved with the Monte Carlo method, leading to an efficient end-to-end optimization procedure. As shown in Table \ref{tab:_discuss_vs_oc}, our method works reasonably well without the multi-stage training and reinforcement learning techniques utilized in \cite{lin2022ocsampler}, while it eliminates the factorial time complexity of the algorithm. Empirically, it reduces the real training time by 2.5x, yet improves the mAP considerably (77.4\% v.s. 74.9\%).

\begin{table*}[!t]
  \centering
  \begin{footnotesize}
  \caption{
    \textbf{Uni-AdaFocus v.s. AdaFocusV3 \cite{wang2022adafocusv3} in terms of the algorithm for learning spatial dynamic computation strategies.}
    The three major advantages of Uni-AdaFocus over AdaFocusV3 are systematically removed to evaluate the performance of the ablated versions of our method. 
    Notably, the training cost can be affected by either varying the number of training epochs or adjusting the input size of $f_{\textnormal{L}}$ during training.
    }
  \label{tab:_discuss_vs_adaV3}
  \vskip -0.2in
  \setlength{\tabcolsep}{2.5mm}{
  \vspace{5pt}
  \renewcommand\arraystretch{1.1}
  \resizebox{1.95\columnwidth}{!}{
  \begin{tabular}{c|ccc|ccccc}
  \toprule
  \multicolumn{1}{c|}{\multirow{3}{*}{\shortstack{Approaches for Learning Spatial\\[-0.25ex]Dynamic Computation Strategies\\[-0.25ex]($P^2$=128$^2$)}}} & 
  \multirow{3}{*}{\shortstack{w/o Additional\\[-0.25ex]Training Cost}} & 
  \multirow{3}{*}{\shortstack{Global Information\\[-0.25ex]Guidance}} & 
  \multirow{3}{*}{\shortstack{w/o Discrepancy\\[-0.25ex]between the Training/Test\\[-0.25ex]Inputs of $f_{\textnormal{L}}$}}  &  
  \multicolumn{5}{c}{ActivityNet mAP after Processing $t$ Frames} \\[-0.25ex]
  &  &  &  & \multicolumn{5}{c}{(\emph{i.e.}, corresponding to $\bm{p}_t$)} \\
  &  &  &  & $t$=1 & $t$=2 & $t$=4 & $t$=8 & $t$=16\\
  \midrule
  Uni-AdaFocus {\scriptsize(spatial adaptive}
  & \multirow{1.75}{*}{\textcolor{red}{\xmark}{\scriptsize\ \ [$E$ epochs]}}
  & \multirow{1.75}{*}{\textcolor{red}{\xmark}{\scriptsize\ \ [supervision from $\bm{e}^{\textnormal{L}}_{t}$]}}
  & \multirow{1.75}{*}{\textcolor{red}{\xmark}{\scriptsize\ \ [$(P+32)^2\!\to\!P^2$]}}
  & 
  \multirow{1.75}{*}{45.2\%} & 
  \multirow{1.75}{*}{57.1\%} & 
  \multirow{1.75}{*}{65.7\%} & 
  \multirow{1.75}{*}{72.7\%} & 
  \multirow{1.75}{*}{77.7\%} 
  \\[-0.8ex]
  {\scriptsize policy: learned with AdaFocusV3 \cite{wang2022adafocusv3})} 
  &&&&&&&&
  \\
  -- % Uni-AdaFocus$\bm{\ -\ -}$  
  & \textcolor{_green}{\cmark}{\scriptsize\ \ [$\frac{P^2}{(P+32)^2}E$ epochs]}
  & \textcolor{red}{\xmark}{\scriptsize\ \ [supervision from $\bm{e}^{\textnormal{L}}_{t}$]}
  & \textcolor{red}{\xmark}{\scriptsize\ \ [$(P+32)^2\!\to\!P^2$]}
  & 44.5\% & 56.7\% & 64.8\% & 71.9\% & 77.3\% \\
  -- % Uni-AdaFocus$\bm{\ -}$   
  & \textcolor{_green}{\cmark}{\scriptsize\ \ [$\frac{P^2}{(P+32)^2}E$ epochs]}
  & \textcolor{_green}{\cmark}{\scriptsize\ \ [supervision from $\bm{e}^{\textnormal{G}}_{t}$]}  
  & \textcolor{red}{\xmark}{\scriptsize\ \ [$(P+32)^2\!\to\!P^2$]}
  & 44.9\% & 57.4\% & 65.9\% & 73.3\% & 78.6\% \\
  -- % Uni-AdaFocus$\bm{\ -}$   
  & \textcolor{red}{\xmark}{\scriptsize\ \ [$E$ epochs]}
  & \textcolor{_green}{\cmark}{\scriptsize\ \ [supervision from $\bm{e}^{\textnormal{G}}_{t}$]}  
  & \textcolor{red}{\xmark}{\scriptsize\ \ [$(P+32)^2\!\to\!P^2$]}
  & 45.1\% & 57.8\% & 66.5\% & 73.5\% & 79.1\% \\
  Uni-AdaFocus      
  & \textcolor{_green}{\cmark}{\scriptsize\ \ [$E$ epochs]}
  & \textcolor{_green}{\cmark}{\scriptsize\ \ [supervision from $\bm{e}^{\textnormal{G}}_{t}$]}  
  & \textcolor{_green}{\cmark}{\scriptsize\ \ [$P^2\!\to\!P^2$]}
  & \textbf{45.7\%} & \textbf{58.3\%} & \textbf{67.4\%} & \textbf{74.8\%} & \textbf{79.6\%} \\
  \bottomrule
  \end{tabular}}}
  \end{footnotesize}
  % \vskip -0.09in
  \vskip -0.1in
  % \vskip -0.1in
\end{table*}

\begin{table*}[!t]
  \centering
  \begin{footnotesize}
  \caption{
    \textbf{Uni-AdaFocus v.s. OCSampler \cite{lin2022ocsampler} in terms of the algorithm for learning temporal dynamic computation strategies.}
    We replace our dynamic frame sampling algorithm in Uni-AdaFocus with OCSampler \cite{lin2022ocsampler} (implemented with the official code and configurations provided by \cite{lin2022ocsampler}), and compare its performance with our method. All other experimental setups remain unchanged for a fair comparison.
    }
  \label{tab:_discuss_vs_oc}
  \vskip -0.2in
  \setlength{\tabcolsep}{2.5mm}{
  \vspace{5pt}
  \renewcommand\arraystretch{1.1}
  \resizebox{1.90\columnwidth}{!}{
  \begin{tabular}{c|ccc|c}
  \toprule
  \multicolumn{1}{c|}{\multirow{3}{*}{\shortstack{Approaches for Learning Temporal\\[-0.25ex]Dynamic Computation Strategies\\[-0.25ex]($P^2$=128$^2$)}}} & 
  \multirow{3}{*}{\shortstack{End-to-end\\[-0.25ex]Training}} & 
  \multirow{2}{*}{\shortstack{Theoretical Complexity for Learning to\\[-0.25ex]Select $T_{\textnormal{L}}$ Frames from $T_0$ Frames}} & 
  \multirow{2}{*}{\shortstack{Empirical Wall-time\\[-0.25ex]Training Cost}}  &  
    \multirow{3}{*}{
    \shortstack{
      ActivityNet mAP\\[-0.25ex]
      (10 frames, following \cite{lin2022ocsampler})
    }
    } \\[-0.25ex]
  &  &  &  &   \\
  &  & {\scriptsize\emph{($M$: sampling size of the Monte Carlo method)}} & {\scriptsize\emph{(using 4 NVIDIA 3090 GPUs)}} &   \\
  \midrule
  Uni-AdaFocus {\scriptsize(temporal adaptive}
  & \multirow{1.75}{*}{\textcolor{red}{\xmark}{\scriptsize\ \ (Multi-stage + RL)}} 
  & \multirow{1.75}{*}{
    $\mathcal{O}\Big(T_0 + T_{\textnormal{L}} \log T_0 + T_{\textnormal{L}}! \cdot T_{\textnormal{L}}\Big)$
    }
  & \multirow{1.75}{*}{9.6h}
  & \multirow{1.75}{*}{74.9\%} \\[-0.8ex]
  {\scriptsize policy: learned with OCSampler \cite{lin2022ocsampler})} 
  &  &  &  &  \\
  Uni-AdaFocus
  & \textcolor{_green}{\cmark}
  & $\mathcal{O}\Big(T_{\textnormal{L}}(T_0 + T_{\textnormal{L}} \log T_0) \cdot M\Big)$
  & 3.8h \scriptsize{(\textcolor{blue}{$\downarrow$2.5x})}
  & \ \!\textbf{77.4\%} \\[0.1ex]
  \bottomrule
  \end{tabular}}}
  \end{footnotesize}
  % \vskip -0.09in
  \vskip -0.1in
  % \vskip -0.1in
\end{table*}

\begin{figure}[!t]
  % \vskip -0.015in
  % \hskip -0.1in
  \begin{center}
  \centerline{\includegraphics[width=\columnwidth]{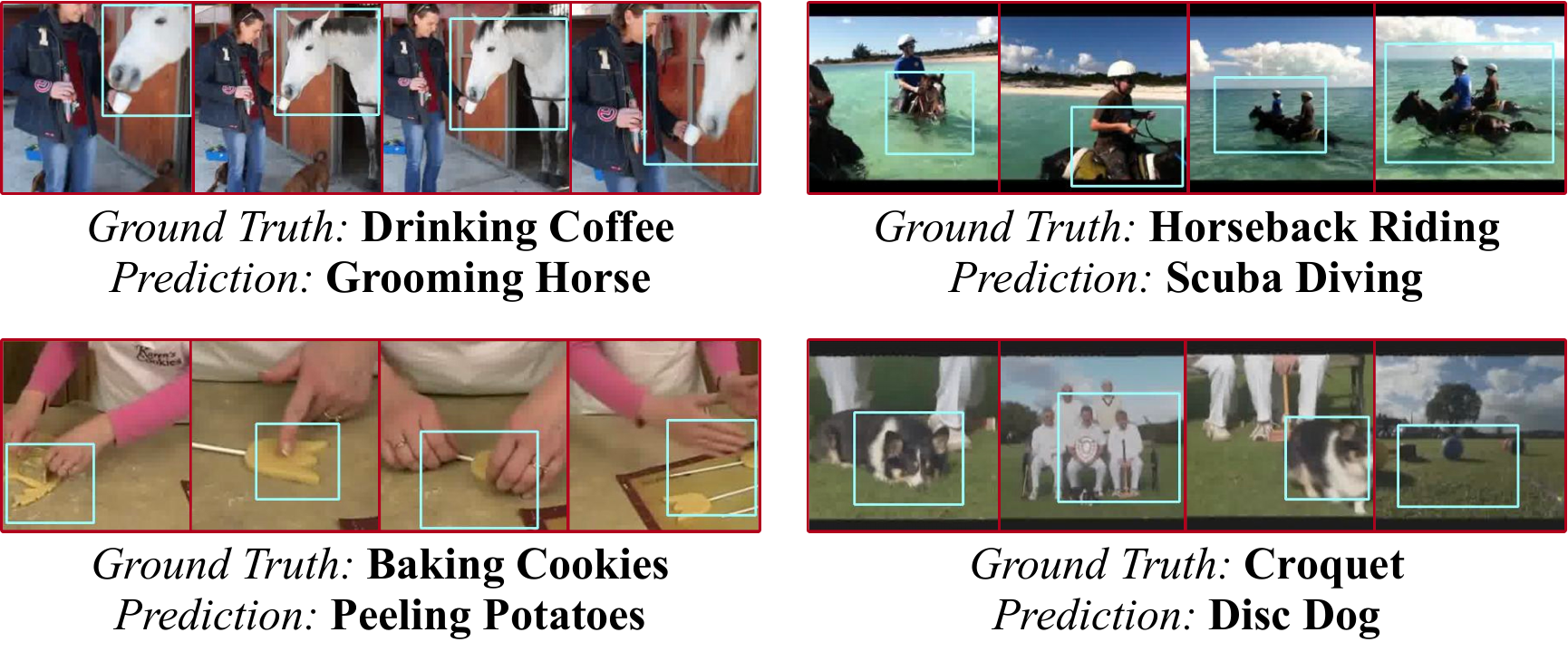}}
  \vskip -0.175in
  \caption{
    \textbf{Representative examples of the failure cases of our model (zoom in for details).}
    Blue boxes indicate the patches selected by Uni-AdaFocus. ``Ground Truth'' and ``Prediction'' denote the ground truth labels of the videos and the predictions of Uni-AdaFocus, respectively.
    \label{fig:fail_case}
  }
  \end{center}
  \vspace{-2ex}
\end{figure}

\begin{figure}[!t]
  % \vskip -0.015in
  % \hskip -0.1in
  \begin{center}
  \centerline{\includegraphics[width=\columnwidth]{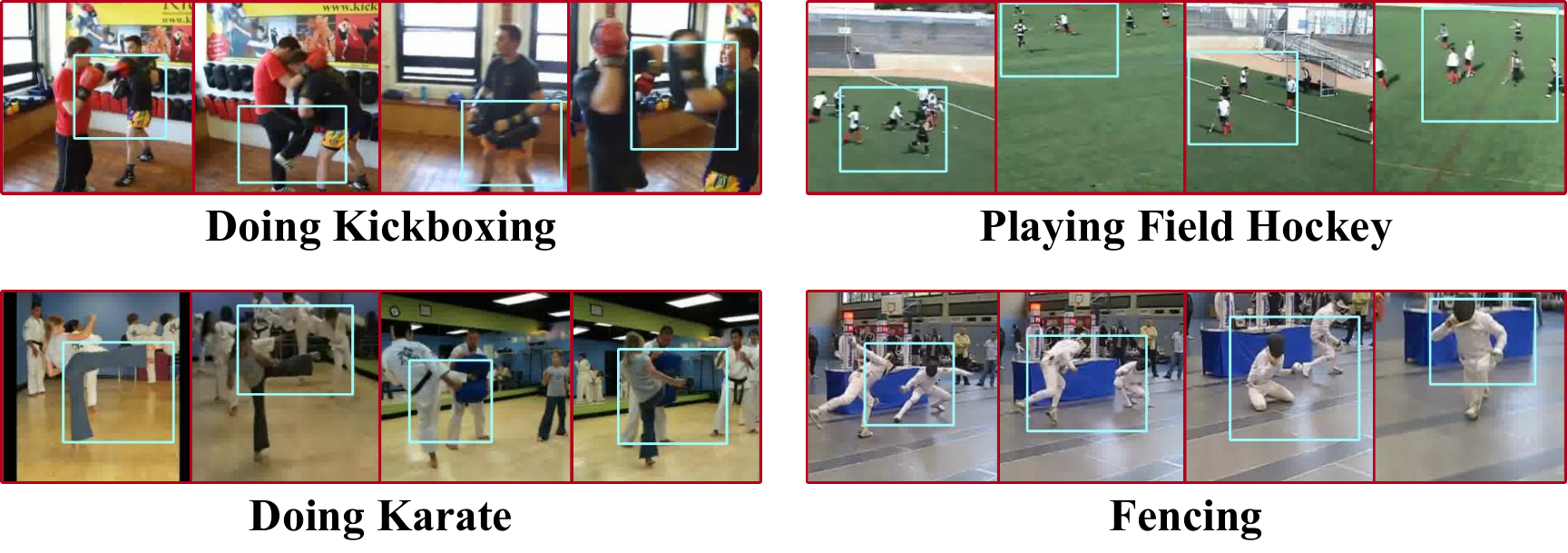}}
  \vskip -0.175in
  \caption{
    \textbf{Additional visualization results focusing primarily on videos that incorporate multi-person/object interactions (zoom in for details).}
    Blue boxes indicate the patches selected by Uni-AdaFocus. 
    \label{fig:multi_person}
  }
  \end{center}
  \vspace{-3ex}
\end{figure}

\section{Additional Visualization Results}
\label{app:additional_visualization}

In Figure \ref{fig:fail_case}, we present an analysis of the representative failure cases of Uni-AdaFocus. We find that our current model exhibits reduced efficacy in understanding the non-typical actions that are rare and may lie within the tail of the data distribution. For instance, videos such as a horse drinking coffee or riding horses in the sea pose challenges for the model. Moreover, it appears to be more challenging for the model to accomplish video understanding tasks incorporating relatively higher demands of reasoning. For example, in the lower left video of Figure \ref{fig:fail_case}, Uni-AdaFocus succeeds in identifying the interactions between human hands and raw potatoes, yet fails to infer that these actions are part of cookie-baking preparation. Similarly, in the lower right video of Figure \ref{fig:fail_case}, Uni-AdaFocus succeeds in localizing task-relevant regions, yet misinterprets the role of the dog in the action. Future works may focus on addressing these limitations of our current model.
Notably, we do not observe particular differences in the patterns of failure cases across different datasets. Hence, we select the general and interesting cases of ActivityNet video recognition (which incorporates a broader range of various interactions) as representative examples.

% Figure \ref{fig:multi_person} showcases supplementary visualizations focusing primarily on videos featuring multi-person/object interactions. The presented results illustrate our method's capability to precisely delineate spatial regions associated with crucial interactions essential for action recognition.

In Figure \ref{fig:multi_person}, we present additional visualizations results that mainly consider the videos incorporating multi-person/object interactions. It can be observed that our method can accurately capture the spatial regions corresponding to the interactions critical for the action to be recognized.

% -Visualization:  Current visualizations mainly showcase single-person/object actions. An evaluation of the algorithm's ability to also accurately capture multi-person/object interactions is suggested.

% The left image depicts a horse in canonical pose and is easy to classify, whereas the right image is
% taken from a rare viewpoint and is likely in the tail of the data distribution.

\section{Datasets}
\label{app:dataset}

% In this section, we present an introduction of the benchmark datasets used in our experiments, as well as our data pre-processing pipeline.

\textbf{Introduction of the datasets.}
Our experiments are based on seven widely-used large-scale video understanding benchmark datasets. For all of them, we use the official training-validation split.
\begin{itemize}
  \item ActivityNet \cite{caba2015activitynet} contains 10,024 training videos and 4,926 validation videos sorted into 200 human action categories. The average duration is 117 seconds.
  \item FCVID \cite{TPAMI-fcvid} contains 45,611 videos for training and 45,612 videos for validation, which are annotated into 239 classes. The average duration is 167 seconds.
  \item Mini-Kinetics is a subset of the Kinetics \cite{kay2017kinetics} dataset. We establish it following \cite{wu2019liteeval, meng2020ar, meng2021adafuse, sun2021dynamic, kim2021efficient, ghodrati2021frameexit}. The dataset includes 200 classes of videos, 121k for training and 10k for validation. The average duration is around 10 seconds \cite{kay2017kinetics}.
  \item Something-Something (Sth-Sth) V1\&V2 \cite{goyal2017something} datasets include 98k and 194k videos respectively. Both of them are labeled with 174 human action classes. The average duration is 4.03 seconds.
  \item Jester \cite{materzynska2019jester} dataset consists of 148,092 videos in 27 action categories. The average duration is 3 seconds.
  \item Kinetics-400 \cite{kay2017kinetics} is a large-scale human action video dataset collected from YouTube, which contains 400 human action classes. It includes 240k videos for training and 20k videos for validation. The average duration is around 10 seconds \cite{kay2017kinetics}.
\end{itemize}

\textbf{Data pre-processing.}
Following \cite{lin2019tsm, meng2020ar,Wang_2021_ICCV}, the training data is augmented via random scaling followed by 224x224 random cropping, after which random flipping is performed on all datasets except for Sth-Sth V1\&V2 and Jester. At test time, since we consider improving the inference efficiency of video recognition, we resize the short side of video frames to 256 and perform 224x224 centre-crop, obtaining a single clip per video for evaluation. On Kinetics-400, we also report the results of 3-crop prediction \cite{feichtenhofer2019slowfast, wang2018non}, where we take three 224x224 crops to cover the spatial dimensions, and average the SoftMax scores for prediction. Notably, these experimental settings are adopted under the consideration of ensuring efficiency. Since our aim is to improve the computational efficiency of the models, we do not consider the settings of more inference views (\emph{e.g.}, 30 views), which dramatically increase the inference cost.

\section{Additional Implementation Details}
\label{app:implementation_details}

\subsection{Details of Network Architectures}
\label{app:net_arch}

\textbf{ActivityNet, FCVID and Mini-Kinetics.}
Following the common practice of our baselines, we adopt MobileNet-V2 \cite{sandler2018mobilenetv2} and ResNet-50 \cite{he2016deep} as the global encoder $f_{\textnormal{G}}$ and local encoder $f_{\textnormal{L}}$ in Uni-AdaFocus. The temporal accumulated max-pooling module proposed in \cite{ghodrati2021frameexit} is deployed as the classifier $f_{\textnormal{C}}$. The policy network $\pi$ adopts an efficient two-branch architecture, corresponding to producing the temporal and spatial dynamic computation policies. Given the input coarse global features with the size $C\!\times\!T\!\times\!H\!\times\!W$, the temporal branch pools them to $C\!\times\!T\!\times\!1\!\times\!1$ (since here we do not need to preserve the spatial information), and processes them with two $3\!\times\!1\!\times\!1$ convolutional layers followed by a fully-connected layer. In contrast, the spatial branch compresses the channel number to $128\!\times\!T\!\times\!H\!\times\!W$ with $1\!\times\!1\!\times\!1$ convolution to reduce the computational cost, and then processes the features with two $3\!\times\!3\!\times\!3$ convolutional layers followed by a fully-connected layer. Compared to directly processing $C\!\times\!T\!\times\!H\!\times\!W$ features, this design eliminates the redundant information in the corresponding branch, introducing minimal computational overheads, which are generally negligible. The training hyper-parameters of our approach can be found in Appendix \ref{app:hyper}.

\textbf{Sth-Sth V1\&V2, Jester and Kinetics-400.}
We implement Uni-AdaFocus on top of the recently proposed efficient network architectures, ConvNets with temporal shift module (TSM) \cite{lin2019tsm} and X3D networks \cite{feichtenhofer2020x3d}, to demonstrate that our method can further improve the efficiency of such state-of-the-art lightweight models. For TSM, we still use MobileNet-V2 and ResNet-50 as $f_{\textnormal{G}}$ and $f_{\textnormal{L}}$, but add TSM to them. Following the original design of TSM \cite{lin2019tsm}, a fully-connected layer is deployed as the classifier $f_{\textnormal{C}}$, and we average the frame-wise predictions as the output. For a fair comparison, we augment the vanilla TSM by introducing the same two backbone networks as ours (named as TSM+), where their output features are also concatenated to be fed into a linear classifier. In other words, TSM+ differentiates itself from our method only in that it feeds the uniformly-sampled whole video frames into ResNet-50, while we feed the dynamically-selected informative image patches of task-relative frames. For X3D, we simply replace MobileNet-V2/ResNet-50 by X3D-S/X3D-L on top of the settings of TSM. Here we directly compare the performance of our method with X3D, since both our two backbones come from the family of X3D networks.

Following the experimental settings in the original papers of TSM \cite{lin2019tsm} and X3D \cite{feichtenhofer2020x3d}, the video recognition task on Sth-Sth V1\&V2, Jester and Kinetics-400 is considered here. We uniformly sample $T_0\!=\!24/36/48/96$ frames from each video, and set $T_{\textnormal{L}}\!=\!8/12/16/32$ correspondingly, with $T_{\textnormal{G}}\!=\!8/16$. 
Notably, as the videos in these datasets are very short (average duration $\approx\!$ 3-4s for Sth-Sth V1\&V2 and Jester, $\approx\!$ 10s for Kinetics-400), we find the networks require the visual angle of adjacent inputs (frames/patches) to be similar for high generalization performance. We also observe that the locations and sizes of task-relevant regions do not significantly change across the frames in the same video (due to the short duration). Therefore, here we let Uni-AdaFocus generate a common patch location/size for the whole video. In addition, as the backbone networks usually necessitate processing all the inputs simultaneously, here our sample-wise conditional-exit is performed by adaptively activating/deactivating $f_{\textnormal{L}}$. Importantly, such simplification does not affect the main idea of our method since different videos have varying patch locations, shapes and sizes, while our method is still able to unevenly allocate computation across ``easy'' and ``hard'' videos. The architecture of the policy network $\pi$ and the training algorithm remain unchanged. Details of training hyper-parameters are deferred to Appendix \ref{app:hyper}.

\subsection{Real-world Application Scenarios}
\label{app:real_app}

In our paper, we evaluate the effectiveness of Uni-AdaFocus in three representative realistic application scenarios, including 1) fine-grained diving action classification \cite{li2018resound}, and 2) Alzheimer's and Parkinson's diseases diagnosis with brain magnetic resonance images (MRI) \cite{url_1, url_2, url_3}, and 3) violence recognition for online videos \cite{soliman2019violence}. For 1), we adopt the Diving48 dataset \cite{li2018resound}, which consists of $\sim$18k trimmed video clips of 48 unambiguous dive sequences. All the videos have similar backgrounds, while the networks need to have a strong ability to model temporal dynamics and distinguish between diving actions with subtle visual differences. Here we implement Uni-AdaFocus-TSM following Section \ref{sec:tsm_uniada}. For 2), we adopt the ADNI \cite{url_1} and OASIS \cite{url_2} datasets for Alzheimer's disease, and the PPMI \cite{url_3} dataset for Parkinson's disease. Since the amount of data in each dataset is relatively small, here we report the accuracy of 5-fold cross validation. The pre-processing of the MRI data follows from \cite{cao2023end} and \cite{zhang2019deep}. For 3), we adopt the RLVS dataset \cite{soliman2019violence}, which contains 1000 violence and 1000 non-violence videos collected from YouTube. For both 2) and 3), we simply utilize the implementation of Uni-AdaFocus in Section \ref{sec:uni_ada_mnv2_r50}, which is sufficient to achieve superior performance. Besides, for these applications, our major focus is to achieve relatively high accuracy, and hence we do not perform sample-wise dynamic computation.

\subsection{Training Hyper-parameters}
\label{app:hyper}

In this subsection, we present the configurations and hyper-parameters for training Uni-AdaFocus. The values of hyper-parameters are determined with the following procedure: we hold out 25\% of the training samples as a mini-validation set to search for the hyper-parameters, and then we put these samples back to train our model on the whole training set with our selected optimal hyper-parameters, with the final performance reported on the official validation or test set. 

\textbf{ActivityNet, FCVID and Mini-Kinetics.}
As stated in the paper, all the components (\emph{i.e.}, $f_{\textnormal{G}}$, $f_{\textnormal{L}}$, $f_{\textnormal{C}}$ and $\pi$) of Uni-AdaFocus are trained simultaneously in a standard end-to-end fashion. An SGD optimizer with cosine learning rate annealing and a momentum of 0.9 is adopted. On ActivityNet, the size of the mini-batch is set to 32, and the L2 regularization co-efficient is set to 2e-4. The number of training epochs is set to 50. The two encoders $f_{\textnormal{G}}$ and $f_{\textnormal{L}}$ are initialized using the official ImageNet pre-trained models provided by PyTorch \cite{paszke2019pytorch}, while $f_{\textnormal{C}}$ and $\pi$ are trained from random initialization. The initial learning rates of $f_{\textnormal{G}}$, $f_{\textnormal{L}}$, $f_{\textnormal{C}}$ and $\pi$ are set to 0.001, 0.002, 0.04 and 0.0004, respectively. The value of $\alpha$ is selected within $\{0.25, 0.5, 0.75, 1\}$ on a per-dataset basis. On FCVID and Mini-Kinetics, due to the relatively larger size of training data, we reduce the L2 regularization co-efficient to 1e-4, and increase the batch size to 64 (the initial learning rates are linearly scaled correspondingly). The experiments on ActivityNet, FCVID and Mini-Kinetics with all patch sizes use the same aforementioned training configurations.

\textbf{Sth-Sth V1\&V2 and Jester.}
When TSM \cite{lin2019tsm} is implemented as the backbones in Uni-AdaFocus, the initial learning rates of $f_{\textnormal{G}}$, $f_{\textnormal{L}}$, $f_{\textnormal{C}}$ and $\pi$ are set to 0.01, 0.02, 0.02 and 0.004, respectively. The L2 regularization co-efficient is set to 1e-3 on Sth-Sth V1 and Jester, and 5e-4 on Sth-Sth V2. These changes mainly follow the official implementation of TSM \cite{lin2019tsm}. All other training settings are the same as the experiments on FCVID and Mini-Kinetics. All the experiments on Sth-Sth V1\&V2 and Jester adopt the same aforementioned training configurations.

\textbf{Kinetics-400.}
When X3D \cite{feichtenhofer2020x3d} is implemented as our backbones, we initialize $f_{\textnormal{G}}$ and $f_{\textnormal{L}}$ with the official pre-trained models provided by \cite{feichtenhofer2020x3d}. The initial learning rates of $f_{\textnormal{G}}$, $f_{\textnormal{L}}$, $f_{\textnormal{C}}$ and $\pi$ are set to 0.0025, 0.005, 0.005 and 0.002, respectively. The L2 regularization co-efficient is set to 5e-5. All other training settings are the same as the experiments on FCVID and Mini-Kinetics. All the experiments on Kinetics-400 adopt the same aforementioned training configurations.

\end{document}